\documentclass{article}

\usepackage{arxiv}

\usepackage[utf8]{inputenc} 
\usepackage[T1]{fontenc}    
\usepackage{hyperref}       
\usepackage{url}            
\usepackage{booktabs}       
\usepackage{amsfonts}       
\usepackage{nicefrac}       
\usepackage{microtype}      
\usepackage{lipsum}		
\usepackage{graphicx}
\usepackage{natbib}
\usepackage{doi}
\usepackage{subcaption}
\usepackage{float}
\usepackage{amsmath}
\usepackage{authblk}
\usepackage{array}

\author[1,4]{Maxime Ryckewaert}
\author[1]{Diego Marcos}
\author[1]{Christophe Botella}
\author[2]{Maximilien Servajean}
\author[3]{Pierre Bonnet}
\author[1]{Alexis Joly}

\affil[1]{Inria, Univ Montpellier, Montpellier, France}
\affil[2]{LIRMM, AMIS, Univ Paul Valéry Montpellier, Univ Montpellier, CNRS, Montpellier, France}
\affil[3]{AMAP, Univ Montpellier, CIRAD, CNRS, INRAE, IRD, Montpellier, France}
\affil[4]{UMR AGAP Institut, Univ Montpellier, CIRAD, INRAE, Institut Agro, Montpellier, France}

\title{Applying the maximum entropy principle to neural networks enhances multi-species distribution models}

\renewcommand{\shorttitle}{\textit{arXiv}}

\hypersetup{
pdftitle={submitted to MEE},
pdfsubject={deep learning and SDM},
pdfauthor={Maxime Ryckewaert},
pdfkeywords={species distribution modelling; neural networks, maximum entropy principle; deepmaxent; deep learning; presence-only data, sampling bias;target-group background},
}

\begin{document}
\maketitle

\begin{abstract}
The rapid expansion of citizen science initiatives has led to a significant growth of biodiversity databases, and particularly presence-only (PO) observations. PO data are invaluable for understanding species distributions and their dynamics, but their use in a Species Distribution Model (SDM) is curtailed by sampling biases and the lack of information on absences.
Poisson point processes are widely used for SDMs, with Maxent being one of the most popular methods. Maxent maximises the entropy of a probability distribution across sites as a function of predefined transformations of variables, called features.  
In contrast, neural networks and deep learning have emerged as a promising technique for automatic feature extraction from complex input variables. Arbitrarily complex transformations of input variables can be learned from the data efficiently through backpropagation and stochastic gradient descent (SGD). Yet, deep learning was mainly developed for classification problems, and learning robust features and species abundances across space while properly correcting for sampling biases has remained a challenge so far.
In this paper, we propose DeepMaxent, which harnesses neural networks to automatically learn shared features among species, using the maximum entropy principle. 
To do so, it employs a normalised Poisson loss where for each species, presence probabilities across sites are modelled by a neural network. 
We evaluate DeepMaxent on a benchmark dataset known for its spatial sampling biases, using PO data for calibration and presence-absence (PA) data for validation across six regions with different biological groups and covariates. 
Our results indicate that DeepMaxent performs better than Maxent and other leading SDMs across all regions and taxonomic groups. The method performs particularly well in regions of uneven sampling, demonstrating substantial potential to increase SDM performances. 
The method opens the possibility to learn more robust features predicting simultaneously many species to arbitrary large datasets without increased memory requirements. The model likelihood, arising from a Poisson process, makes the method compatible with the integration of more standardised types of data to further increase sampling bias correction. 
In particular, our approach yields more accurate predictions than traditional single-species models, which opens up new possibilities for methodological enhancement.

\end{abstract}


\section{Introduction}
\label{introduction}

In recent years, the rapid {growth} of citizen science projects has contributed significantly to the expansion of biodiversity databases. {Among the different types of data collected, a large amount consists of presence-only (PO) observations}~\citep{callaghan_benefits_2022,bonnet_how_2020}. PO records have been instrumental {in improving} our understanding of species distributions and {helping} inform conservation strategies~\citep{carvalho_conservation_2011,guisan_predicting_2013}. 


Maxent~\citep{phillips_maximum_2006} is one of the most widely used and effective methods for {Species Distribution Model (SDM) b}ased on PO data ~\citep{warren_ecological_2011, elith_novel_2006,elith_presence-only_2020,valavi_predictive_2022}. Maxent generates a {relative probability of species} occurrence across {space as a function of environmental variables. This function is applied to various predefined transformations of input environmental variables. Maxent's output can be interpreted as the probability of observing the species in each site relative to the other sites and knowing that it has been observed once. This probability estimate of Maxent is actually equivalent to the one that is derived from a related Poisson regression or a spatially discretized Poisson process~\citep{renner2013equivalence}.} Maxent's name arises from the fact that {its formulation leads to finding, among all solutions that fit the PO training data, the one that maximizes the entropy of this spatial probability distribution. This behaviour, in which spatially smooth solutions are favoured for regions with sparse PO data, participated to Maxent's robust performances against many other SDM methods under such data regimes and in diverse contexts~\citep{elith_novel_2006,valavi_predictive_2022}.}

{Yet, a} major issue when calibrating SDM{s} using PO data is spatial sampling bias, which leads to clustering of PO records in areas with high sampling effort, typically of higher accessibility or greater human activity. Such bias can distort SDM outputs, leading to inaccurate species distribution estimates~\citep{yackulic_presence-only_2013, fithian_bias_2015, phillips_sample_2009}. 


A wide range of strategies have been proposed to correct for spatial sampling bias in SDMs, including methods based on background points manipulation, spatial filtering of records or explicit bias modelling~\citep{phillips_sample_2009,boria_spatial_2014,fithian_bias_2015}. Specifically for Maxent,~\citep{phillips_sample_2009} proposed the Target-Group Background correction (hereafter TGB), which restricts the background sites used by Maxent to those where at least one species was observed among a Target Group of species. This Target Group should contain species being sampled along with the focal one. \cite{phillips_sample_2009} evaluated the TGB correction with Maxent on a large standardized dataset. This correction was also robust in later studies~\citep{fourcade_mapping_2014}{. Besides its simplicity and empirical robustness, the TGB correction also comes with theoretical guarantees. Indeed, with the assumptions that all species occurrences are drawn from independent Poisson point processes thinned by a same sampling bias, as it is often assumed~\citep{fithian_bias_2015,botella2021jointly}, and that the TG species cumulated intensities are constant, TGB yields an unbiased estimate of the species relative intensity across sites~\citep{botella_bias_2020}, which then applies within Maxent~\citep{renner2013equivalence}.} More recent studies have proposed other bias corrections in the context of deep learning based SDMs, by adapting the loss function to weight presences and background points~\citep{zbinden_selection_2024,gillespie_deep_2024}.

Feature design, i.e. the definition of pre-defined transformations {(features)} of the input variables, is an important step in traditional SDMs, including Maxent~\citep{phillips2008modeling,komori_cumulant-based_2024}. Deep learning is a family of data-driven methods that removes the need for feature design by allowing the model to learn arbitrary non-linear features from the data using neural networks, backpropagation and stochastic gradient descent~\citep{goodfellow_deep_2016,hornik_multilayer_1989,lecun_backpropagation_1989}. {While other approaches are also capable of learning non-linear relationships, such as Generalized Additive Models (GAM), Multivariate Adaptive Regression Splines (MARS) or Boosted Regression Trees (BRT), and have been used for SDMs~\citep{phillips_sample_2009}, deep learning methods offer a broader and more flexible class of models that can automatically learn rich, hierarchical features while integrating efficient regularisation strategies to mitigate overfitting ~\citep{lecun_deep_2015,schmidhuber_deep_2015}.} {Additionally, like other multi-species SDMs do, although typically restricted to  the linear case}~\citep{ovaskainen2017make,van2023concurrent}.{ Deep learning based SDM (deepSDMs) can also learn shared features to simultaneously predict multiple species distributions. Furthermore, these learnt features tend to be more predictive and robust the more species are included ~\citep{chen2017deep,botella2018deep}, leading to a recent interest in deep learning for multi-SDMs ~\citep{kellenberger_performance_2024}. In addition, deepSDM architectures can capture predictive features from structured and high-dimensional input data, such as remote sensing imagery ~\citep{deneu_convolutional_2021, estopinan_modelling_2024,deneu_convolutional_2021,estopinan2022deep}.} {In spite of this, deepSDM performance have remained limited so far when using} low-dimensional input variables ~\citep{zbinden_selection_2024}. Besides, they remain susceptible to sampling biases ~\citep{zbinden_selection_2024}, and potentially amplifying them due to their capacity of fitting arbitrary functions.

\begin{figure}[htbp]
    \centering
    \includegraphics[width=0.45\linewidth]{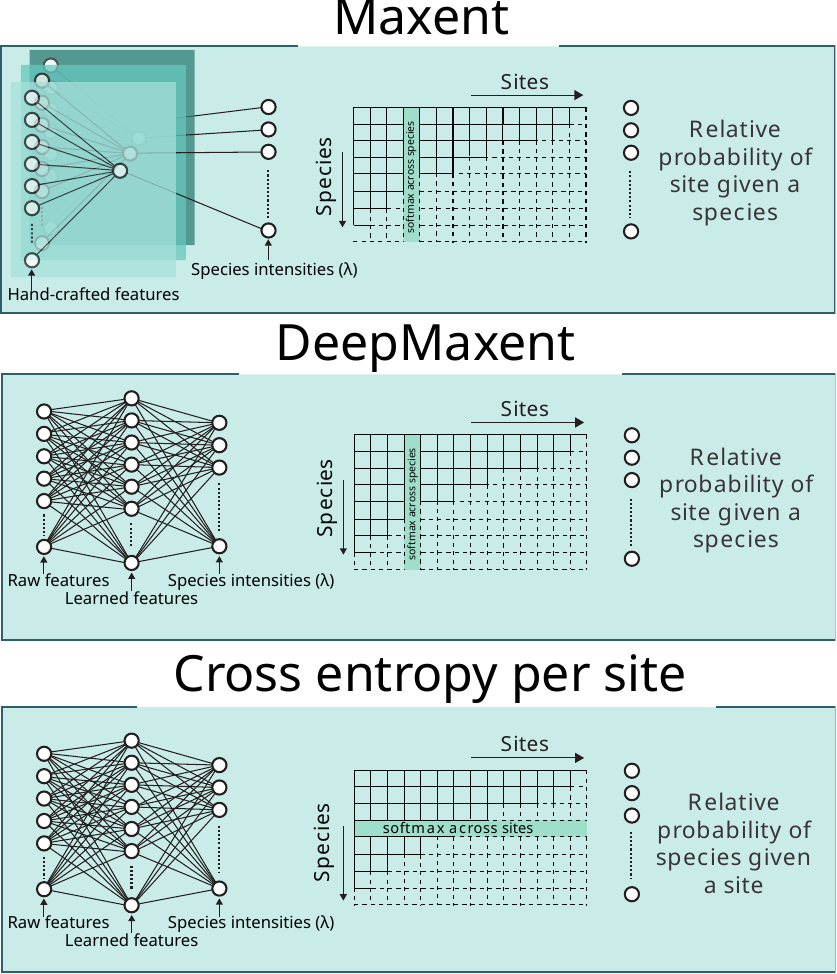}
    \caption{Illustration of {three species distribution modelling approaches (Maxent, DeepMaxent, and a common cross-entropy). 
    Maxent relies on handcrafted environmental features and trains an independent model for each species. DeepMaxent uses a single deep learning model to predict multiple species simultaneously, incorporating batch normalisation across samples to standardize feature representations. In contrast, the commonly used cross-entropy loss approach applies normalisation along the species dimension, focusing on predicting the presence of species at each site rather than modelling species jointly.}
    }
    \label{fig:SDMorDeepSDM}
\end{figure}

In this study, we propose DeepMaxent, a method that combines the Maxent principle of maximum entropy with the {data-driven feature extraction capabilities of deep learning methods}.
The DeepMaxent model {uses PO data to jointly learn shared} latent features {and the functions that map from these features to the probability distribution across sites for each species.}
{We propose a loss function, hereafter the DeepMaxent loss, that generalizes Maxent for modelling the probability function with a wider class of functions, including neural networks, preserving the equivalence with the Poisson regression loss~\citep{renner2013equivalence}}.
In contrast to loss functions often used for deepSDMs, which attempt at modelling the {relative} probability of each species given a site, DeepMaxent aims at modelling the probability of selecting each site for an observation given a species{, as done in Maxent} (see Figure~\ref{fig:SDMorDeepSDM}).
{Note that the latter is an easier objective, since modelling the relative probability of each species given a site requires capturing the relative abundances between species.}
Unlike the original Maxent loss, which is optimised accounting for the whole dataset at every optimisation step, we adopt a mini batch-based approach to ensure scalability in terms of data and model size. We show that the global minimizer of such loss is the same as the one using the full dataset to inform each optimisation step.
Similarly to Maxent, {we show how the TGB correction can be implicitly incorporated into DeepMaxent to efficiently mitigate spatial sampling bias}.
We evaluate DeepMaxent {and compare it to alternative methods} on {two} reference benchmark{s}~\citep{elith_presence-only_2020,picek_geoplant_2025}, {both} encompassing PO data for {SDM} training and presence-absence (PA) data for evaluation. {We carried an extensive comparative evaluation on the NCEAS dataset ~\citep{elith_presence-only_2020}, comprising} six distinct regions and different biological groups. We compare DeepMaxent to alternative loss functions (Poisson, Cross-Entropy across species, Binary Cross Entropy) with or without the TGB correction, and to various state-of-the-art SDMs, notably Maxent and other multi-species deep learning based SDM. We also conduct sensitivity and ablation studies on this dataset to assess the importance of DeepMaxent's hyper-parameters and components.
We conduct additional experiments on the {GeoPlant dataset~\citep{picek_geoplant_2025}, including an independent comparison of DeepMaxent and Maxent, and an illustration of how DeepMaxent can leverage remote sensing input data with an assessment of the related performance gains compared to tabular climatic data}.

\section{Materials and Methods}





\subsection{DeepMaxent: Maximum entropy principle for SDMs based on neural networks}
\subsubsection{A generalization of Maxent's loss function}

We introduce here the statistical model underlying the investigated methods. {We consider a geographic domain $\mathcal{D} \subset \mathbb{R}^2$ composed of $K$ non-overlapping spatial areas $d_1, ...,d_K \subset \mathcal{D}$ (e.g. a regular mesh), hereafter called sites. We consider $N\in\mathbb{N}^*$ species indexed by $j$, and note $y_{ij}\in\mathbb{N}$ the count of PO observations for species $j$ in site $i\in [1,K]$. For each site $i\in [1,K]$, we have covariates $\mathbf{x}_i\in\mathbb{R}^P$ typically encoding environmental factors (e.g. climate or soil properties), as well as positive intensity values $\lambda_{ij}>0$ for and species $j$. We further assume $\lambda_{ij}$ to be a parametric function of the covariates of the form $\lambda_{ij} = \lambda_j(\mathbf{x}_i)$, where we explicitly define $\lambda_j : \mathbb{R}^P \to \mathbb{R}_+$ as a positive intensity function mapping site-level covariates to expected counts. This function is then parametrized as $\lambda_j(\mathbf{x}_i) = \exp(b_j + f_{\theta_j}(\mathbf{x}_i))$ with real parameters $b_j, \theta_j $. The function $f_{\theta_j}$ can be an arbitrary mapping $f_{\theta_j} : \mathbb{R}^P \to \mathbb{R}$. 
In a Poisson regression setting, the counts $y_{ij}$ are modelled as Poisson random variables $y_{ij}\sim \mathcal{P}(\lambda_{ij})$ where these counts are independent between sites and species. This Poisson regression model notably corresponds to the discrete approximation of a spatial Inhomogeneous Poisson Process (IPP), and is very often used for SDMs~\citep{renner_point_2015} with a known equivalence to Maxent when $f_{\theta_j}$ is linear in $x_i$~\citep{renner2013equivalence}. Such discretization is common and handful to avoid the computational burden of fitting an IPP on a continuous spatial domain where covariates vary at high resolution, and it makes particular sense when covariates come from a stack of geographic rasters.}

{We note for convenience the intensity values (i.e. model predicted counts) across all sites and species by the matrix $\boldsymbol{\Lambda} = \{ \lambda_{ij} \}_{i=1,j=1}^{K,N} \in \mathbb{R}_+^{K \times N}$, and similarly $\mathbf{Y} = \{ y_{ij} \}_{i=1,j=1}^{K,N} \in \mathbb{N}_+^{K \times N}$ for the PO count data. The negative log-likelihood of the above described Poisson regression, hereafter the Poisson loss, is written in equation \ref{eq:poisson_loss}.}

\begin{equation}
\mathcal{L}_{\mathcal{P}}(\boldsymbol{\Lambda}, \mathbf{Y}) = \frac{1}{K N} \sum_{i=1}^{K} \sum_{j=1}^{N} \left( \lambda_{ij} - y_{ij} \log \lambda_{ij} \right),
    \label{eq:poisson_loss}
\end{equation}

{The parameters' estimates for the Poisson loss (i.e. the maximum likelihood estimates of the Poisson regression) can then be noted $(\hat{b}_1^{\mathcal{P}},...,\hat{b}_N^{\mathcal{P}},\hat{\theta}_1^{\mathcal{P}},...,\hat{\theta}_N^{\mathcal{P}})=\underset{b_1,...,b_N,\theta_1,...,\theta_N}{\text{argmin}} \mathcal{L}_{\mathcal{P}}(\boldsymbol{\Lambda},\mathbf{Y})$. It is noteworthy that, due to the equivalence from \cite{renner2013equivalence}, the sites included in the terms of equation \ref{eq:poisson_loss} can be interpreted as the background sites (or points) of Maxent~\citep{phillips_sample_2009}, as developed in the dedicated section below.}

{We introduce in equation \ref{eq:normLoss} a new loss $\mathcal{L}_{\mathcal{H}}$ derived from the Poisson loss of equation \ref{eq:poisson_loss}, named DeepMaxent loss. We can see from this equation that the DeepMaxent loss is a modification of the Poisson loss where the counts $y_{ij}$ and intensity values $\lambda_{ij}$ are normalised by their sum over sites, and each species term is weighted by the species total PO count. The loss measures the discrepancy between the observed and predicted probability distributions of PO across sites for each species. Indeed, equation \ref{eq:normLoss} corresponds to a weighted sum of species-wise cross-entropy losses (or Kullback–Leibler divergences) between the empirical and predicted distribution of PO records over sites. }

\begin{equation}
\label{eq:normLoss}
    \mathcal{L}_\mathcal{H}(\boldsymbol{\Lambda}, \mathbf{Y}) = -\frac{1}{K N} \sum_{j=1}^{N} \sum_{i=1}^{K} y_{ij} \log \left( \frac{\lambda_{ij}}{\sum_{k=1}^{K} \lambda_{kj}} \right) = -\frac{1}{K N}  \sum_{j=1}^{N} (\sum_{k=1}^{K} y_{kj}) \sum_{i=1}^{K} \left( \frac{y_{ij}}{\sum_{k=1}^{K} y_{kj}} \right) \log \left( \frac{\lambda_{ij}}{\sum_{k=1}^{K} \lambda_{kj}} \right)   \\
\end{equation}

{We show in Appendix \ref{annex:equivalence} that the DeepMaxent loss generalizes the loss of Maxent to intensity models $\lambda_{ij}=\exp(b_j + f_{\theta_j}(x_i) )$ which are not necessarily log-linear, e.g. when $f$ is built on neural networks, and obviously to multiple species. That is, if $\lambda_{ij}$ is log-linear, the DeepMaxent loss is equivalent to the one of Maxent. Furthermore, Appendix \ref{annex:equivalence} shows that we preserve the equivalence with the Poisson estimate \cite{renner2013equivalence} for non log-linear intensities: The global minimizer of the DeepMaxent loss for the parameters of the probability distribution across sites $( \hat{\theta}^{\mathcal{H}}_1,...,\hat{\theta}^{\mathcal{H}}_N)$ is equal to the one of the Poisson loss $( \hat{\theta}^{\mathcal{P}}_1,...,\hat{\theta}^{\mathcal{P}}_N)$ of equation \ref{eq:poisson_loss}, and the difference with the Poisson loss is that the latter provides an estimate for $b_j$. Although $b_j$ is not identifiable under the DeepMaxent loss, as it cancels out in equation \ref{eq:normLoss}, we consider it for consistency across all losses introduced and tested below, including the Poisson loss.}

\subsubsection{Feature extraction using neural networks}

In Maxent, the intensity {is defined as a log-linear function of a feature vector $f(\mathbf{x})$, composed of pre-determined transformations of $\mathbf{x}$}. We extend the principle of Maxent by replacing $f(\mathbf{x})$ with a feature extractor instantiated as a neural network $g_\theta: \mathbb{R}^P \mapsto \mathbb{R}^C$ parametrised by $\theta$, where $C$ is the dimensionality of the last hidden representation. {The output $g_\theta(\mathbf{x}) \in \mathbb{R}^C$ is a shared latent representation} {across} all species, as illustrated in Figure \ref{fig:architectures}.
The intensity function $\lambda_j(\mathbf{x})$ of species $j$ in DeepMaxent is then given by:

\begin{equation}
\label{eq:lambda_j}
\lambda_j(\mathbf{x}) = \exp\big(\sum_{c=1}^C \gamma_{jc} g_\theta(\mathbf{x})_c + b_j\big),
\end{equation}
where $\gamma_j \in \mathbb{R}^{C}$ is the species-specific weight {vector} and $b_j\in \mathbb{R}$ {is the species-specific intercept.
Collectively, the weight vectors} for all species form the matrix $\Gamma \in \mathbb{R}^{N \times C}$, and {the intercepts} form the vector $\mathbf{b}\in\mathbb{R}^N$.
The function $g_\theta$ can automatically learn complex, non-linear relationships between environmental variables and the presence of multiple species from the data, potentially enabling the model to identify environmental patterns {that cannot be captured by a linear mapping}.
{With this multi-species architecture, similarly to many earlier deep learning based SDM implementations, we have one shared $g_\theta$ across all species, which offers a key computational advantage compared to training one DeepMaxent model per species, given the complexity of $g_\theta$. Indeed, at each gradient descent step, $g_\theta$ and its gradient are computed only once per site, independent of the number of species. Besides, sharing this  representation of the environment across species yielded more robust performances in deep learning based SDMs~\citep{botella2018deep}, especially for data-poor species.}

\subsubsection{Batched algorithm and partition function approximation in DeepMaxent}

One of the challenges of adapting Maxent to a deep learning framework is the computation of the partition functions $\sum_{k} \lambda_{kj}$ for species $j$, corresponding to the denominator in Equation \ref{eq:normLoss}, which normalises the predicted intensity $\lambda_j(\mathbf{x})$ over the $K$ sites. 
{This may become problematic when the number of observed sites increases, which often happens when considering a large geographic domain or a finer spatial resolution. In this case, the number of terms $K$ in the partition function becomes large, making it challenging to compute it exactly}.
{To address this challenge efficiently, we leverage standard stochastic optimisation techniques widely used in deep learning (e.g., mini-batch Stochastic Gradient Descent~\citep{bottou_stochastic_nodate}), with the specificity that we approximate the normalisation within mini-batches. Specifically, we compute the loss function on a small random subset of sites $B\subset \{1,\dots,K\}$, called mini-batch, hence normalising the intensities within the mini-batch.} 
The mini-batch-wise loss is then written as follows: 

\begin{equation}
\label{eq:batch_loss}
\mathcal{L}(\tilde{\boldsymbol{\Lambda}}_{i\in B},\tilde{\mathbf{Y}}_{i\in B})= - \frac{1}{|B| N}\sum_{i\in B}\sum_{j=1}^N \frac{y_{ij}}{\sum_{i\in B}y_{ij}+\epsilon}\log\left( \frac{\lambda_{j}(\mathbf{x}_i)}{\sum_{i \in B}\lambda_{j}(\mathbf{x}_i) } \right) .
\end{equation}

Note that the denominator $\sum_{i \in B} \lambda_j(\mathbf{x}_i)$ is strictly positive under typical model assumptions where the Poisson intensity functions $\lambda_j(\mathbf{x}_i)$ are positive. However, the denominator $\sum_{i \in B} y_{ij}$, representing the count sum for class $j$ in the mini-batch, may be zero if no samples of class $j$ are present in the mini-batch. In practice, this requires {either} ensuring that each mini-batch contains at least one example of every class to avoid division by zero{, or the addition of a small, but positive, $\epsilon$}.
This mini-batch-wise loss makes the model optimisation computationally feasible by computing it one mini-batch at a time {and avoiding computing the full partition functions for every iteration}, and thus allows it to be scalable to large domains and trained efficiently with any optimiser based on random mini-batches, such as the commonly used Adam {\citep{kingma2014adam}}. 
However, it must be noted that the batch-wise loss is not a simple approximation of the full loss as, for instance, the normalised {occurrences} tend to have a larger value for smaller mini-batches. Nevertheless, we provide the mathematical guarantee that, for any mini-batch size $n$ ($1<n<K$), {minimizing} the model {loss} on all mini-batches also {minimizes} the full loss (see Appendix \ref{annex:batchglobal}). This suggests that our final estimator should be close to the global minimizer of the full loss, even though {there is not guarantee to} obtain the latter{ due to the non-convexity induced by the non-linear feature extractor}. Similarly to supervised contrastive methods~\citep{khosla2020supervised}, the intensity predictions could become more specific, and more concentrated around the occurrences, as the mini-batch size increases.
A small mini-batch size could, on the other hand, result in smoother species intensities. Therefore, we tested the impact of the mini-batch size during training on the final predictions of DeepMaxent. 
{Not that, g}iven the expression of $\lambda_j(\mathbf{x}_i)$ in Equation \ref{eq:lambda_j}, {the normalised intensities} across the mini-batch $B$ {used in Equation \ref{eq:batch_loss} are given by} $\exp(\gamma_{j}^\top g_{\theta}(\mathbf{x}_i))/\sum_{k\in B} \exp(\gamma_{j}^\top g_{\theta}(\mathbf{x}_k))$. This is a particular case of {the normalisation} function commonly referred to as softmax, applied to the logits $\gamma_{j}^\top g_{\theta}(\mathbf{x}_k)$ over the mini-batch $B$. 

\subsubsection{Spatial sampling bias correction with Target-Group Background correction}

When occurrence concentration is biased by spatial variations in sampling effort, a popular SDM correction approach is the Target-Group Background (TGB) method~\citep{phillips_sample_2009}, which was initially proposed to correct sampling bias in Maxent. 
The method basically approximates the spatial sampling effort through the distribution of occurrences of a Target-Group {(TG)} of species, providing background points to Maxent for each site where TG species were reported. In other words, TGB restricts the study domain to the sites with at least an evidence of sampling effort (one observation), which reduces the problem of false absences.
The strategy is expected to work when TG species are reported jointly with the focal species (e.g. the TG is a biological group targeted by {the same} citizen science program). 

In DeepMaxent, which models multiple species simultaneously, {the TGB strategy emerges implicitly by considering all samples within a mini-batch in the computation of the partition function. This makes the implementation of TGB particularly efficient, as the calculation of the intensity prediction for each species is recycled to be used for the TGB-enhanced partition function.}




\subsubsection{L2-regularisation implementation in DeepMaxent}






Maxent makes use of L1 penalisation, on the species weights $\gamma_j$ that model the relation between the features and the density prediction. The L1 term, known as LASSO penalty, encourages $\gamma_j$ to become sparse, thus selecting a subset of features. 
The L1 regularisation is important in Maxent due to a number of features that grows more than quadratically with the number of environmental variables. In DeepMaxent, the latent features are learnt to maximise prediction performances {and can be kept to a fixed dimensionality, removing the need for feature selection~\citep{goodfellow_deep_2016}}.
For DeepMaxent, we employ L2 regularisation, i.e. a penalty on the Euclidean norm of the $\gamma_j$ {and the rest of the model weights}, which encourages small but non-zero weights, which {tends to} induce a smoothing of the estimated species intensities.
{The intercept term $\mathbf{b}$, on the other hand, is not penalized, since it controls only the baseline level of the intensity function per species and does not affect the effective capacity of the model.
The total loss function thus becomes:}

\begin{equation}
\mathcal{L}_\text{total}(\tilde{\lambda}, y; \theta,\gamma) = \mathcal{L}_\mathcal{H}(\tilde{\lambda}, y; \theta,\gamma) + \frac{\tau}{2} (\|\theta\|_2^2+\| \gamma \|_2^2)
\end{equation}

where $\tau$ is the weight decay coefficient. This term penalises large weight values, encouraging the model to learn smaller weights without enforcing sparsity.

\subsection{Evaluation of model performance}
\subsubsection{Datasets}

For our experiments we used two openly {available datasets}: (i) one from the National Centre for Ecological Analysis and Synthesis (NCEAS)~\citep{elith_presence-only_2020} and (ii) GeoPlant~\citep{picek_geoplant_2025}.

\paragraph{The NCEAS dataset} This dataset includes 52,605 PO records (for SDM training) and presence-absence (PA, for SDM evaluation) data from six global regions: Australian Wet Tropics (AWT), Canada (CAN), New South Wales (NSW), New Zealand (NZ), South America (SA) and Switzerland (SWI)~\citep{elith_presence-only_2020}. Each regions is associated with a specific set of species, and sometimes from several biological groups (see Table \ref{tab:dataset}), with a total of 226 anonymous species. The dataset provides specific environmental variables {with specific spatial resolution} for each region, including climatic, soil or location variables (see more details in \cite{elith_presence-only_2020}). {The total area of all regions is of 13,607,500 km$^2$ (Table \ref{tab:dataset}), so that there is $0.004$ PO records per km$^2$ on average when pooling all species and regions. However, the spatial concentration of records is extremely heterogeneous across regions and taxonomic groups, varying from $0.0002$ records/km$^2$ in South America (SA) to $0.887$ records/km$^2$ in Switzerland (SWI), as computed from Table \ref{tab:dataset}}. 

Various SDM methods have been evaluated using this dataset~\citep{elith_novel_2006, phillips_sample_2009, zbinden_selection_2024, valavi_predictive_2022}, which allows for a direct comparison of DeepMaxent's performance {to many state-of-the-art SDM methods}.
\cite{phillips_sample_2009} studied spatial sampling biases and found that the PO data in certain regions (AWT, CAN, and SWI) contained high levels of such biases, {making this a good benchmark to assess model robustness}.

\begin{table}[htbp]
  \caption{The total number of species, the occurrence number in PO data and the total number of species presence in PA data for each region and biological group}
  \label{tab:dataset}
  \centering
  \begin{tabular}{lllllll}
    \toprule
        \textbf{Code} & \textbf{Location}& \textbf{Biological Group} & \textbf{Species number} & \multicolumn{2}{c}{\textbf{Occurrences number}} & \textbf{{Area (‘000 km2)}} \\
                      &                           &                         & & PO  & PA \\
\midrule
AWT & Australian wet tropics & bird & 20 & 3105 & 340 & 24 \\
AWT & Australian wet tropics  & plant & 20 & 701 & 102 & 24 \\
CAN & Ontario, Canada &  bird & 20 & 5063 & 14571 & 979.3 \\
NSW & New South Wales  &  bates & 10 & 187 & 570 & 76.2\\
NSW & New South Wales  &  birds & 7 & 1781 & 1839 & 76.2\\
NSW & New South Wales  &  plants & 29 & 680 & 5329 & 76.2\\
NSW & New South Wales  &  reptile & 8 & 675 & 1008 & 76.2\\
NZ & New Zealand & plant & 52 & 3088 & 19120 & 265.4\\
SA & South America &  plant & 30 & 2220 & 152 & 12223.2\\
SWI & Switzerland &  tree & 30 & 35105 & 10013 & 39.6 \\
\end{tabular}
\end{table}

\paragraph{The GeoPlant dataset} {Described by \cite{picek_geoplant_2025}, we use this dataset to evaluate DeepMaxent under a different data regime, with 100 times more PO records, much more concentrated in space, and 40 times more species covered, than NCEAS. Indeed, GeoPlant contains 5,079,797 presence-only observations of 9,709 plant species (i.e. almost half of Europe's flora) from 13 selected datasets of the Global Biodiversity Information Facility (GBIF, \href{www.gbif.org}{www.gbif.org}), to be used for model training. These records cover 38 European countries spanning a total area of about 5,914,500 km$^2$, with $0.859$ records per km$^2$ on average. This spatial concentration in Geoplant is thus 200 times higher than the average of NCEAS, and comparable to the one of the SWI region, but for an area 150 times larger.} Additionally, 88,987 presence-absence records from the European Vegetation Archive (EVA) were used for model evaluation.

{Within GeoPlant, two types of input data were considered in this work for training SDMs.} 
The first {configuration} (Bioclim-GeoPlant) used bioclimatic variables aggregated to a 10 km resolution, enabling direct comparison with MaxEnt. 
The second {configuration} (LandSat-GeoPlant) focused on applying {deep learning-based approaches} to multi-band time series of satellite {data. Only methods that include a neural network feature extractor were considered here. We used Landsat-based covariates at 30 }m resolution, derived from the seasonally aggregated and gap-filled GLAD analysis-ready dataset~\citep{potapov_landsat_2020}, and accessed via the EcoDataCube platform~\citep{witjes_ecodatacubeeu_2022}.
These data covered six spectral bands: Red (R), Green (G), magenta (B), Near Infrared (NIR), Shortwave Infrared 1 (SWIR1), and Shortwave Infrared 2 (SWIR2).

\subsubsection{Evaluation metrics}

To directly compare our results to \cite{phillips_sample_2009}, \cite{zbinden_selection_2024} and \cite{valavi_predictive_2022}, we evaluated our method performances with the Area Under the ROC Curve (AUC) computed for each species across the PA plots, {none of} which were used in model training. The AUC is the empirical probability that a presence site has a higher model-predicted value than an absence site. In other words, it measures the model ability to distinguish between presence and absence classes based on its predicted scores. The NCEAS dataset being decomposed into regions and biological groups, we first averaged each species-wise AUC per biological group and then per region (across groups), and then took the average over regions as our general performance metric.

\subsubsection{Implementation details}

\paragraph{NCEAS Dataset}

For the NCEAS dataset, the feature extractor $g_\theta$ was implemented as a multilayer perceptron (MLP) with {rectifier linear unit (ReLU) non-linearities and} skip connection between hidden layers (see Figure \ref{fig:architectures}) {in order to mitigate potential vanishing gradient issues when exploring deeper architectures~\citep{he_deep_2016}}. {The use of an MLP as a feature extractor was justified both by the structure of the input data (vectors of environmental variables for each site) and as a standard reference in deep learning-based multi-species SDMs
~\citep{zbinden_selection_2024,kellenberger_performance_2024,hu_introduction_2025}.}

{In this study, the parameters related to the neural network architecture and the optimiser, referred to as hyper-parameters, were selected through a cross-validation procedure. The search included the number of hidden layers, learning rate, mini-batch size and weight decay, and was performed using spatially blocked folds based on geographic data~\citep{valavi_blockcv_2019,roberts_cross-validation_2017}}. As suggested in \citet{zbinden_selection_2024}, the cross-validation was performed using PO data (see details in appendix~\ref{annex:cv}). {All models were trained for 100 epochs}. Once cross-validation has been performed, the {hyper-parameter values were chosen to be the same for all regions and biological-groups. These hyper-parameters were: 2 hidden layers, Adam as optimiser, a learning rate of $0.0002$, and a both the mini-batch size and hidden layer size of $250$. Each loss function was evaluated both with and without TGB correction. Among the hyper-parameters tested, these values consistently yielded the best results for all loss functions (see Annex \ref{annex:cv}). Regarding weight decay $\tau$, we did not observe any performance improvement across the tested loss functions, except in the case of DeepMaxent, where it led to better results.} The final model was then calibrated with these {hyper-parameter} values on the whole PO data and applied to the PA data. To account for variability arising from model initialization, each model was trained and evaluated across ten different random seeds, following the same procedure as used in~\citet{zbinden_selection_2024}. 


For the NCEAS dataset, {although DeepMaxent implicitly implements Target Group Background (TGB) correction, we additionally added random background points in locations with no species observations to verify that adding such points would not contribute to improve model performance. Background points were sampled uniformly from the raster data, producing a dataset ten times larger than the presence-only (PO) occurrences.} This procedure ensured a consistent representation of environmental conditions across the study area and maintained a percentage-based approach to guarantee comparability across datasets (e.g., SWI or CAN).

\paragraph{GeoPlant Datasets}
For the Bioclim-GeoPlant dataset, we used the same MLP architecture as for NCEAS. This choice was {again} driven by the structure of the data, {consisting of 19 bioclimatic variables}.
For the LandSat-GeoPlant dataset, each occurrence in our dataset was represented as a multidimensional data cube {of 6 spectral bands × 4 seasons × 21 years}, which serves as input to the model (see Figure~\ref{fig:landsatCUBE}). {As a feature extractor we used the adapted ResNet-18 model proposed in~\cite{picek_geoplant_2025}}. The output features from the ResNet-18 {were} processed through two fully connected layers, resulting in the final predictions. {The number of epochs was fixed at 20 throughout the training process and each hidden layer was composed of 250 neurons}.

For {both} GeoPlant datasets, cross-validation was not performed due to computational constraints. Instead, a validation set was created by randomly splitting the presence-only (PO) data to {select the best model}. This approach ensured that model performance {was} assessed without excessive computational overhead. The {hyper-parameters} used for training {were adapted} from \citet{picek_geoplant_2025}, maintaining consistency with previously established configurations.

\subsubsection{Baseline losses}

We implemented the Poisson regression loss (see Equation~\ref{eq:poisson_loss}){, using the same neural network architecture as in DeepMaxent to model $\lambda_j$, } to test the effect of the density normalisation in DeepMaxent on the estimator quality. 
Other commonly used loss functions in deep learning, and notably for SDMs, namely Cross-Entropy over species (CE, \cite{deneu_convolutional_2021,brun2024multispecies}) and Binary Cross-Entropy (BCE, \cite{benkendorf2020effects,zbinden_selection_2024}) were implemented. The Cross-Entropy loss (CE) {over species}, $\mathcal{L}_{\mathrm{CE}}(\boldsymbol{\Lambda}, \mathbf{Y})$ (Equation~\ref{eq:CE}), measures for each site the deviation between a predicted probability distribution across species and the associated empirical distribution based on the species observations in that site. In this case, the predicted probabilities are obtained by normalising the intensity values over the species $\boldsymbol{\Lambda} := \{\lambda_{ij}\}_{i=1,j=1}^{K,N}$, {at each site separately,} implemented {using the} softmax {function}:

\begin{equation}
\label{eq:CE}
\mathcal{L}_\mathrm{CE}(\boldsymbol{\Lambda}, \mathbf{Y}) = -\frac{1}{K} \sum_{i=1}^K \sum_{j=1}^N \frac{y_{ij}}{\sum_{k=1}^N y_{ik}} \log \left( \frac{\lambda_{ij}}{\sum_{k=1}^N \lambda_{ik}} \right)
\end{equation}

The Binary Cross-Entropy {(BCE)} loss, $\mathcal{L}_{\mathrm{BCE}}(\boldsymbol{\Lambda}, \mathbf{Y}_\mathbb{b})$, was implemented for the case where $\mathbf{Y}_\mathbb{b} \in \{0, 1\}$ is treated binary variable of $\mathbf{Y}$, taking the value 1 if the species was observed at least once in the pixel, and 0 otherwise. Unlike the CE case, where the probabilities across species are required to sum to 1 in each site, the predicted probability of a species learned using BCE does not directly restrict the probabilities of other species.
{In this setting, the softmax function reduces to a sigmoid function. }
The BCE loss is defined as:

\begin{equation}
    \mathcal{L}_\mathrm{BCE}(\boldsymbol{\Lambda}, \mathbf{Y}_\mathbb{b}) = -\frac{1}{K N} \sum_{i=1}^K \sum_{j=1}^N \left( y_{\mathbb{b},ij} \log \sigma(\log (\lambda_{ij})) + (1 - y_{\mathbb{b},ij}) \log (1 - \sigma(\log (\lambda_{ij}))) \right),
\end{equation}

where $\sigma$ is the logistic sigmoid function here applied to the linear predictor, or "logit", of each species $\log(\lambda_{ij})=\gamma_j^\top g_{\theta}(\mathbf{x}_i) + b_j$.

\section{Results}

\subsection{Comparative analysis of SDM methods}

Table \ref{tab:results} shows the performances of various {traditional SDM} methods, including Maxent, Boosted Regression Tree (BRT) with or without TGB correction, and the {recent} neural network model for multi-species proposed by \cite{zbinden_selection_2024}, all evaluated with the average AUC per region{, along with overall average}~\citep{phillips_maximum_2006,valavi_predictive_2022,zbinden_selection_2024}. It also contains the performance of our main DeepMaxent implementation and the baseline {deep learning} losses, with {and} without the TGB correction.

\begin{table}[htbp]
  \caption{Comparison of method performance by region-averaged AUC and general averaged AUC over all regions. The best average AUC for each column is highlighted in bold, while the second-best averaged AUC is underlined. The references correspond to results from the following articles: [1] \citet{valavi_predictive_2022}, [2] \citet{phillips_sample_2009} and [3] \citet{zbinden_selection_2024}. }
  \label{tab:results}
  \centering
  \begin{tabular}{lllccccc}
    \toprule
    & \multicolumn{6}{c}{Regions}                          \\
    \cmidrule(r){2-7}
                    & AWT & CAN & NSW & NZ & SA & SWI & avg \\
                        \midrule
        \multicolumn{8}{l}{\textbf{Results from the literature}} \\ 
    \multicolumn{8}{l}{\textit{Single-species models}} \\
    Maxent [1] & 0.686    & 0.587    & 0.700    & 0.738   & 0.804  & 0.809    & 0.721  \\
    BRT [1]    & 0.681    & 0.577    & 0.701    & 0.735   & 0.795 & 0.816    &  0.718 \\
    RF down-sampled [1] & 0.675 & 0.572 & 0.715 & 0.746& \underline{0.813}& 0.818 & 0.723\\
    Ensemble [1] & 0.683 & 0.580 & 0.710 & \underline{0.749} & {0.806} & 0.812 & 0.723\\ 
    IWLR-GAM [1] & 0.674 & 0.595 & 0.689 & 0.747& 0.796 & 0.798 & 0.716\\
    Maxent (using TGB) [2] & \textbf{0.732}    & 0.716    & 0.741    & 0.738   & 0.798   & 0.837   & 0.760  \\
    BRT (using TGB) [2]   & 0.700    & {0.728}    & 0.738    & {0.740}   & 0.792 & {0.842}   &  0.757  \\
     \multicolumn{8}{l}{\textit{Multi-species models}} \\
    Zbinden et al. [3] & 0.704 & 0.714 & 0.719 & 0.741 & \textbf{0.815} & 0.838 & 0.755 \\
    \midrule
    \multicolumn{8}{l}{\textbf{Results from our implementations}} \\
    \multicolumn{8}{l}{\textit{Baseline losses}} \\
    
    CE & 0.701 & 0.661 & 0.732 & 0.724 & 0.772 & 0.793  & 0.731 $\pm$ 0.001 \\
    CE (using TGB) & \underline{0.727} & 0.708 & 0.739 & 0.732 & 0.771 & 0.792 & 0.745 $\pm$ 0.001\\
    BCE & 0.656 & 0.600 & 0.718 & 0.736 & 0.804 & 0.799 & 0.719 $\pm$ 0.002\\
    BCE (using TGB) & {0.722} & \underline{0.730} & \underline{0.743} & 0.738 & 0.804 & \underline{0.849} & \underline{0.764} $\pm$ 0.002\\
    Poisson loss   & 0.658 & 0.599 & 0.714 & 0.737 & 0.804 & 0.799 & 0.719 $\pm$ 0.002 \\
    Poisson loss (using TGB)  &  {0.712} & \underline{0.730} & {0.732} & {0.729} & 0.801 & \underline{0.849} & 0.759 $\pm$ 0.002\\
    \multicolumn{8}{l}{\textit{Proposed loss}} \\
    DeepMaxent & 0.654 & 0.593 & 0.718 & 0.744 & 0.803 & 0.810 & 0.720 $\pm$ 0.001\\
    \textbf{DeepMaxent (using TGB)} & 0.712 & \textbf{0.732} & \textbf{0.752} &  \textbf{0.753} & 0.806 & \textbf{0.850} & \textbf{0.768} $\pm$ 0.001\\
    \bottomrule
  \end{tabular}
\end{table}

Without TGB sampling bias correction, performances are overall {lower} and close among methods, ranging from 0.716 to 0.723 in {overall} average AUC (Table \ref{tab:results}), except for the CE loss which {achieves} 0.731. Except for the latter, we observe no general performance gain for the tested deep learning losses (BCE, Poisson, DeepMaxent, ranging from 0.719 to 0.720) compared to the literature methods, e.g. Maxent (0.721) or the best SDM Ensemble of \cite{valavi_predictive_2022} (0.723). 

The TGB bias correction brings a consistent performance improvement for all methods, including the ones {in} the literature and our implementations. However, not all approaches respond equally strongly to {TGB}. For instance, Maxent gains 0.039 in {overall} averaged AUC by using TGB, and it is the same for BRT. Regarding our implemented baseline losses, TGB induces an AUC gain of 0.011 for the CE loss, 0.045 for the BCE loss and 0.040 for the Poisson loss. Finally, DeepMaxent achieves an improvement of 0.048 with TGB, resulting in the highest {overall} AUC (0.768).
These results show that the proposed DeepMaxent is well adapted to this bias correction technique while it enables to leverage the predictive potential of multi-species neural networks for spatial density estimation.
The largest region average AUC gains were mostly seen in regions CAN and AWT, where spatial sampling bias is the strongest according to~\citep{phillips_sample_2009}. Note that the best method of \cite{zbinden_selection_2024}, achieving an {overall} averaged AUC of 0.755, incorporated both random and TGB points as absences in their BCE loss, and their results specifically showed the key role of the TGB points in this performance.
DeepMaxent {with TGB} also had the best AUC in four of the six regions (CAN, NSW, NZ, SWI), showing that it is robust across regions and biological groups (NSW includes four biological groups, see Table \ref{tab:dataset}). 
BCE {with TGB} is the second-best method in {overall} AUC (0.764). In contrast, {using TGB,} CE and Poisson yield poorer {overall} AUC (0.745 and 0.759) than Maxent (0.760) or BRT (0.759).

{Indeed, although we would expect similar results for DeepMaxent and Poisson when using the same TGB strategy, given the equivalence of their global minimizer (see Appendix \ref{annex:equivalence}, DeepMaxent resulted in a 0.023 gain in AUC, closing the gap towards a perfect score by close to 10\% (i.e. 10\% less mis-ordered pairs of presence and absence sites) compared to Poisson, a significant gain according to statistical tests (Appendix \ref{annex:statistical_tests})}.

\begin{figure}[htbp]
    \centering
    \includegraphics[width=0.5\linewidth]{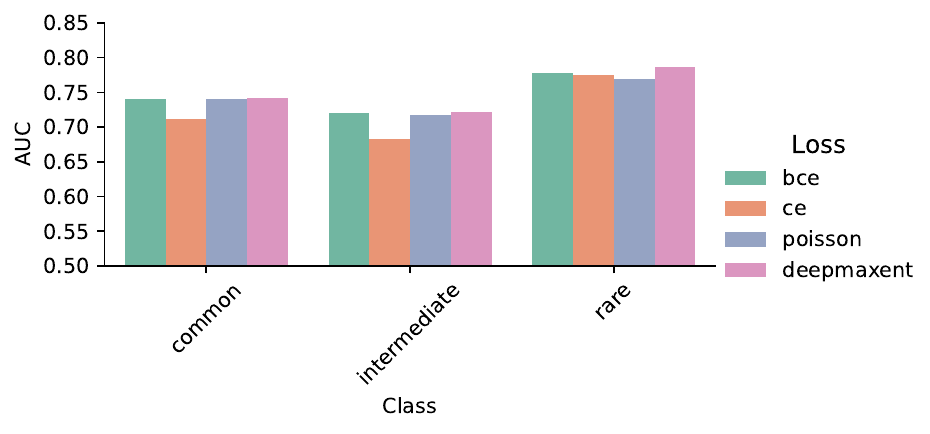}
    \caption{Comparison of average AUC values across all regions by loss and abundance classes on NCEAS dataset}
    \label{fig:auc_by_abundance}
\end{figure}

Figure~\ref{fig:auc_by_abundance} shows AUC scores by {observation} abundance class (rare, common and abundant) for each loss function, averaged over all regions. Averaging takes into account region-specific abundance class distributions (see appendix \ref{annex:distribution_biodivdata}).

DeepMaxent consistently outperforms the other losses across all abundance classes{, while} BCE ranks second in every class. 
{Poisson loss performs moderately well for abundant and common species. For rare species, the Poisson loss is dominated by the many zero-observation terms, which directly penalize the predicted intensities via the $\lambda_{ij}$ component. This tends to enforce uniformly low intensity values, potentially leading to underfitting beyond what data scarcity alone would justify.}
In contrast, CE shows the opposite trend: it achieves its best results for rare species, but lower AUC scores for abundant and common species, suggesting that its sensitivity to class imbalance limits its overall effectiveness in these abundance regimes. 
{These results suggest that DeepMaxent is particularly well-suited for modelling species with heterogeneous distributions and low occurrence counts, where classical Poisson-based methods may underperform due to their tendency to over-penalize predictions in data-sparse regions.}

\begin{figure}[htbp]

    \centering
        \begin{subfigure}[b]{0.40\linewidth}
        \centering\large
        \includegraphics[width=\linewidth]{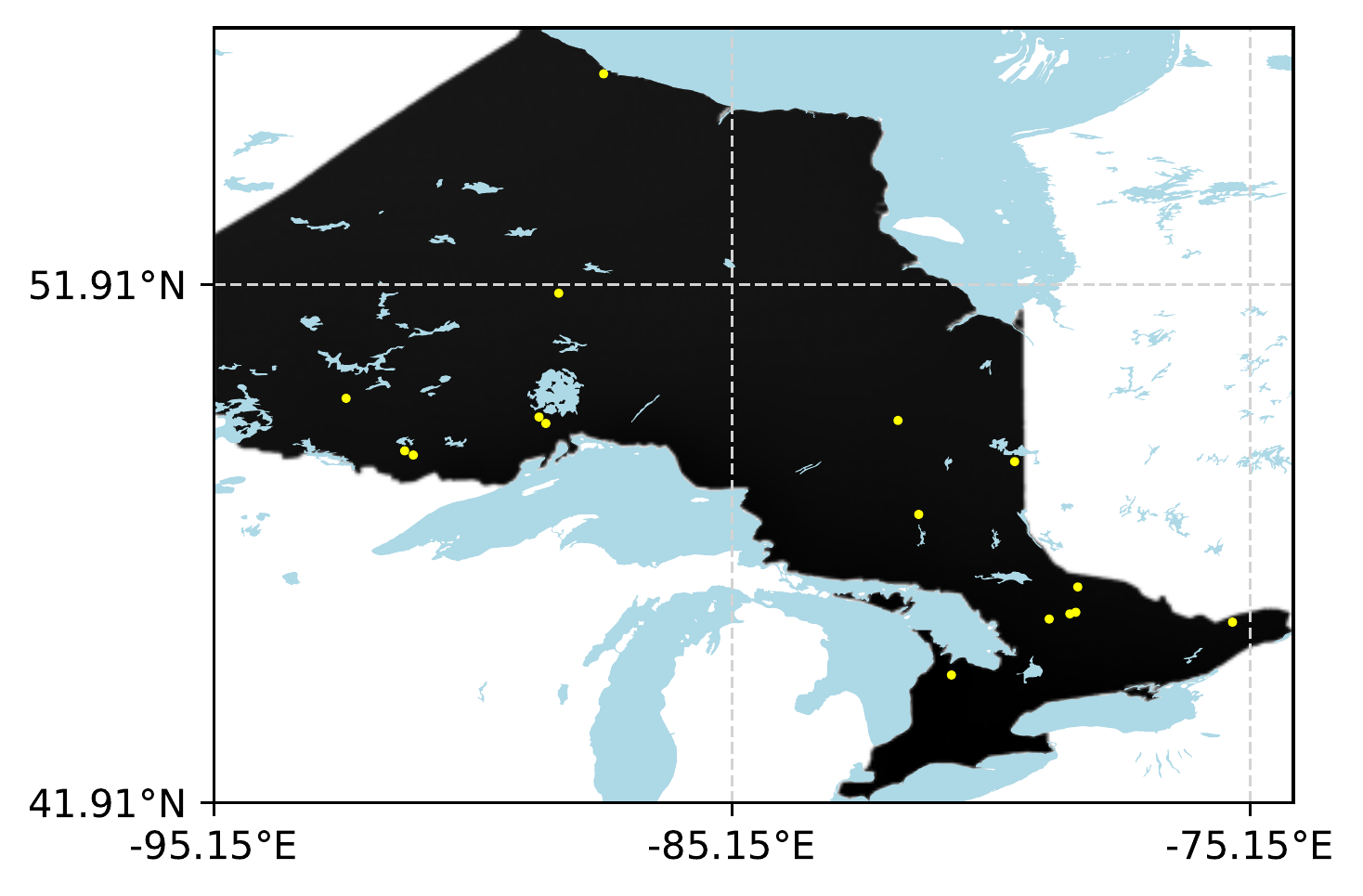}
      
        \caption{PO data for can01}
    \end{subfigure}
      \begin{subfigure}[b]{0.40\linewidth}
        \centering
        \includegraphics[width=\linewidth]{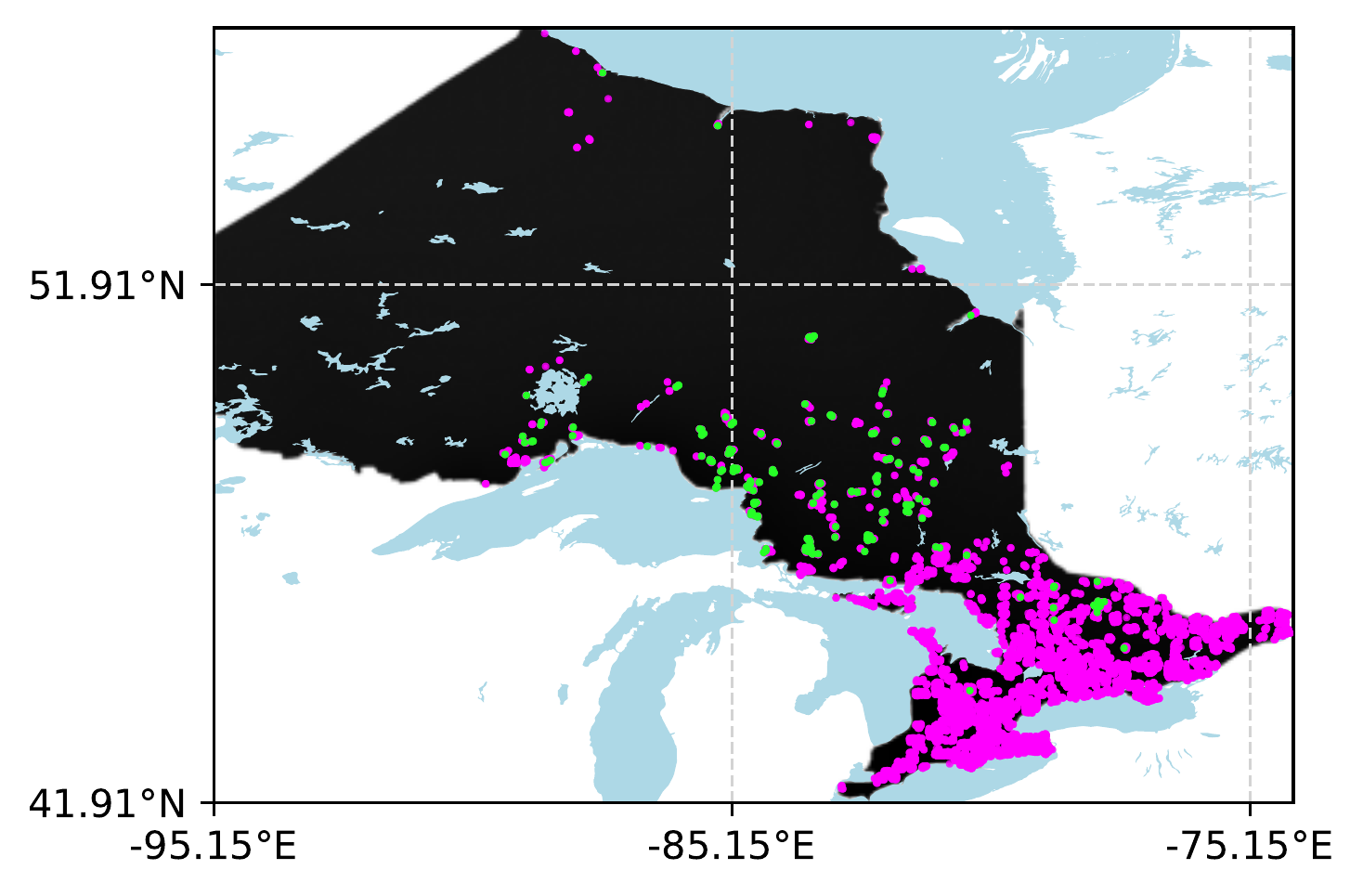}
       
        \caption{PA data for can01}
    \end{subfigure}
    \\
        \begin{subfigure}[b]{0.45\linewidth}
        \centering
        \includegraphics[width=\linewidth]{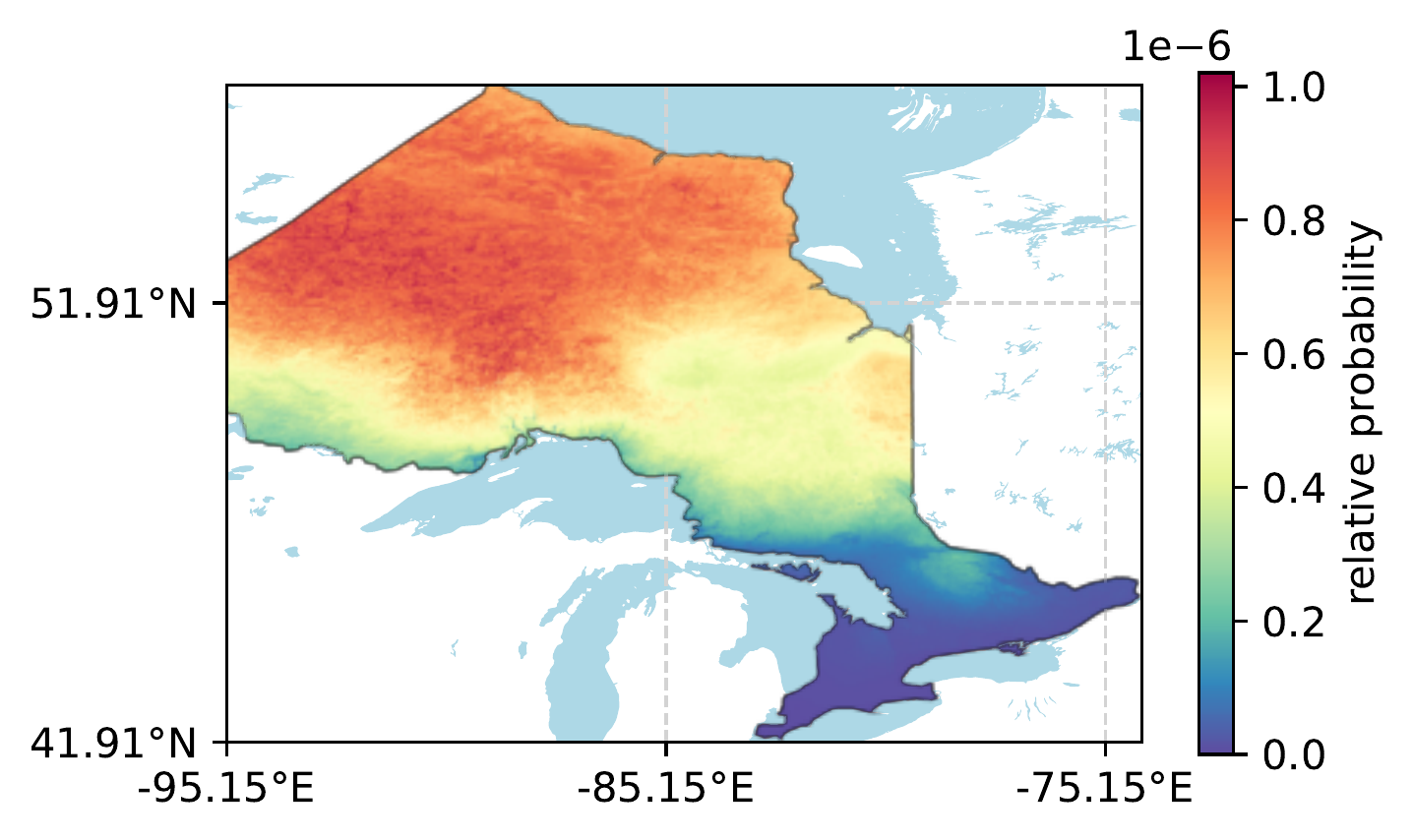}
     
        \caption{DeepMaxent - AUC of 0.942}
    \end{subfigure}
    \begin{subfigure}[b]{0.45\linewidth}
        \centering
        \includegraphics[width=\linewidth]{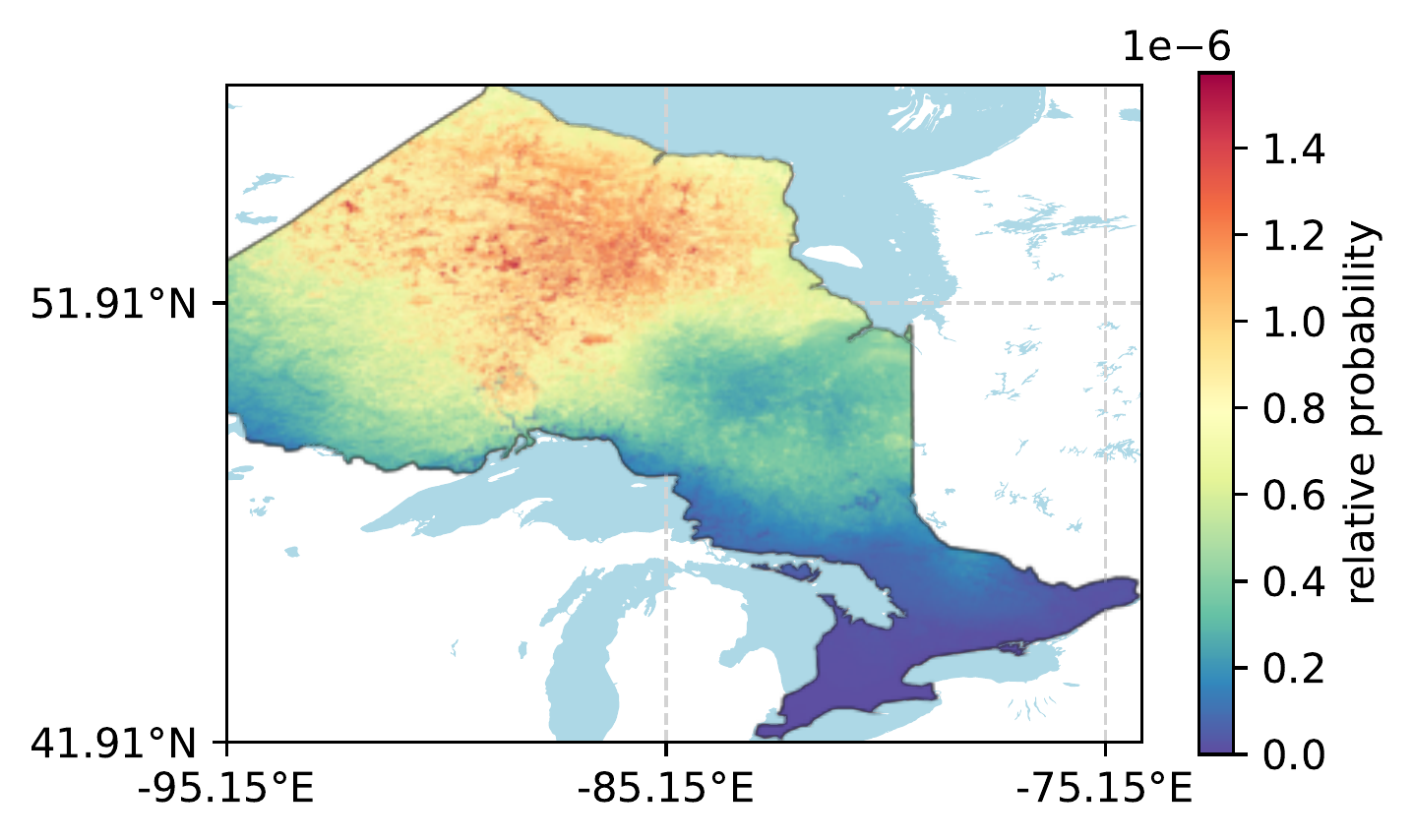}

        \caption{BCE - AUC of 0.926}
    \end{subfigure}
    \\
    \begin{subfigure}[b]{0.45\linewidth}
        \centering
        \includegraphics[width=\linewidth]{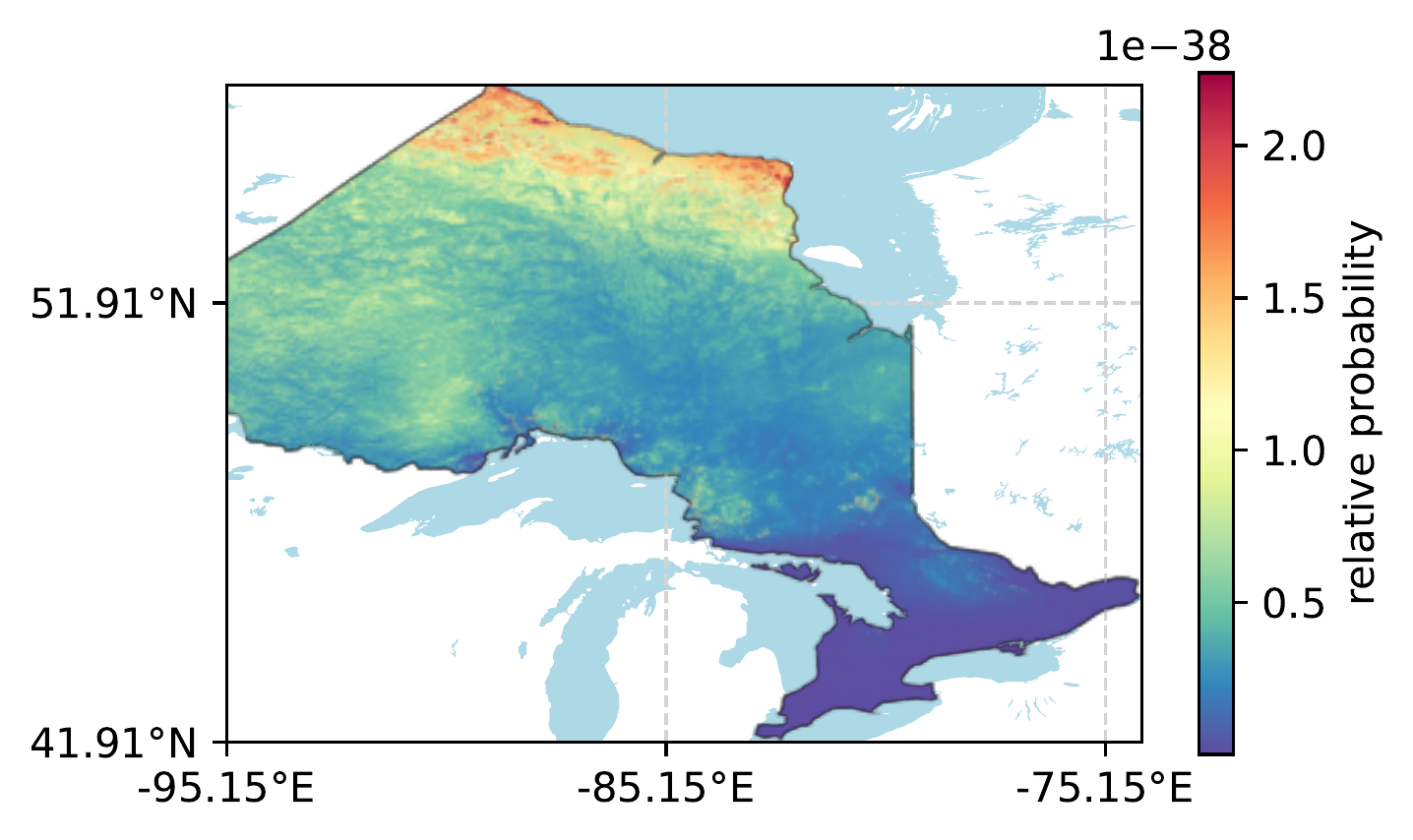}

        \caption{CE - AUC of 0.932}
    \end{subfigure}
        \begin{subfigure}[b]{0.45\linewidth}
        \centering
        \includegraphics[width=\linewidth]{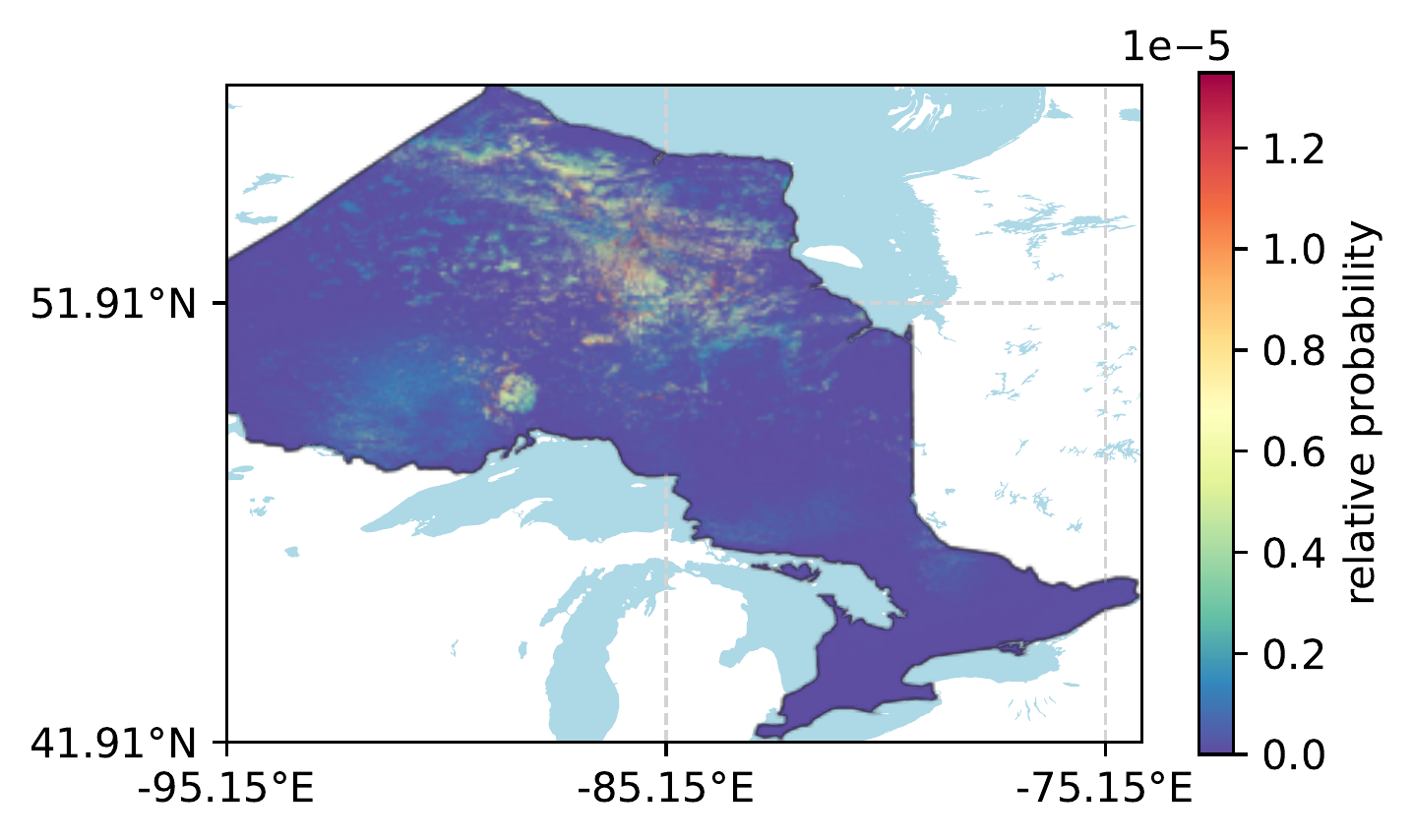}

        \caption{Poisson - AUC of 0.923}
    \end{subfigure}

    \caption{Estimated relative probabilities for the species can01 (a rare species  {with 16 PO points}):
(a) Presence-Only (PO) data,  {yellow points};
(b) Presence-Absence (PA) data  {where green corresponds to presences, and magenta to absences.}
(c–f) Estimated from different loss functions: (c) DeepMaxent; (d) Binary Cross-Entropy (BCE); (e) Cross-Entropy (CE); (f) Poisson loss.
    }
    \label{fig:allmaps_comparative}
\end{figure}

Estimated probabilities for can01, a rare species, differ widely depending on the loss function applied (see Fig. \ref{fig:allmaps_comparative}). For this species, presence-only (PO) data are rare, while presence-absence (PA) data are more abundant. In contrast, the maps generated for can02, a common species, give much more consistent estimates between the different methods (see appendix \ref{fig:allmaps_comparative_common_species}).

Table \ref{tab:results_geoplant} reports the general averaged AUC values for different loss functions evaluated on two datasets, Bioclim-GeoPlant and Landsat-GeoPlant, using TGB sampling bias correction. 

\begin{table}[htbp]
  \caption{Comparison of method performance by general averaged AUC. The best average AUC for each column is highlighted in bold, while the second-best averaged AUC is underlined. {All methods use TGB for the background samples.}}
  \label{tab:results_geoplant}
  \centering
  \begin{tabular}{lc}
    \toprule
    Loss & AUC \\
    \midrule
    \multicolumn{2}{l}{\textbf{Results for Bioclim-GeoPlant}} \\ 
    Maxent   & 0.823  \\
    CE  & 0.830 $\pm$ 0.001\\
     BCE  & \underline{0.839} $\pm$ 0.001\\
     Poisson loss   &0.837 $\pm$ 0.002\\
     DeepMaxent  & \textbf{0.860} $\pm$ 0.001\\ 
    \midrule
    \multicolumn{2}{l}{\textbf{Results for Landsat-GeoPlant}} \\
    CE  & 0.829 $\pm$ 0.001\\
     BCE  & \underline{0.885} $\pm$ 0.002\\
     Poisson loss   & 0.862 $\pm$ 0.002 \\
     DeepMaxent  & \textbf{0.887} $\pm$ 0.001 \\ 
    \bottomrule
  \end{tabular}
\end{table}

On the Bioclim-GeoPlant dataset, the DeepMaxent model achieves the highest average AUC (0.860), outperforming all losses, including BCE (0.839), Poisson (0.837), and CE (0.830), as well as the Maxent model (0.823). These results highlight the improved predictive capacity of DeepMaxent when using environmental variables from Bioclim.
A similar result is observed on the Landsat-GeoPlant dataset, where DeepMaxent also performs best with a general averaged AUC of 0.887, closely followed by BCE (0.885), while CE and Poisson yield lower performances (0.829 and 0.862, respectively). DeepMaxent performs well on satellite time series data. BCE Loss also performs well in this context, especially for a large dataset. {Notably, the standard deviations across random initializations are very small for all models ($\leq$0.002), indicating high consistency across runs. }

\subsection{Sensitivity study}

\begin{figure}[htbp]
    \centering
    \begin{tabular}{ccc}
      &   Smaller value & Higher value\\
    \toprule
    \textbf{can01} \\
     &  \multicolumn{2}{c}{Batch size}  \\ 

    &    \begin{subfigure}[b]{0.4\linewidth}
        \centering
        \includegraphics[width=\linewidth]{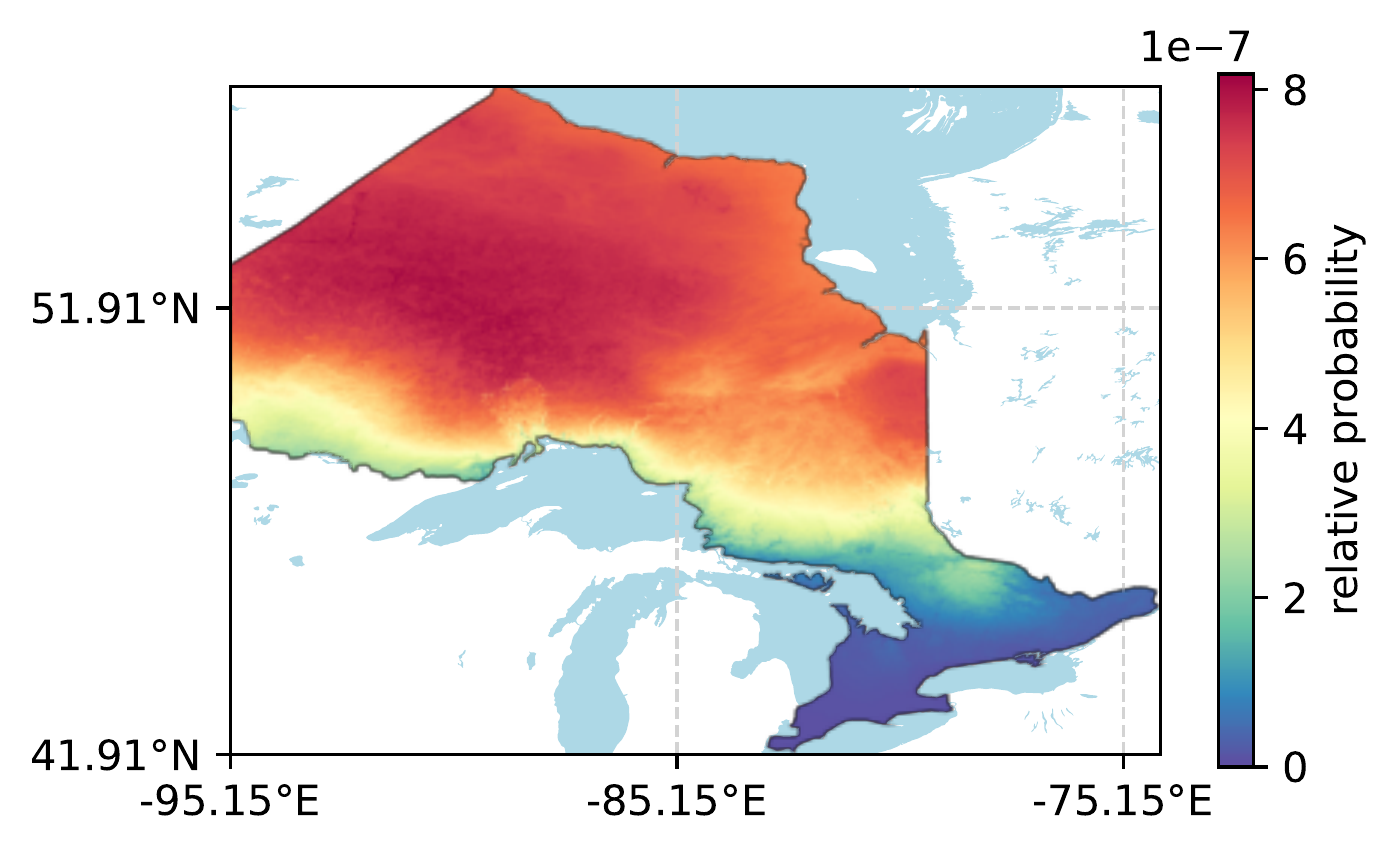}
        
    \end{subfigure}
    &    \begin{subfigure}[b]{0.4\linewidth}
        \centering
        \includegraphics[width=\linewidth]{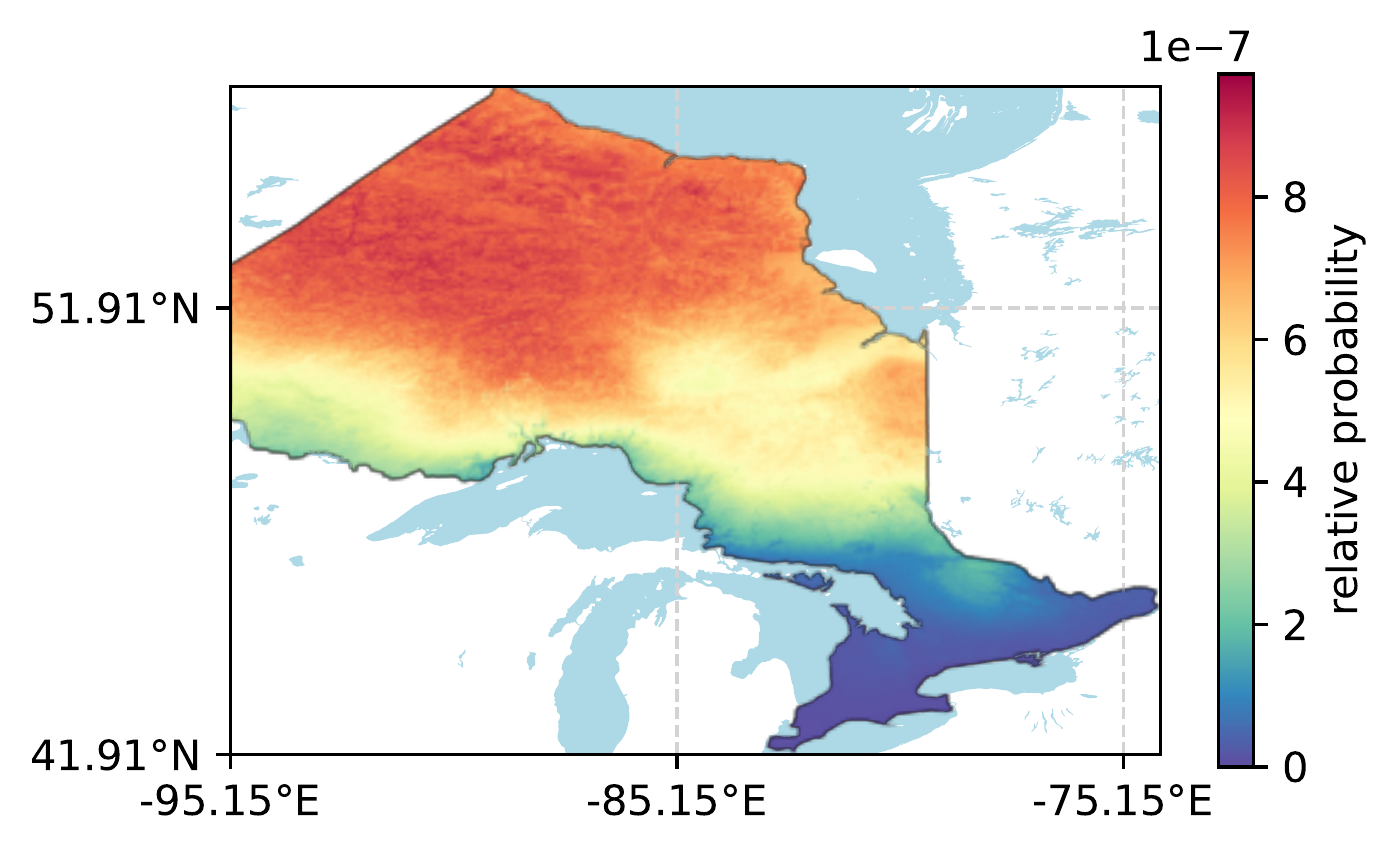}
        
    \end{subfigure}
     \\
    & \multicolumn{2}{c}{Weight decay}  \\ 
    & 
    \begin{subfigure}[b]{0.4\linewidth}
        \centering
        \includegraphics[width=\linewidth]{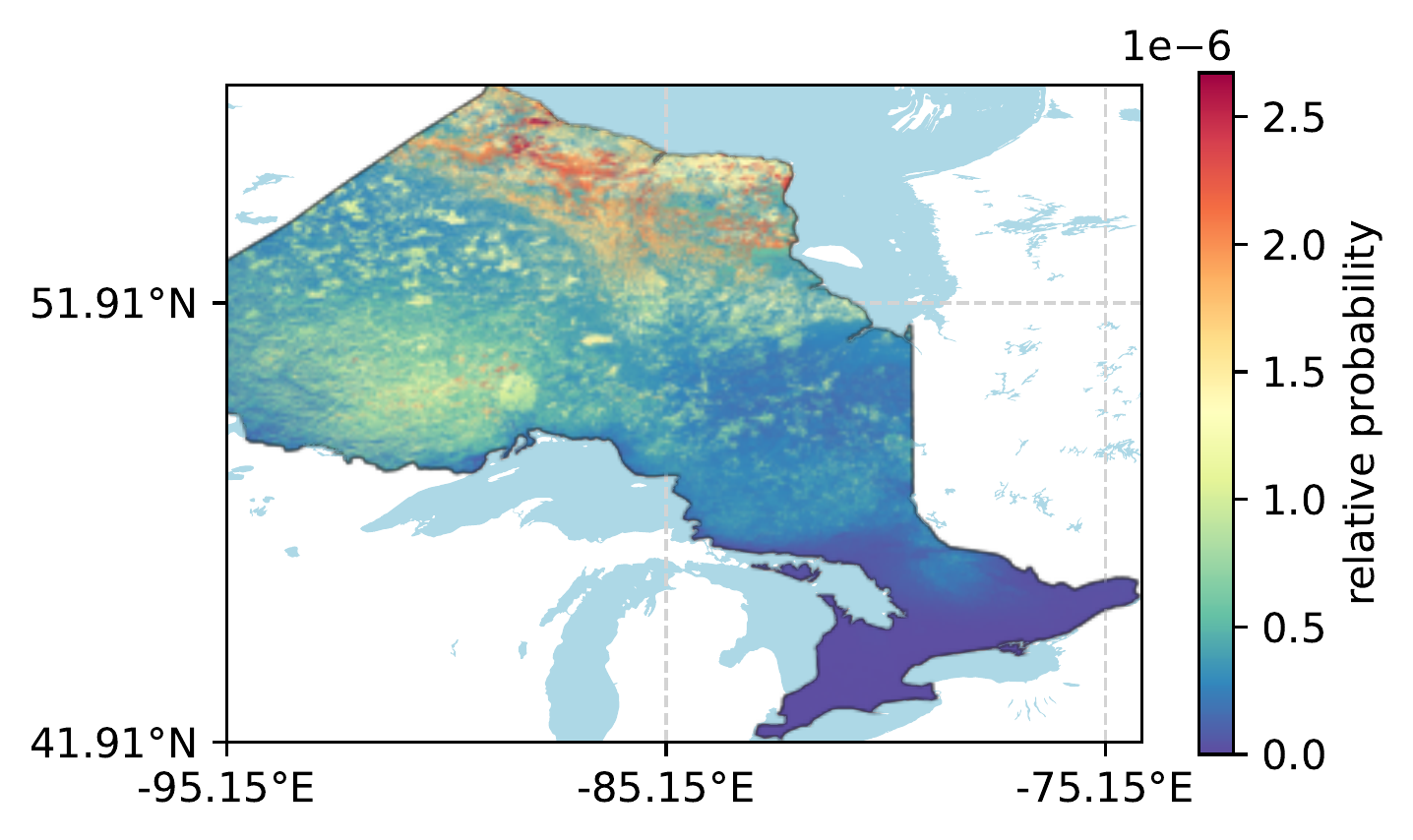}
        
    \end{subfigure}
    & 
        \begin{subfigure}[b]{0.4\linewidth}
        \centering
        \includegraphics[width=\linewidth]{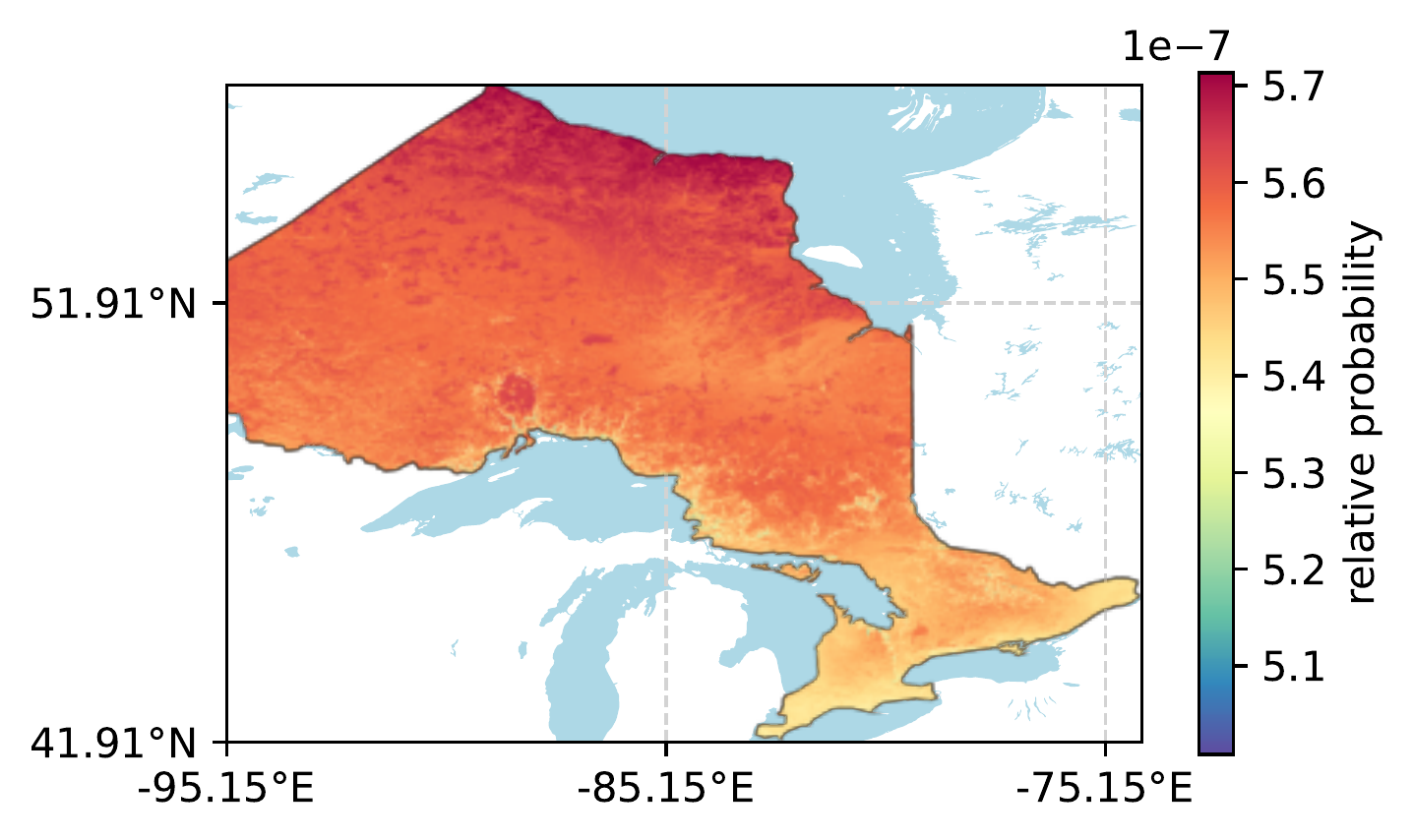}
        
    \end{subfigure}
    \\ 
    \midrule

\textbf{can02} \\
       &  \multicolumn{2}{c}{Batch size}  \\ 

    &    \begin{subfigure}[b]{0.4\linewidth}
        \centering
        \includegraphics[width=\linewidth]{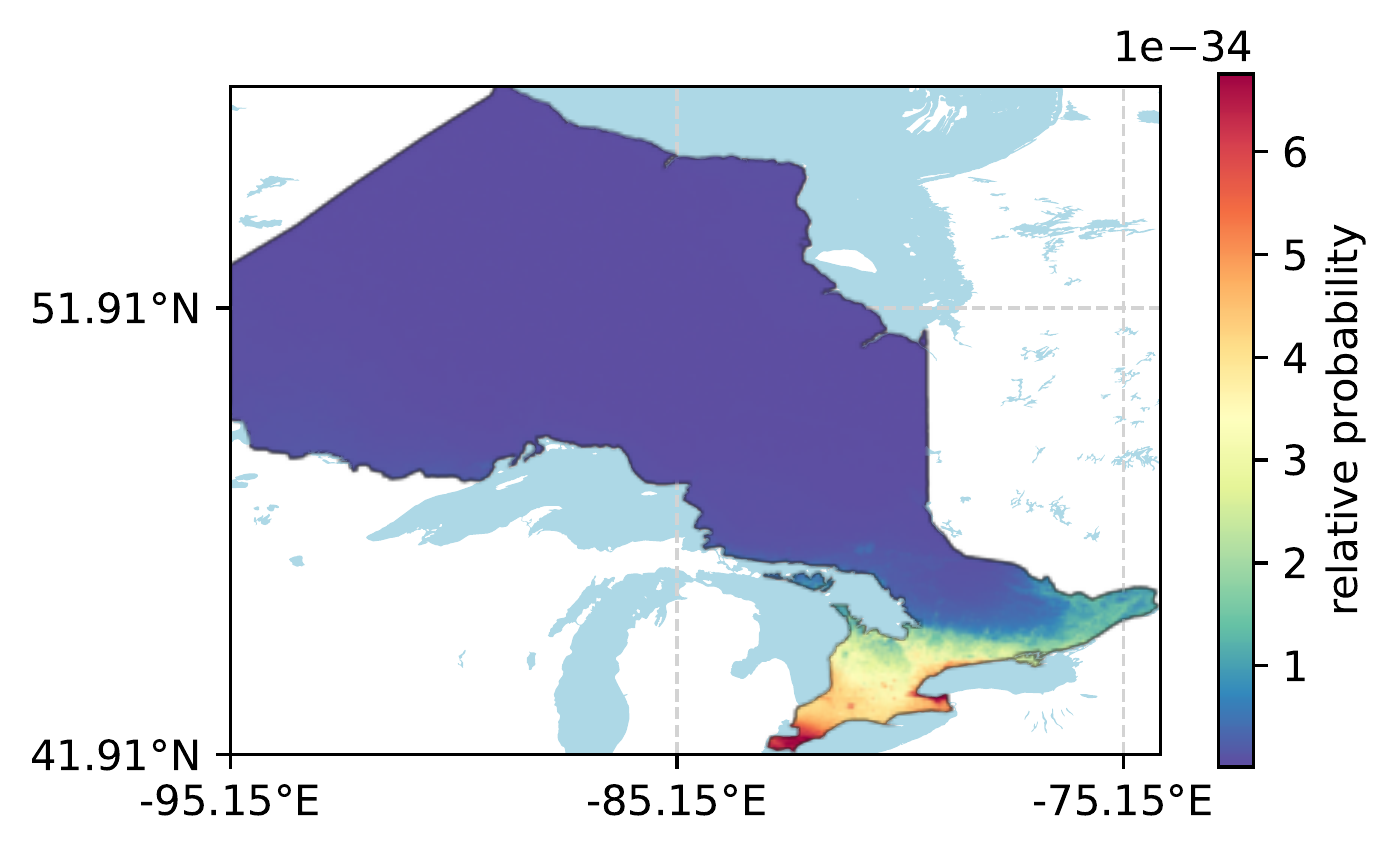}
        
    \end{subfigure}
    &    \begin{subfigure}[b]{0.4\linewidth}
        \centering
        \includegraphics[width=\linewidth]{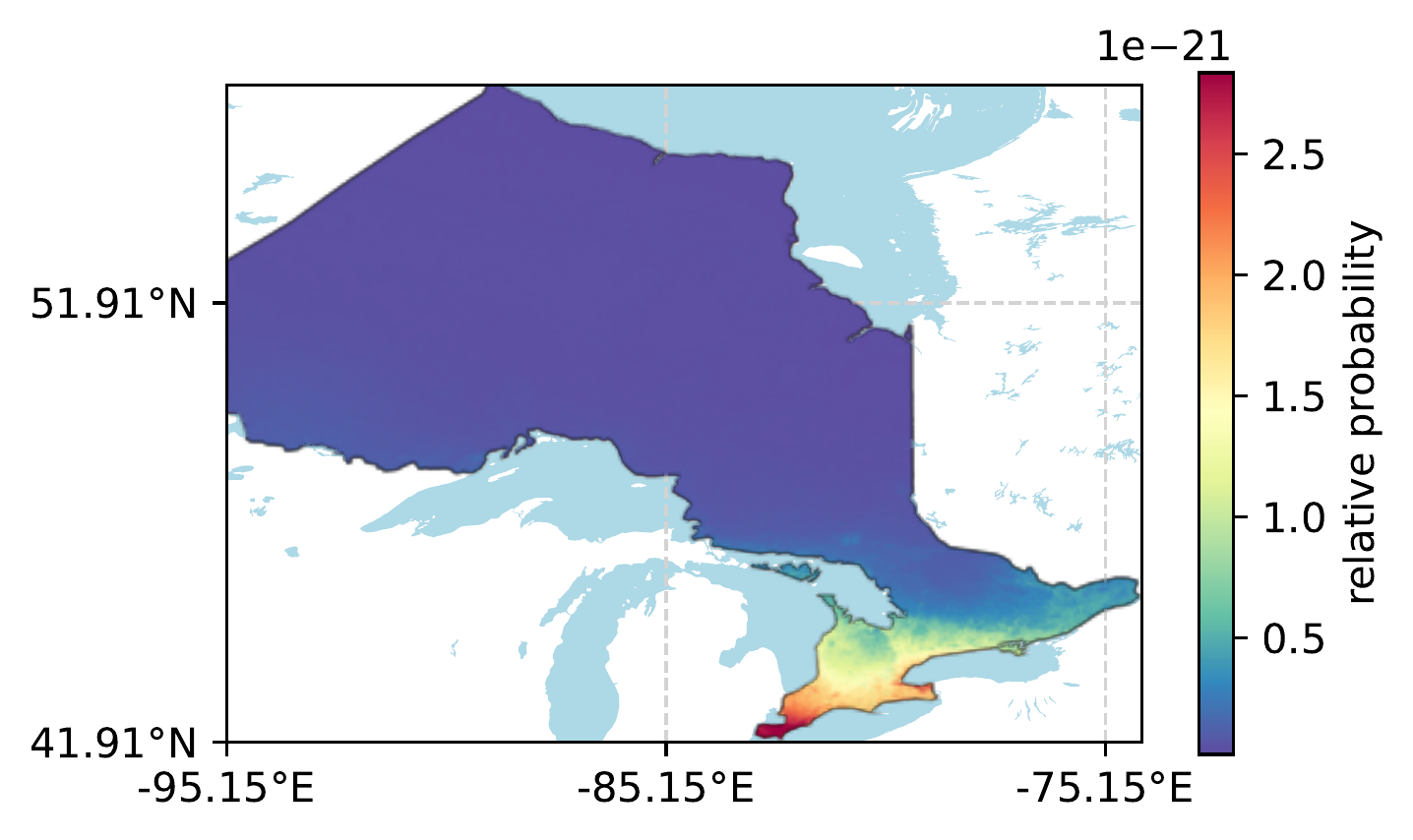}
        
    \end{subfigure}
     \\
    & \multicolumn{2}{c}{Weight decay}  \\ 
    & 
    \begin{subfigure}[b]{0.4\linewidth}
        \centering
        \includegraphics[width=\linewidth]{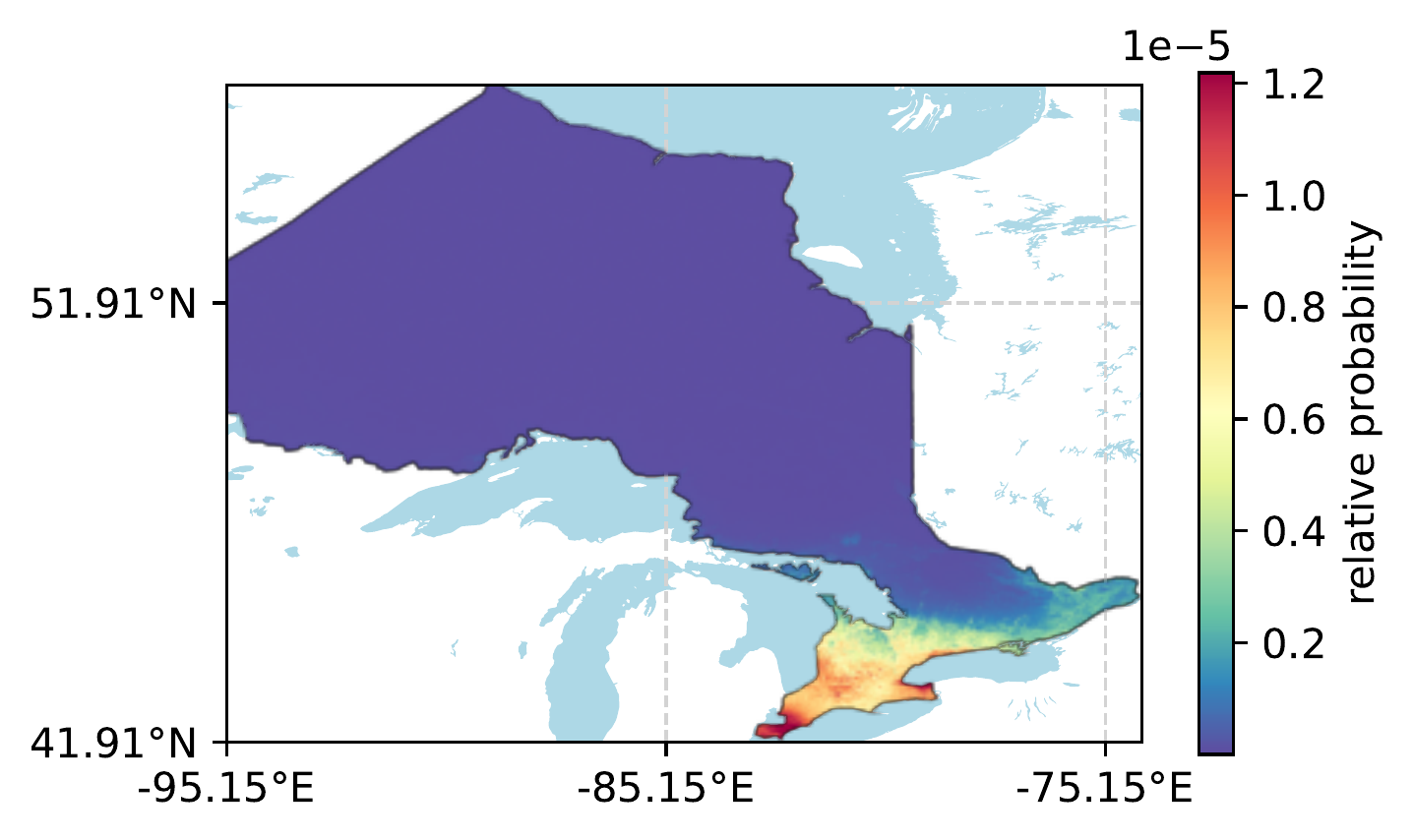}
        
    \end{subfigure}
    & 
        \begin{subfigure}[b]{0.4\linewidth}
        \centering
        \includegraphics[width=\linewidth]{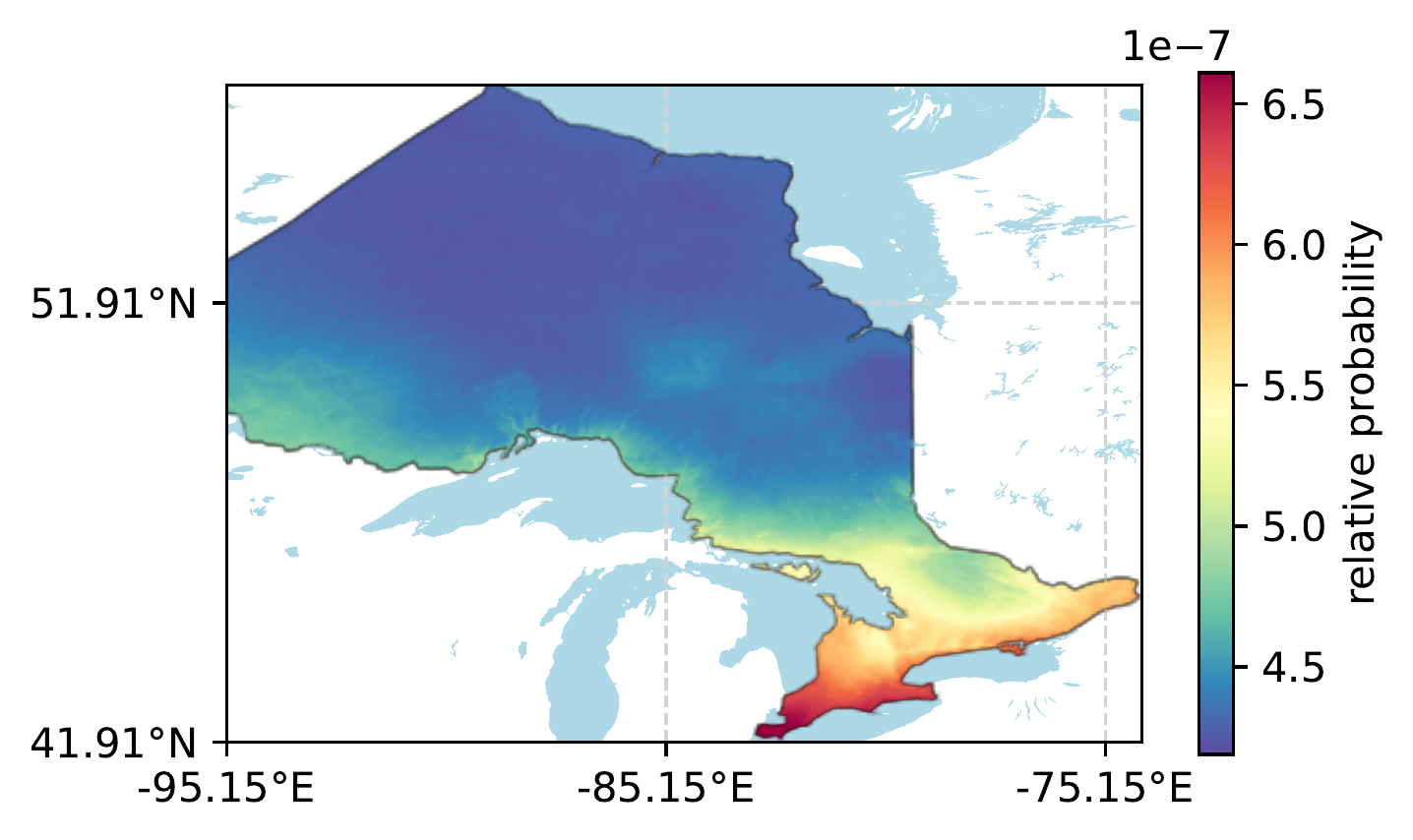}
        
    \end{subfigure}
    \\ 
    
    \end{tabular}

    \caption{Estimated relative probabilities for the species can01 and can02 (CAN) by varying mini-batch size and weight decay, while keeping other {hyper-}parameters at their default values (mini-batch size = 250, hidden layers = 2, weight decay = $3 \times 10^{-4}$). 
    {Smaller mini-batch sizes correspond to 10, and larger sizes to 3000. For weight decay, smaller values correspond to $3 \times 10^{-6}$, and larger values to 0.1.}
    }
    \label{fig:allmaps}
\end{figure}

Table \ref{tab:sensitivity_study} shows the average AUCs calculated for all regions, according to six different values for each {hyper-parameter}: mini-batch size, number of hidden layers and weight decay. A detailed analysis of AUC values for each region is provided in the appendix (\ref{annex:detailed_sensitivity}). 
The general performance of DeepMaxent-TGB was quite robust to hyper-parameter choices{, with the largest difference in average AUC across all tested values being only 0.010, illustrating the model’s stability with respect to mini-batch size, number of hidden layers, and weight decay.} In particular, DeepMaxent {with TGB} kept a general average AUC above 0.764, i.e. above the best results using all other methods, for all tested mini-batch sizes, ranging from 10 to 2500. Qualitatively, a smaller mini-batch size induces smoother species intensity maps, while larger mini-batch size tends to concentrate the intensity in higher abundance areas, as illustrated for one species in the region CAN in Figure~\ref{fig:allmaps}. The L2 regularisation (weight decay) has an {important} impact on DeepMaxent {performance}. {It} has a small but consistently positive impact on the performance up to a value of $3\times10^{-4}$ (see Table \ref{tab:sensitivity_study}), while further increasing the weight decay value results on a performance degradation due to oversmoothing (see Figure~\ref{fig:allmaps}). 
Varying the number of hidden layers in the neural network architecture of DeepMaxent from one to two had almost no effect, with a same general averaged AUC of 0.767 (Table \ref{tab:sensitivity_study}), and the AUC softly and progressively decreased for three (0.766), four (0.764), five (0.762) and six layers (0.759). 

\begin{table}[htbp]
    \centering
    \caption{Average AUC values for DeepMaxent-TGB across all regions, calculated for six different values of each {hyper-}parameter: mini-batch size, number of hidden layers, and weight decay. The default values used are a mini-batch size of 250, two hidden layers, and a weight decay of 3e-4.}
    \begin{tabular}{lc|lc|lc}
         Mini-batch size & AUC & Hidden layers & AUC & Weight decay & AUC
         \\ \hline
          10 & 0.765     & 1& 0.767 &0 &0.762 \\
          25&  0.765     & 2 & 0.767 & 3e-5& 0.763\\
          100& 0.767     &3 &    0.766 & 1e-4& 0.765\\
          250&  0.767    &4 &    0.764 &3e-4& 0.767\\
          1000&  0.766   &5 &   0.762 & 1e-3& 0.765\\
          2500&  0.764   &6 &    0.759 & 3e-3 &0.757\\
    \end{tabular}
    \label{tab:sensitivity_study}
\end{table}

\section{Discussion}




{In this study, we propose DeepMaxent, a new SDM method for PO data that generalizes Maxent to multi-species deep neural networks with a direct relationship to a Poisson count loss. DeepMaxent can be trained with a scalable batched algorithm, and implicitly implements the Target-Group Background sampling bias correction (TGB) initially proposed for Maxent~\citep{phillips_sample_2009}}. We {conduct an extensive evaluation} of the method on two SDM benchmark datasets: NCEAS~\citep{elith_presence-only_2020} and GeoPlant~\citep{picek_geoplant_2025}, {which together span a wide range of species, biological groups, regions, and data regimes}. {The} NCEAS {dataset enables} comparison {with a broad set of }state-of-the-art SDM methods previously evaluated on this dataset~\citep{phillips_sample_2009,valavi_predictive_2022,zbinden_selection_2024}, such as Maxent and BRT. {We also use NCEAS to compare the DeepMaxent loss against batched implementations of alternative losses (Poisson, BCE, CE), and to assess the efficiency of the implicit TGB bias correction compared to uniform random background for each loss}. 
{Consistent} with {previous} findings on traditional SDM methods~\citep{phillips_sample_2009, ranc2017performance, barber2022target}, deep learning-based SDMs benefit strongly and {systematically} from TGB regardless of the loss function. DeepMaxent {is well adapted to the TGB correction as it outperforms the same model architecture trained using the three alternative losses: the Poisson loss from which it derives, the cross-entropy over species~\citep{deneu_convolutional_2021,brun2024multispecies}) and the binary-cross-entropy loss (encoding occurrences as presence and pseudo-absences\citep{benkendorf2020effects,zbinden_selection_2024})}. DeepMaxent with TGB achieve{s} the highest average AUC across all NCEAS regions and outperform{s} all other methods in four of the six regions (see Table \ref{tab:results}). It notably surpasse{s} the single-species Maxent and BRT methods with TGB correction.

{More broadly, our results illustrate that, like other SDM methods, deep learning-based SDMs must properly account for spatial sampling biases to reveal their performance potential. This can be achieved through the compatibility of DeepMaxent and TGB, but many other proposed bias correction methods could bring further improvements to DeepMaxent (e.g. \cite{boria_spatial_2014,botella2021jointly}).} 
Future efforts could consider more sophisticated bias correction strategies, such as adapting background sampling strategies to biases specific to each biological group and species, as these biases are often a major source of variation in the performance of species distribution models~\citep{schartel_background_2024}, or explicitly model spatial sampling effort~\citep{warton2013model,botella2021jointly}.

For feature extraction on the NCEAS dataset, DeepMaxent with a fully-connected neural network with just two hidden layers is found to be optimal across most regions. This simple MLP architecture performs well on tasks involving low-dimensional and unstructured tabular data. While deep learning is often associated with large models, simpler architectures can be more efficient and beneficial for such tasks. 
{For more complex tasks such as the GeoPlant dataset, DeepMaxent is compatible with any type of neural network architecture, allowing the model to obtain the best overall results on Landsat-GeoPlant using a convolutional architecture for time series}.

{In multi-species settings, DeepMaxent,} like other deep learning–based SDMs (and some specific approaches such as concurrent ordination~\citep{van2023concurrent}), reduces overall computational cost by learning a single feature extractor shared across species~\citep{ba_deep_2014,raghu_expressive_2017}, in contrast to most methods that rely on more resource-intensive, per-species computations~\citep{merow_what_2014}.
{Similar ideas of reducing multi-species complexity into shared latent dimensions exist in statistical ecology (e.g. concurrent ordination~\citep{van2023concurrent} and community-level driver models~\citep{ovaskainen2017species}), but DeepMaxent implements it through deep neural feature learning rather than explicit latent-variable modelling}. 
It is important to note that each layer added increases the number of model parameters. Although these parameters are shared between the different species, this increase in complexity can lead to overfitting, particularly when data are limited or noisy. A prudent approach, therefore, is to start with simple architectures and then gradually make the model more complex, by adding hidden layers, as long as validation performance continues to improve. A deterioration in validation performance then serves as a warning signal that the model is becoming too complex in relation to the amount of information available in the training data.

More broadly, the proposed approach is flexible regarding the type of input and species observation data and should facilitate data integration approaches in the future by using neural networks. In the case of more structured input data, the neural network architecture could {also} be adapted to ingest {other} types of inputs (e.g. spatial remote sensing imagery), which might lead to capturing complementary spatio-temporal environmental patterns~\citep{deneu_convolutional_2021,estopinan2022deep}. Importantly, DeepMaxent is the first method to bridge the gap between deep learning and point process-based SDM~\citep{renner_point_2015}. 
{Using a point-process formulation of species distributions within our loss framework, the present work focuses on presence-only data through a Poisson likelihood. However, this formulation is general and could, in principle, be extended to incorporate other types of ecological observations, such as presence–absence surveys~\citep{fithian_bias_2015}, detection/non-detection histories~\citep{koshkina2017integrated}, or abundance and imperfect count data~\citep{dorazio2014accounting}. In such extensions, all observation types would remain linked to the same underlying predicted ecological intensity, while each type would be associated with its own appropriate observation likelihood. When these likelihoods involve additional parameters (e.g., detection probabilities), they could be jointly estimated within the same optimisation framework or through hierarchical extensions under suitable independence assumptions (see, e.g., \cite{isaac2020data})}.

These approaches for combining various {observation} types, called integrated SDMs, have recently been highlighted as a promising avenue to enhance the reliability of SDMs~\citep{miller2019recent,isaac2020data,mostert2023pointedsdms}. Extending DeepMaxent with this approach could use standardised datasets to disentangle the real relative abundance of each species from detection biases, while harnessing the extensive geographic coverage of opportunistic presence-only data. To further increase ecological realism, the DeepMaxent framework could be extended to incorporate species co-occurrence patterns, following the principles of Joint Species Distribution Models~\citep{pollock2014understanding}. This integration, which involves modelling the joint probability of species occurrences given the environment, has already proven feasible within deep learning-based SDM architectures~\citep{chen2017deep}.

Maxent and log-linear Poisson regression have been shown to be equivalent in terms of the estimated probability across sites~\citep{renner2013equivalence}, and we found a similar equivalence between the more general DeepMaxent and Poisson losses, in that they share the same global minimizer when applied to a single species (Appendix \ref{annex:equivalence}). This would lead us to expect similar results for DeepMaxent and Poisson, but DeepMaxent consistently outperformed Poisson in our experiments. One key difference that could explain the discrepancy lies in the treatment of the absolute intensities in the multi-species setting: in the un-normalised Poisson loss, the model will try to capture differences in observation counts between species, leading to more importance in the loss for frequently observed species. In contrast, the DeepMaxent loss normalises intensities across sites for each species, which removes the effect of the number of observations in the per-species loss magnitude and decouples parameter updates from absolute intensity. 
Additionally, the stochastic mini-batch optimisation interacts with this normalisation, further differentiating the parameter estimates between the two losses.

Even though the CE loss has the best results without TGB correction, and thus appears natively less sensitive to spatial sampling bias, its performances with TGB remain below the rest of methods, which benefit more from this correction (DeepMaxent, BCE, Poisson, Maxent). 
This limited performance may be due to the loss of information on the spatial variations of the intensity for each species when normalising the intensity across species for each site. 
More broadly, regarding the estimation of multiple species spatial intensities from biased presence-only data, these results suggest that learning to classify the most likely observed species~\citep{deneu_convolutional_2021,estopinan2022deep,brun2024multispecies} per site leads to sub-optimal results.

In addition to the smoothing of species spatial densities achieved by the DeepMaxent loss function, which induces entropy maximization, this study highlights that L2 regularisation can be used to further encourage smoothing {of predicted probabilities} (see Figure~\ref{fig:allmaps}). By penalizing large model parameters, L2 reduces overfitting and produces smoother predictions, resulting in more gradual changes in species probability across space and improved performance in presence-only settings, potentially alleviating overfitting in the low-data regime.
{Concerning the optimisation process of DeepMaxent, w}e provided a mathematical guarantee to justify the use of a stochastic batched gradient descent algorithm and we further showed that varying the batch size (from 10 to 2500) had {little} impact on general performances (see Table \ref{tab:sensitivity_study}). 
From a computational perspective, this algorithm is highly scalable for large datasets, as it drastically reduces memory and computational requirements by processing only one mini-batch at a time. Moreover, it can take full advantage of GPU parallelization, further accelerating training and enabling efficient handling of high-dimensional or large-scale data (see \ref{annex:computation_time}).
Yet, the mini-batch size may affect the learning trajectory due to the approximation of the partition function, and we noticed its influence on the final model behaviour. Similarly to increasing the weight decay, we observed that decreasing the mini-batch size may smooth species spatial densities. Although mini-batch size is one of the least sensitive hyper-parameters, identifying an optimal mini-batch size remains important, and it may interact with other hyper-parameters such as the learning rate and number of epochs. 

{
One classical limitation of PO-based SDMs is that presence–absence (PA) data are often unavailable to validate model hyper-parameters. To address this issue, we evaluated hyper-parameter tuning directly on PO data and found that it reliably identifies suitable values, thereby removing the need for PA data during model training. The corresponding analyses are provided in the Appendix \ref{annex:cv}.}


{Overall,} DeepMaxent provides improvements over both traditional presence-only SDMs and previous deep learning-based {approaches} on a variety of case studies. {It} leverages neural networks to learn complex, high-order non-linear relationships directly from data, enabling flexible and expressive modelling, while {benefiting} from the principles behind the success of the original Maxent. 
However, several methodological directions remain open. 
Future extensions could integrate additional ecological information, such as species traits or phylogenetic relationships, to model inter-species dependencies more explicitly, an approach explored in models like HMSC~\citep{ovaskainen2017species}. {Additionally,} DeepMaxent {could} offer a framework for developing and testing bias-reduction strategies in deep learning-based SDMs, such as improved background sampling or observation models, to enhance robustness in heterogeneous and biased datasets.

\bibliography{template}%

\bibliographystyle{plainnat}
\newpage

\appendix

\section{Supplementary theoretical aspects}

\subsection{The relationship between Poisson, DeepMaxent and Maxent loss functions}
\label{annex:equivalence}

{We show here that the global minimizer of DeepMaxent's loss for a species probability distribution across sites $(\tilde{\lambda}_{ij})_{i\in 1,...,K}$ is equivalent to the one of the Poisson regression estimate, even though the Poisson regression adds the estimation of a multiplicative factor for the intensity captured by the parameter $b_j$. Then, using the equivalence between the log-linear Poisson regression and Maxent (\cite{renner2013equivalence}), we show that DeepMaxent is a non-linear generalization of Maxent, i.e. it reduces to Maxent for a log-linear model where $g_{\theta}(x_i)=x_i$.}

{We consider a single species $j$ for simplicity and the site index $i\in \lbrace 1,...,K\rbrace $.}d

{The occurrence count of $j$ at site $i$ is noted  $y_{ij}\in\mathbb{N}$. We further note the total count $y_j^+= \sum_{i=1}^K y_{ij} >0$ and normalised count $\tilde{y}_{ij} = y_{ij} / y_j^+ $. }

{The model predicted intensity in cell $i$ is $\lambda_{ij}=\lambda_j(x_i)\ge 0$ with our main notations, i.e. some parameterized function of the site covariates, not necessarily log-linear. We also note $Z_j = \sum_{i=1}^K \lambda_{ij}$ and $\tilde{\lambda}_{ij}=\lambda_{ij}/Z_j$.}

{First, we'll show the equivalence between the Poisson loss (negative log-likelihood) and DeepMaxent's loss. This equivalence is strongly related to the Poisson trick (\cite{lee2017poisson}), but we write the proof below with our notations for transparency.}

{The Poisson loss (negative log-likelihood) for species $j$ is written in equation \ref{eq:poisson_loss_single} below consistently with equation \ref{eq:poisson_loss} (we simply removed the constant factor)}:

\begin{equation}
    \label{eq:poisson_loss_single}
\begin{array}{rcl}
      \mathcal{L}_{\mathcal{P}}(\Lambda_{.j},Y_{.j}) & = & \sum_{i=1}^K (\lambda_{ij} - y_{ij} \log(\lambda_{ij})) \\
      & = & Z_j - \sum_{i=1}^K y_{ij} \log(\lambda_{ij}) \\
      & = & Z_j - \left( \sum_{i=1}^K y_{ij} \log(\lambda_{ij}/Z_j) + y_j^+ \log(Z_j) \right) \\
     & = & Z_j - y_j^+ \log(Z_j) - \sum_{i=1}^K y_{ij} \log(\lambda_{ij}/Z_j) \\
     & = & \underbrace{Z_j - y_j^+ \log(Z_j)}_{\text{Poisson loss for total count } y_j^+} + y_j^+ \underbrace{(-\sum_{i=1}^K \tilde{y}_{ij} \log(\tilde{\lambda}_{ij}))}_{\text{DeepMaxent loss for }j}
\end{array}
\end{equation}

{Hence, the Poisson loss can be separated into a first term for the total count and a second term for the distribution across sites, which turns out to be the loss term of DeepMaxent for species $j$. Now we can easily show that, with the parametric functional form $\lambda_{ij}=\lambda_j(x_i)=\exp(b_j + \sum_{c=1}^C \gamma_{jc} g_{\theta}(x_i))$ proposed in DeepMaxent, the second term is independent of $b_j$, as it simplifies in $\tilde{\lambda}_{ij}=\frac{\exp(b_j + \sum_{c=1}^C \gamma_{jc} g_{\theta}(x_i))}{ \sum_{i'} \exp(b_j + \sum_{c=1}^C \gamma_{jc} g_{\theta}(x_{i'})} = \frac{ \exp(\sum_{c=1}^C \gamma_{jc} g_{\theta}(x_i)) }{ \sum_{i'} \exp(\sum_{c=1}^C \gamma_{jc} g_{\theta}(x_{i'}) }$. 
Consequently, the parameters of this Poisson regression can be estimated in two steps, (i) minimizing the DeepMaxent loss (cross-entropy) to estimate $(\hat{\theta},\hat{\gamma_{j1}},...,\hat{\gamma_{jC}})$, and (ii) estimate $b_j$ separately via the total count term. For this latter step, we simply have to solve $\hat{Z}_j = e^{\hat{b_j}} \sum_i e^{\sum_c \hat{\gamma}_{jc} g_{\hat{\theta}}(x_i)} = y_j^+$ for $\hat{b}_j$, that is $\hat{b_j}=\log\left(\frac{y_j^+}{\sum_i \exp(\sum_c \hat{\gamma}_{jc} g_{\hat{\theta}}(x_i))} \right) $. }

{Consider now the particular case where $\lambda_{ij}$ is a log-linear function of the covariates $x_i$, i.e. $g_{\theta}(x_i)=x_i$, in the Poisson loss above. This is the negative log-likelihood of the log-linear Poisson regression that is equivalent to Maxent, as shown by theorem 1 of \cite{renner2013equivalence}. Hence, the equivalence that we have shown above means that the DeepMaxent loss and estimate are equivalent to the ones of Maxent in this simple particular case. More broadly, DeepMaxent's loss generalizes Maxent's loss by allowing a more flexible model for the distribution across sites, and preserving the link to the Poisson loss. }

\subsection{A modified batched SGD algorithm for DeepMaxent: The empirical minimizer of all batch-wise losses minimizes DeepMaxent's loss}
\label{annex:batchglobal}

In DeepMaxent, the term batch is not to be understood in its classical sense in machine learning or statistics. This is because the formula of the loss for a batch is not simply the restriction of the full loss to the terms of that batch: The partition function,{ used to normalise the intensity, is also reduced to the terms of the batch}. Thus, unlike in a classical setting, using the SGD with this type of batches might not necessarily lead to an estimator that is good regarding the full loss. To address this, we show below that {the empirical} estimator that minimises all the batch-wise losses {is the empirical estimator minimizing the full loss. However, this estimator exactly predicts the data, which, in general, is neither possible given a constrained model nor what we want. Even though the property doesn't guarantee that the actual model parameters minimizing all batch-wise losses will actually minimize the full loss, it gives us the insight that they should be close, which justifies this training procedure.}

\textbf{Notations:} Note for any $n\in \mathbb{N}^\star$, $\Delta^n=\lbrace t_0,...,t_n \in (\mathbb{R}^+)^{n+1} \/ \sum_{i=0}^n t_i=1,\; \forall i, t_i\ge 0 \rbrace$ the $n$-probability simplex, i.e. the space of probability distributions on a set of $n+1$ elements. Without loss of generality, we consider a single species with occurrence (pseudo-)count $y_k$ in site $k$. $\forall k\in \{1,\dots,K\}, y_k>0$, consistently with our implementation, where $0<\epsilon <<1$ is added to each raw occurrence count to avoid numerical problems. For any $B \subset \{1,\dots,K\}$, we note $y_B:=\lbrace y_k \rbrace_{k\in B} $, and also apply this notation to $\lambda_B$. 
We express the batch cross-entropy loss function of DeepMaxent with our notation in equation \ref{eq:ce_batch}.

\begin{equation}
    \mathcal{L}(\lambda_B,y_B)= -\frac{1}{|B|} \sum_{k \in B} \frac{y_k}{\sum_{i\in B}y_i} \log\left( \frac{\lambda_k}{\sum_{i\in B}\lambda_i} \right)
\label{eq:ce_batch}
\end{equation}

\; \\
According to Gibbs inequality, we have that $\underset{p \in \Delta^{|B|-1}}{\text{argmin}} \;  \mathcal{L}(p,y_B) =\lbrace y_k / \sum_{i\in B} y_i \rbrace_{k\in B}$. 

Now, assume that an estimator of the intensity $\lambda^*_{\{1,\dots,K\}}$ ($\forall k \in \{1,\dots,K\}, \lambda^*_k > 0$) minimises our loss for any mini-batch of size $n$ ($1<n<K$). Formally, $\lambda^*$ fulfils the property $P(n)$: $\forall B\subset \{1,\dots,K\}$ such that $|B|=n, \; \lbrace \lambda^*_k/\sum_{i\in B}\lambda^*_i \rbrace_{k\in B}  = \lbrace y_k / \sum_{i\in B} y_i \rbrace_{k\in B} $

\;\\
As a preliminary corollary, we have that $P(n) \Rightarrow P'(n)$: $\forall B_1,B_2 \subset \{1,\dots,K\} \slash |B_1|=|B_2|=n $ and $B_1 \cap B_2 \neq \emptyset, \; \frac{\sum_{i\in B_1}\lambda_i^*}{\sum_{i\in B_2}\lambda_i^*}=\frac{\sum_{i\in B_1}y_i}{\sum_{i\in B_2}y_i}$, because from $P(n)$ we get $\frac{\lambda_k^*}{y_k}=\frac{\sum_{i\in B_1}\lambda_i^*}{\sum_{i\in B_1}y_i} =\frac{\sum_{i\in B_2}\lambda_i^*}{\sum_{i\in B_2}y_i}$ which leads to the result.

\textbf{Property:} $P(n)\Rightarrow P(K)$, that is,
 if $\lambda^*$ fulfils $P(n)$, it minimises as well the full loss $\mathcal{L}(\lambda^*_{\{1,\dots,K\}},y_{\{1,\dots,K\}} )$. 

 \; \\*
 
\textbf{Proof:} Let's show that $P(n) \Rightarrow P(n+1)$, implying $P(K)$ by induction.

$P(n+1) \Leftrightarrow \forall B\subset \{1,\dots,K\}, l\in \{1,\dots,K\}, \, k\notin B$, we have $\lbrace \lambda^*_k/\sum_{i\in B \cup \lbrace l \rbrace}\lambda^*_i \rbrace_{k\in B \cup \lbrace l \rbrace}= \lbrace y_k/\sum_{i\in B \cup \lbrace l \rbrace}y_i \rbrace_{k\in B \cup \lbrace l \rbrace} $

Let be such $B$ and $l$, then $\forall i,j\in B, i \neq j$, 

\begin{equation}
    \frac{\lambda^*_i}{\sum_{k\in B \cup l } \lambda^*_k} = \frac{\lambda^*_i}{\sum_{k\in B \cup l \backslash j } \lambda^*_k + \lambda^*_j}  = \frac{\lambda^*_i/\sum_{k\in B} \lambda^*_k}{\frac{\sum_{k\in B \cup l \backslash j } \lambda^*_k}{\sum_{k\in B} \lambda^*_k} + \frac{\lambda^*_j}{\sum_{k\in B} \lambda^*_k}} \underset{P(n),P'(n)}{=}  \frac{y_i/\sum_{k\in B} y_k}{\frac{\sum_{k\in B \cup l \backslash j } y_k}{\sum_{k\in B} y_k} + \frac{y_j}{\sum_{k\in B} y_k}}  = \frac{y_i}{\sum_{k\in B \cup l \backslash j } y_k + y_j} = \frac{y_i}{\sum_{k\in B \cup l  } y_k }
\end{equation}

So $\lambda^*$ satisfies $P(n+1)$.

\; \\
\; \\

\section{DeepMaxent architecture}

\subsection{Residual MLP architecture}

\begin{figure}[htbp]
    \centering
        \centering
        \includegraphics[width=0.7\linewidth]{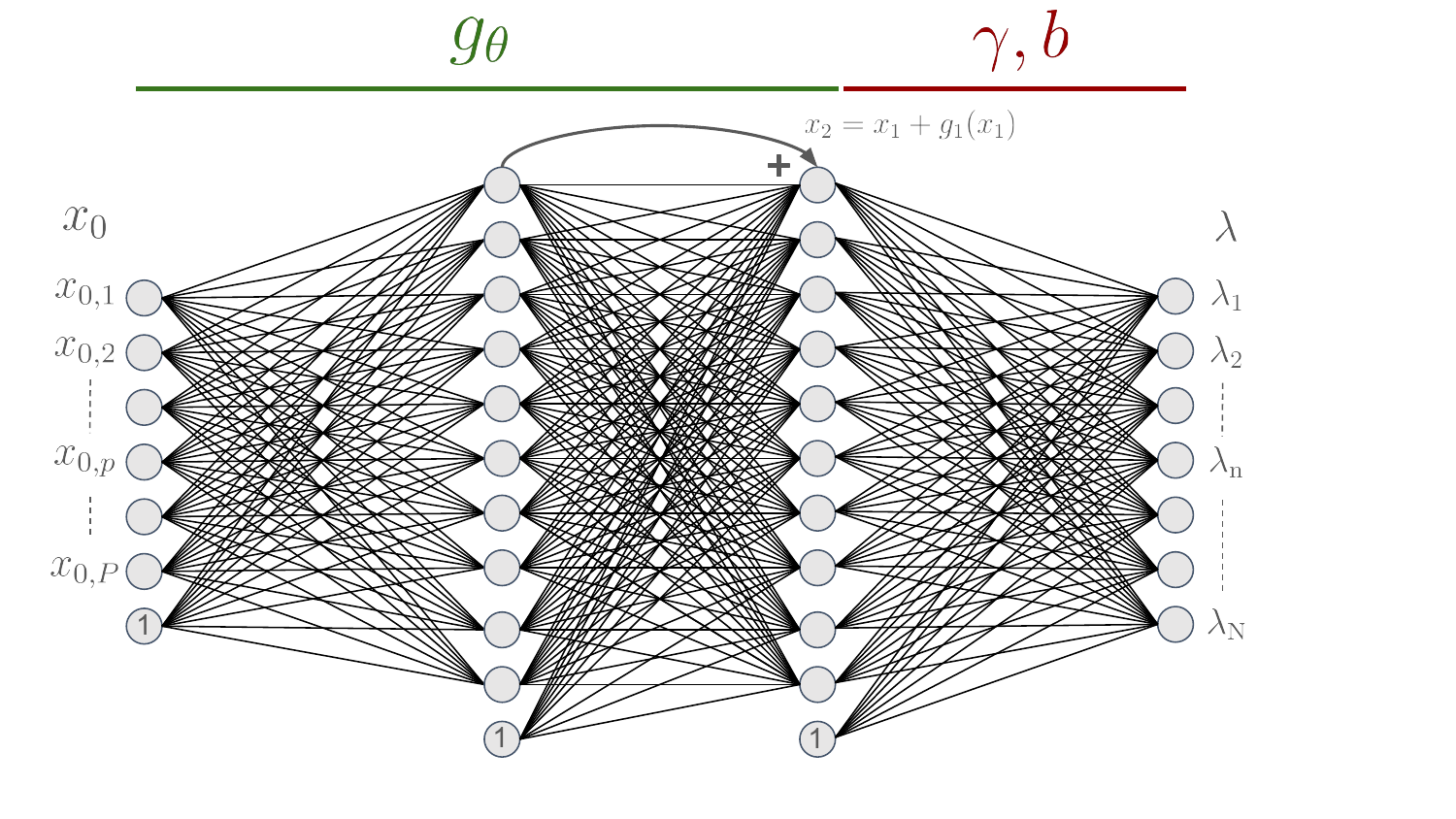}
    \caption{The residual neural network to estimate the intensity $\lambda$ from variable input ($x_0$), where $P$ as is the variable number input, where $C$ is the number of hidden layer nodes, and $N$ denotes the number of species (or target categories). The illustrated case involves two hidden layers. In the special case where there is only one hidden layer, no residual addition is applied.}
    \label{fig:architectures}
\end{figure}

DeepMaxent is architecture-agnostic, and in this work we chose a $g$ based on a residual neural network architecture (see Figure \ref{fig:architectures}), which takes the vector of environmental variables $\mathbf{x}$ as input. 
$\mathbf{x}$ is transformed by a sequence of hidden layers $g=g_1\circ g_2\circ \dots \circ g_L$, whose number $L$ and size are the main hyper-parameters. Residual connections between adjacent hidden layers were incorporated introducing shortcuts that allow the input of a layer to bypass the non-linear transformations and be directly added to the output ($\mathbf{x}_{l+1} = \mathbf{x}_l + g_l(\mathbf{x}_l)$)~\citep{he_deep_2016}. 
The output of the last hidden layer ($g(\mathbf{x})$), used as our latent feature vector, leads to the output layer, which is composed of each species intensity measure $\mathbf{\lambda}_j(\mathbf{x})$.

\subsection{Convolutional DeepMaxent architecture (ResNet18-based)}

{In addition to the multilayer perceptron architecture described above, DeepMaxent can accommodate feature extractors based on convolutional neural networks when the environmental covariates have an inherent spatial, spectral or temporal structure. We explored an instantiation of $g_\theta$ based on a modified ResNet-18 backbone~\citep{he_deep_2016}, adapted from~\citep{picek_geoplant_2025} specifically for the Landsat-Geoplant use case. The input tensor, representing multi-channel layers, is first normalised and then passed through a ResNet-18 in which the initial convolutional layer is adapted to accept $C_{\text{in}}$ channels instead of the standard three. As large-scale pooling may remove fine-grained spatial information relevant to species distributions, the initial max-pooling operation is disabled. The convolutional blocks learn spatially structured latent features at multiple scales, producing a 1000-dimensional representation after the final global pooling stage. This representation is further normalised and transformed through two fully connected layers to obtain the final latent feature vector $g_\theta(\mathbf{x})$, which plays the same role as in the MLP-based version of DeepMaxent. This convolutional variant enables the model to exploit local spatial patterns, textures, or temporal gradients present in high-resolution environmental rasters, while preserving the multi-species output structure described previously.}

\section{Cross-validating DeepMaxent using PO data}
\label{annex:cv}
In this study, model calibration was performed using presence-only (PO) data, whereas model evaluation used presence-absence (PA) data with a specific metric, the Area Under the Curve (AUC). This difference introduces a fundamental challenge in choosing the appropriate evaluation metric in the cross-validation step using PO data. PO and PA data are collected under different conditions with bias related to inhomogeneous sampling effort in the case of PO data. 
To fine-tune the DeepMaxent model’s hyper-parameters, spatial blocking was employed for cross-validation, drawing on geographical data to optimise the model’s performance. 
The cross-validation was performed following the approach of \citet{zbinden_selection_2024,roberts_cross-validation_2017}. 
In more detail, we performed a spatially stratified 10-fold cross-validation using a grid of 25 spatial blocks. Each fold used a unique subset of blocks for validation, distinct from previous folds, with blocks selected randomly yet balanced to ensure an even distribution of presence data across folds. 

\begin{table}[h!]
    \centering
    \caption{Parameters and Tested Values for cross-validation.}
    \label{tab:parameter_values}
    \begin{tabular}{ll}
\toprule
        \textbf{Parameter} & \textbf{Tested Values} \\
       \midrule
        Learning rate & 0.0002, 0.002, 0.02\\
        Mini-batch size & 250, 2500 \\
        Hidden layers & 1, 2, 3  \\
    \end{tabular}
\end{table}

To achieve this, all combinations of hyper-parameter values visible in Table \ref{tab:parameter_values} were evaluated.
The evaluation criterion used was the average AUC across all validation sets.
The {best} model was obtained with two hidden layers, a learning rate of 0.0002 and a mini-batch size of 250. 

For DeepMaxent's loss, the best scores were achieved with a learning rate of 0.0002, two hidden layers, and a mini-batch size of 250.
However, for the Poisson and BCE losses, cross-validation suggests a different optimal learning rate (lr = 0.002).

\begin{table}[h!]
    \centering
    \caption{{All combinations of hyper-parameter values and their cross-validation results on PO data for each loss.}}
    \label{tab:parameter_combinations}
    \begin{tabular}{lllllll}
        \toprule
{Learning rate} &{Mini-batch Size} & {Hidden Layers}  & \multicolumn{4}{c}{AUC} \\ %
 & & & BCE & CE & DeepMaxent &  Poisson \\
\toprule
0.0002 & 250 & 1 & 0.582 & 0.645 & 0.662 & 0.568 \\
0.0002 & 250 & 2 & 0.611 & \textbf{0.654} & \textbf{0.668} & 0.591 \\
0.0002 & 250 & 3 & 0.600 & 0.640 & 0.653 & 0.588 \\
0.0002 & 2500 & 1 & 0.518 & 0.586 & 0.622 & 0.505 \\
0.0002 & 2500 & 2 & 0.548 & 0.619 & 0.645 & 0.535 \\
0.0002 & 2500 & 3 & 0.540 & 0.609 & 0.634 & 0.525 \\
0.002 & 250 & 1 & 0.649 & 0.647 & 0.654 & 0.644 \\
0.002 & 250 & 2 & \textbf{0.656} & 0.645 & 0.659 & \textbf{0.655} \\
0.002 & 250 & 3 & 0.652 & 0.646 & 0.646 & 0.650 \\
0.002 & 2500 & 1 & 0.597 & 0.654 & 0.661 & 0.578 \\
0.002 & 2500 & 2 & 0.616 & 0.649 & 0.661 & 0.596 \\
0.002 & 2500 & 3 & 0.602 & 0.646 & 0.648 & 0.585 \\
0.02 & 250 & 1 & 0.644 & 0.634 & 0.633 & 0.643 \\
0.02 & 250 & 2 & 0.645 & 0.627 & 0.631 & 0.643 \\
0.02 & 250 & 3 & 0.643 & 0.621 & 0.640 & 0.633 \\
0.02 & 2500 & 1 & 0.637 & 0.644 & 0.634 & 0.609 \\
0.02 & 2500 & 2 & 0.626 & 0.624 & 0.650 & 0.604 \\
0.02 & 2500 & 3 & 0.596 & 0.621 & 0.643 & 0.582 \\
\bottomrule
\end{tabular}
\end{table}

\begin{table}[htbp]
  \caption{Comparison of BCE and Poisson loss performances by region-averaged AUC and general averaged AUC over all regions.}
  \label{tab:CV_poisson_and_bce}
  \centering
  \begin{tabular}{lllccccc}
    \toprule
    & \multicolumn{6}{c}{Regions}                          \\
    \cmidrule(r){2-7}
                    & AWT & CAN & NSW & NZ & SA & SWI & avg \\
                        \midrule
        \multicolumn{8}{l}{\textbf{Results from the literature}} \\ 
    \multicolumn{8}{l}{\textit{lr=0.002; mini-batch size = 250; hidden-layers=2}} \\
   BCE (using TGB)& 0.699 & 0.658 & 0.731 & 0.722 & 0.802 & 0.809 & 0.737\\
      Poisson (using TGB)& 0.709 & 0.677 & 0.740 & 0.729 & 0.803 & 0.819 & 0.746\\
          \multicolumn{8}{l}{\textit{lr=0.0002; mini-batch size = 250; hidden-layers=2}} \\
             BCE (using TGB)& {0.723} & 0.726 & {0.743} & 0.739 & 0.803 & {0.846} & {0.763} \\
      Poisson (using TGB)& {0.712} & {0.727} & {0.732} & {0.731} & 0.800 & {0.846} & {0.758}\\
    \bottomrule
  \end{tabular}
\end{table}

Although cross-validation initially suggested different optimal hyper-parameters for BCE and Poisson loss (lr = 0.002, Table \ref{tab:parameter_combinations}), these two losses were also tested with optimal hyper-parameters retained with DeepMaxent's loss on PA data to ensure fairness. Interestingly, the results confirmed that the best-performing hyper-parameters were the same as those identified for DeepMaxent's loss (Table \ref{tab:CV_poisson_and_bce}), suggesting that these settings were optimal across different loss functions.
This raises an important question about how to properly perform cross-validation when working with presence-only (PO) data, which may contain biases due to sampling effort or environmental factors. Since models are ultimately evaluated using AUC on unbiased presence-absence (PA) test data, there is a risk that hyper-parameters optimised on biased PO data may not generalize well to true PA conditions. Ideally, cross-validation should be performed using PA data, at least in part, before applying the models to the PA test set. This would help ensure that the chosen hyper-parameters lead to models that generalize well to unbiased evaluation data, reducing the risk of overfitting to PO-specific biases.

\section{Sensitivity analyses}
\label{annex:detailed_sensitivity}

\subsection{Sensitivity to the mini-batch size}
The impact of mini-batch size on the average AUC values obtained by the DeepMaxent method in the different regions of the dataset is illustrated in Table \ref{tab:batch}.

\begin{table}[htbp]
        \caption{Impact of mini-batch size on DeepMaxent performance across different regions. The table presents the accuracy scores for each region (AWT, CAN, NSW, NZ, SA, SWI) and the average (avg) performance across all regions.}
    \label{tab:batch}
      \centering
\begin{tabular}{lllccccc}
    \toprule
  Mini-batch size     & \multicolumn{6}{c}{regions}                          \\
    \cmidrule(r){2-7}
                 & AWT & CAN & NSW & NZ & SA & SWI & avg \\
                 \midrule
10 & 0.702 & \underline{0.732} & {0.747}& \textbf{0.755} & \textbf{0.805} & {0.848} & 0.765\\
25 & 0.705 & {0.731} & {0.748} & \textbf{0.755} & \textbf{0.805} & \underline{0.849} & {0.765}\\
100 & 0.711 & {0.730} & \textbf{0.754} & \textbf{0.755} & \underline{0.804} & \underline{0.849} & \textbf{0.767} \\
250 & {0.714} & \underline{0.732} & \underline{0.752} & \underline{0.754} & 0.803 & \textbf{0.850} & \textbf{0.767}\\
1000 & \underline{0.715} & \textbf{0.733} & 0.750 & 0.749 & 0.801 & 0.848 & \underline{0.766}\\
2500 & \textbf{0.716} & \textbf{0.733} & 0.748 & 0.742 & 0.800 & 0.847 & 0.764 \\
    \end{tabular}
\end{table}

The mean AUC values for small mini-batch sizes such as mini-batch size 10 averaged 0.765. Increasing the mini-batch size to 25 does not improve the overall AUC. Even though very small, the improvement comes mainly from AWT, NSW, and SWI. 
The average AUC in all regions shows better values for mini-batch sizes of 100 and 250, with values of 0.767. 
Conversely, when the mini-batch size increases to 1000 or 2500, the average AUC values for all regions decrease, reaching 0.766 and 0.764, respectively. 

When analysing specific regions, we notice that regions like AWT and NSW show high variability in their AUC values. AWT exhibits the most instability, as its AUC value fluctuates between 0.702 and 0.716 across different mini-batch sizes, suggesting that it may require a larger mini-batch size for consistent performance. Conversely, SWI maintains a high AUC value around 0.848 with minimal changes, indicating that it is relatively unaffected by mini-batch size variations. Similarly, regions such as CAN, NSW and SA also show limited impact, with AUC values remaining close to 0.732, 0.751 and 0.801, respectively, regardless of the mini-batch size.

For some regions, such as SWI and CAN, the effects of mini-batch size appear to be minimal, with good approximations obtained regardless of mini-batch size. However, a medium mini-batch size can improve computational efficiency on machines, providing a balance between performance and resource utilisation compared to larger mini-batch sizes.
On the other hand, the AWT remains the most unstable region, indicating that it requires more precise parametrisation to achieve optimal performance.

\subsection{Sensitivity to the number of hidden layers}

Table \ref{tab:layers} presents the impact of hidden layer number on the performance of the DeepMaxent model, measured by AUC values across regions: AWT, CAN, NSW, NZ, SA, and SWI.

\begin{table}[htbp]
    \caption{Impact of hidden layer number on DeepMaxent performance with L2 regularisation (w=1e-2) across different regions. The table presents AUC value for each region (AWT, CAN, NSW, NZ, SA, SWI) and the average (avg) AUC across all regions.}
    \label{tab:layers}
    \centering
\begin{tabular}{lllccccc}
    \toprule
  Hidden layer number  & \multicolumn{6}{c}{regions}                          \\
    \cmidrule(r){2-7}
                 & AWT & CAN & NSW & NZ & SA & SWI & avg \\
                 \midrule
1 & \textbf{0.719} & \textbf{0.735} & \textbf{0.753} & 0.751 & 0.800 & 0.847  & \textbf{0.767}\\
2 & \underline{0.714} & \underline{0.732} & \underline{0.752} & \textbf{0.754} & 0.803 & \textbf{0.850}  & \textbf{0.767}\\
3 &  0.710 & 0.729 & \underline{0.752} & \underline{0.752} & \underline{0.805} & \textbf{0.850}  & \underline{0.766}\\
4 &  0.705 & 0.727 & 0.748 & 0.752 & \underline{0.805} & \underline{0.849} & {0.764}\\
5 & 0.703 & 0.722 & 0.743 & 0.749 & \textbf{0.806} & 0.848  & 0.762\\
6 & 0.698 & 0.719 & 0.739 & 0.747 & \textbf{0.806} & 0.847  & 0.759\\
    \end{tabular}

\end{table}

In some regions, the AUC values decrease as the number of hidden layers increases. 
For instance, in AWT, CAN and NSW, the highest AUC values are achieved with just one hidden layer. 
In contrast, for regions like SA, increasing the number of layers leads to a slight improvement in AUC performance, reaching an AUC value of 0.806. In other regions, an optimum is reached, such as in SWI with an AUC of 0.850 and NZ with 0.754, both with two hidden layers.
With one layer, the model achieves an average AUC of 0.767, the highest overall, particularly performing well in regions like AWT with a AUC of 0.720 and NSW with an AUC of 0.754. 

The impact of the number of hidden layers varies by region. In some cases, the changes are minimal, such as for SWI, where the AUC fluctuates by only 0.003, while for the AWT dataset, the variations are more pronounced. Overall, using one or two hidden layers results in the highest average AUC across all regions, at 0.767. However, with two hidden layers, the model produces the highest or second-highest scores in most regions, suggesting that this configuration may represent the optimal balance across all regions rather than favouring a specific one. This is further supported by the Pearson correlation values provided in the appendix (see Table \ref{annex:pearson_layers}).

\subsection{Sensitivity to the number of neurons per hidden layer}

{Table \ref{tab:neurons} presents the impact of the number of neurons per hidden layer on the performance of DeepMaxent on the NCEAS dataset, all other hyper-parameters kept the same as for in main results, measured by AUC values across regions: AWT, CAN, NSW, NZ, SA, and SWI.}

\begin{table}[h]
\label{tab:neurons}
\centering
\begin{tabular}{c c}
\hline
\textbf{Hidden Size} & \textbf{AUC across all region} \\
\hline
50  & 0.75853 \\
100 & 0.76583 \\
250 & 0.76775 \\
500 & 0.76662 \\
750 & 0.76548 \\
\hline
\end{tabular}
\caption{AUC value as a function of hidden layer size for deepMaxent models.}
\end{table}

\subsection{Sensitivity to the L2 regularisation weight}

The average AUC values per region and across all regions for various weight decay values are shown in Table \ref{tab:results_regul}. 

\begin{table}[htbp]
  \caption{Comparison of DeepMaxent performance according to different weight decay values for L2 regularisation. The criteria are the average AUC per region and the average AUC for all regions. The best average AUC for each column is highlighted in bold, while the second-best averaged AUC is underlined.}
  \centering
  \begin{tabular}{lllccccc}
    \toprule
   weight decay value & \multicolumn{6}{c}{regions}                          \\
    \cmidrule(r){2-7}
                    & AWT & CAN & NSW & NZ & SA & SWI & avg \\
     \\ 
    0 & 0.713 & 0.719 & {0.750} & 0.744 & 0.804 & 0.843 & 0.762 \\
    \midrule
    1e-6  & 0.713 & 0.719 & 0.750 & 0.744 & \textbf{0.804} & 0.843  & 0.762\\
    3e-6  & 0.713 & 0.719 & 0.750 & 0.744 & \textbf{0.804} & 0.843 & 0.762\\
    1e-5  & 0.713 & 0.719 & 0.750 & 0.744 & \textbf{0.804} & 0.844 & 0.762\\
    3e-5  & 0.713 & 0.721 & 0.750 & 0.746 & \textbf{0.804} & 0.845 & 0.763\\
    1e-4  & 0.713 & 0.726 & 0.751 & \underline{0.750} & \textbf{0.804}& \underline{0.848} &  \underline{0.765}\\
    3e-4  & 0.714 & \underline{0.732} & \underline{0.752} & \textbf{0.754} & \underline{0.803} & \textbf{0.850} & \textbf{0.767}\\
    1e-3 & \underline{0.718} & \textbf{0.734} & \textbf{0.753} & 0.745 & 0.799 & 0.843 & \underline{0.765}\\
    3e-3 & \textbf{0.722} & \underline{0.732} & 0.749 & 0.723 & 0.780 & 0.833 & 0.757\\
    1e-2 & 0.716 & 0.730 & 0.740 & 0.707 & 0.754 & 0.805  & {0.742}\\
    3e-2 & 0.691 & 0.705 & 0.715 & 0.662 & 0.737 & 0.780 & {0.715}\\
    1e-1 & 0.642 & 0.661 & 0.670 & 0.585 & 0.669 & 0.718 & 0.657\\
    3e-1 & 0.592 & 0.636 & 0.615 & 0.560 & 0.604 & 0.644 &  0.609\\
    \bottomrule
  \end{tabular}
    \label{tab:results_regul}
\end{table}

Without regularisation (with w=0), DeepMaxent method achieved an average AUC value of 0.762, highlighted already a strong performance in comparison of classic method (see Table \ref{tab:results}). 
When L2 regularisation with a weight decay value of $w = 1 \times 10^{-6}$ or $w = 1 \times 10^{-5}$  is applied, the AUC values for each region, as well as the overall average AUC, remain approximatively unchanged. Marginal improvements on the order of 0.001 are observed for Switzerland.
These results are therefore approximately the same as those without regularisation. This suggests that the weight decay value is too small to significantly affect the outcomes, resulting in minimal L2 regularisation.

When the weight decay value is increased to $w = 3 \times 10^{-5}$, slight changes are observed, with noticeable improvements particularly in CAN and SWI regions. The overall average AUC begins to be affected by regularisation, with an overall AUC value of 0.763.
With a weight decay value of $1 \times 10^{-4}$, regularisation has a more pronounced effect on the results, yielding an overall average AUC of 0.765. Significant improvements are observed, particularly for CAN and SWI, with AUC values of 0.726 and 0.848, respectively.
When weight decay has a value of $3 \times 10^{-4}$, the overall AUC is the highest with a value of 0.767. In addition, five of the six regions obtained the best or the second-best values in this ablation study. This is the case for CAN, NSW, NZ, SA and SWI with AUC values of 0.732, 0.752, 0.754, 0.803 and 0.850, respectively. 

On the other hand, with higher weight decay values, the overall average AUC across all regions decreases. 
For a weight decay value of $3 \times 10^{-2}$, the mean AUC decreases slightly to reach a value of 0.742. 
All AUC values for all regions decrease.
Finally, with higher weight decay values of $1 \times 10^{-1}$ and $3 \times 10^{-2}$, the overall mean AUC decreases significantly.

\begin{table}[htbp]
    \centering
        \caption{Impact of hidden layer number on DeepMaxent performance. The table presents average AUC and Pearson coefficient values across all regions.}
    \label{annex:pearson_layers}
\begin{tabular}{lcc}
    \toprule
  Hidden layer number     & avg AUC & avg Pearson Coefficient \\
                 \midrule
1 & \textbf{0.767} & \underline{0.244}\\
2 & \textbf{0.767} & \textbf{0.247}\\
3 & \textbf{0.767} & {0.242}\\
4 & \underline{0.764} & 0.234\\
5 &  0.762 & 0.228\\
6 &  0.759 & 0.222\\
    \end{tabular}

\end{table}

\begin{table}[htbp]
    \centering
        \caption{Impact of mini-batch size on DeepMaxent performance across all regions. The table presents average AUC and Pearson coefficient values across all regions.}
    \label{annex:pearson_batch}
\begin{tabular}{lcc}
    \toprule
  Mini-batch size    & avg AUC & avg Pearson Coefficient \\
                 \midrule
10 & 0.765 & 0.231 \\
25 & {0.765} & 0.236 \\
100 & \textbf{0.767} & {0.244} \\
250 & \textbf{0.767} & \textbf{0.247} \\
1000  & \underline{0.766} & \underline{0.245} \\
2500 & 0.764 & 0.241 \\
    \end{tabular}

\end{table}


In a multi-species Maxent model, applying a normalisation based on the sum of $y$ (presences weighted by probability) for each species can indeed impact the distribution of relative presence probabilities between abundant and rare species. When a uniform weight is assigned to each species (for example, setting each species’ weight to 1), there is a risk that abundant species will have their occurrences assigned lower probabilities, while, conversely, the occurrences of rare species may be prioritised.

\begin{table}[htbp]
  \caption{Comparison of method performance of DeepMaxent using TGB and using a equal weight between species by region-averaged AUC and averaged over all regions. }
  \label{tab:results_equal_weight}
  \centering
  \begin{tabular}{llcccccc}
    \toprule
   weight decay value & \multicolumn{6}{c}{regions}\\
    \cmidrule(r){2-7}
            & AWT & CAN & NSW & NZ & SA & SWI & avg \\
    \midrule
     0 & 0.717 & 0.728 & 0.737 & 0.745 & 0.804 & 0.844 & 0.762 \\
     1e-6 &  0.716 & 0.725 & 0.742 & 0.745 & 0.804 & 0.844 & 0.763\\
     3e-6 & 0.716 & 0.726 & 0.742 & 0.746 & 0.804 & 0.844 & 0.763 \\
    1e-5 & 0.716 & 0.727 & 0.742 & 0.746 & 0.805 & 0.846 & 0.763\\
    3e-5 & 0.717 & 0.729 & 0.741 & 0.746 & 0.804 & 0.845 & 0.764 \\
    1e-4& 0.716 & 0.731 & 0.740 & 0.745 & 0.799 & 0.839 & 0.761  \\
    3e-4& 0.711 & 0.730 & 0.735 & 0.738 & 0.782 & 0.826 & 0.754 \\
    1e-3& 0.687 & 0.727 & 0.716 & 0.707 & 0.755 & 0.795 & 0.731\\
    3e-3& 0.629 & 0.700 & 0.681 & 0.682 & 0.708 & 0.776  & 0.696\\
    \bottomrule
  \end{tabular}
\end{table}

\begin{table}[htbp]
  \caption{Comparison of performance using DeepMaxent without TGB approach according to different weight decay values. }
  \label{tab:weight_decay_full}
  \centering
  \begin{tabular}{lllccccc}
    \toprule
   Weight decay value & \multicolumn{6}{c}{regions}                          \\
    \cmidrule(r){2-7}
                    & AWT & CAN & NSW & NZ & SA & SWI & avg \\
  1e-1 & 0.691 & 0.455 & 0.594 & 0.679 & 0.746 & 0.647 & 0.635\\
  1e-2 & 0.665 & 0.469 & 0.691 & 0.719 & 0.773 & 0.770 & 0.681\\
  1e-3 & 0.654 & 0.593 & 0.718 & 0.744 & 0.803 & 0.810 & 0.720\\
  1e-4 & 0.652 & 0.601 & 0.713 & 0.731 & 0.802 & 0.803 & 0.717\\
  1e-5 & 0.651 & 0.596 & 0.712 & 0.729 & 0.802 & 0.799 & 0.715\\
  1e-6 & 0.652 & 0.595 & 0.712 & 0.730 & 0.802 & 0.798 & 0.715\\
    \bottomrule
  \end{tabular}
\end{table}

\begin{table}[htbp]
    \centering
        \caption{Impact of weight decay value on performances of baseline losses. The table presents average AUC and Pearson coefficient values across all regions.}
    \label{annex:pearson_weight}
\begin{tabular}{llcc}
    \toprule
  Weight decay value     & loss& AUC & avg Pearson Coefficient \\
no   & BCE & 0.763 & 0.233 \\
3e5  & BCE & 0.764 & 0.235 \\
1e4  & BCE & 0.764 & 0.239 \\
3e4  & BCE & 0.763 & 0.240 \\
1e3  & BCE & 0.755 & 0.231 \\
3e3  & BCE & 0.751 & 0.208 \\

no   & CE & 0.744 & 0.162 \\
3e5  & CE & 0.743 & 0.161 \\
1e4  & CE & 0.744 & 0.162 \\
3e4  & CE & 0.744 & 0.162 \\
1e3  & CE & 0.744 & 0.162 \\
3e3  & CE & 0.744 & 0.162 \\

no   & Poisson loss & 0.758 & 0.241 \\
3e5  & Poisson loss & 0.759 & 0.242 \\
1e4  & Poisson loss & 0.759 & 0.243 \\
3e4  & Poisson loss & 0.757 & 0.241 \\
1e3  & Poisson loss & 0.748 & 0.228 \\
3e3  & Poisson loss & 0.721 & 0.198 \\

    \bottomrule
  \end{tabular}
    \label{tab:results_regul_other_methods}
\end{table}

\subsection{Ablation study of MLP components in DeepMaxent}

{To better understand the contributions of architectural components in the DeepMaxent MLP model, we conducted an ablation study using the optimal hyper-parameters identified in the main experiments. We evaluated the model under four variants:}

\begin{itemize}
    \item Full model: with skip connections and activation functions (ReLU).
    \item No skip connections: activation functions kept.
    \item No activation functions: skip connections kept.
    \item No skip connections and no activation functions: a baseline MLP.
\end{itemize}

{All models were retrained on the same dataset (NCEAS) with the same training procedure over ten same seeds. Performance was assessed via average AUC over species.}

\begin{table}[ht]
\centering
\caption{{Ablation study of MLP components in DeepMaxent across different regions. Models trained with optimal hyper-parameters.}}
\label{tab:ablation_deepmaxent_regions}
\begin{tabular}{lccccccc}
\toprule
                    & \multicolumn{6}{c}{Regions}                          & \\
\cmidrule(r){2-7}
\textbf{Model variant} & AWT  & CAN  & NSW  & NZ   & SA   & SWI  & \textbf{avg} \\
\midrule
Full model             & 0.712 & 0.732 & 0.752 & 0.753 & 0.806 & 0.850 & 0.768\\
No skip connections    & 0.714 & 0.731 & 0.752 & 0.747 & 0.802 & 0.848 & 0.766  \\
No activations         & 0.704 & 0.729 & 0.738 & 0.736 & 0.763 & 0.831 &  0.750 \\
No skip connections, no activations & 0.702 & 0.731 & 0.738 & 0.741 & 0.764 & 0.830 & 0.750 \\
\bottomrule
\end{tabular}
\end{table}

{The full model, incorporating both skip connections and activation functions, achieves the best average performance (0.768). Removing skip connections results in a slight marginal decrease, indicating that their contribution is beneficial but moderate. 
On the other hand, removing activation functions has a greater impact on performance, with a more pronounced decrease in the average AUC (0.750). These results highlight the importance of activation functions for the network's learning ability, while skip connections play a secondary but positive role.}

{These results highlight the importance of activation functions for the learning capacity of the network, reflecting the non-linear nature of the data. In addition, the flexibility provided by residual connections (skip connections) improves model optimisation by facilitating gradient propagation and enabling more stable and efficient learning of complex representations, which contributes to better generalisation.}

\subsection{The importance of multi-species modelling in DeepMaxent: a case study in Switzerland}
{
DeepMaxEnt was tested on each species independently over the SWI region, keeping the same set of parameters to build single-species models. In both cases, we applied the TGB correction to account for potential biases. For each species, we trained the model separately and evaluated its performance using the AUC metric. To compare these results with a multi-species approach, we calculated the average of the AUC values obtained for all the single-species models.}

\begin{table}[htbp]
    \centering
        \caption{Comparison of DeepMaxEnt performance between single-species and multi-species models in the Swiss region. The table reports AUC values for each approach, with results averaged across all species.}
    \label{annex:single_species_deepmaxent}
\begin{tabular}{lc}
    \toprule
  Method    & avg AUC for SWI \\
                 \midrule
Single-species DeepMaxent & 0.820  \\
Multi-species DeepMaxent & 0.850 \\
\midrule 
Maxent (with TGB) & 0.837 \\
    \bottomrule
  \end{tabular}
\end{table}

{
Table \ref{annex:single_species_deepmaxent} shows the results for the single-species and multi-species models. It shows that the multi-species model performs much better. This {may be explained by several reasons: (i) The} set of {hyper-parameter}s optimal for {the multi-species model may not be suited for each single species model, and (ii) learning a feature extractor across multiple species is more robust one independent feature extractor for each individual species, as already showed in the past (e.g. \cite{botella2018deep}). Adjusting hyper-parameters individually for each species could potentially improve performance of the single-species DeepMaxent models, but this approach would add up an important computational overload to the already important computational cost of fitting individual deep models.}}

\section{Statistical significance of performance differences}
\label{annex:statistical_tests}

{To assess whether the observed differences in predictive performance are statistically significant, we conducted pairwise Wilcoxon signed-rank tests between DeepMaxent and three alternative loss functions: Poisson, BCE and CE. These tests were applied to the macro AUC scores obtained over 10 random seeds, ensuring a fair comparison under identical experimental conditions.}

{As shown in Table~\ref{tab:wilcoxon_tests}, the resulting p-values are both below 0.002, indicating that the improvements achieved by DeepMaxent over both BCE and CE are statistically significant at the 0.01 level. The Wilcoxon statistic equals 0.0 in both cases, which further supports the consistent results of DeepMaxent across all tested seeds.}

\begin{table}[ht]
\centering
\caption{{Wilcoxon signed-rank tests over 10 random seeds comparing DeepMaxent to BCE and CE in terms of macro AUC. The low p-values indicate that the observed differences are statistically significant.}}
\begin{tabular}{lcc}
\toprule
\textbf{Comparison} & \textbf{Wilcoxon Statistic} & \textbf{p-value} \\
\midrule
DeepMaxent vs BCE & 0.0 & 0.00195 \\
DeepMaxent vs CE  & 0.0 & 0.0019 \\
DeepMaxent vs Poisson  & 0.0 & 0.0019 \\
\bottomrule
\end{tabular}
\label{tab:wilcoxon_tests}
\end{table}

\section{Comparison of performances across abundance classes on the NCEAS dataset}

\begin{figure}[h!]
    \centering
    \includegraphics[width=0.5\linewidth]{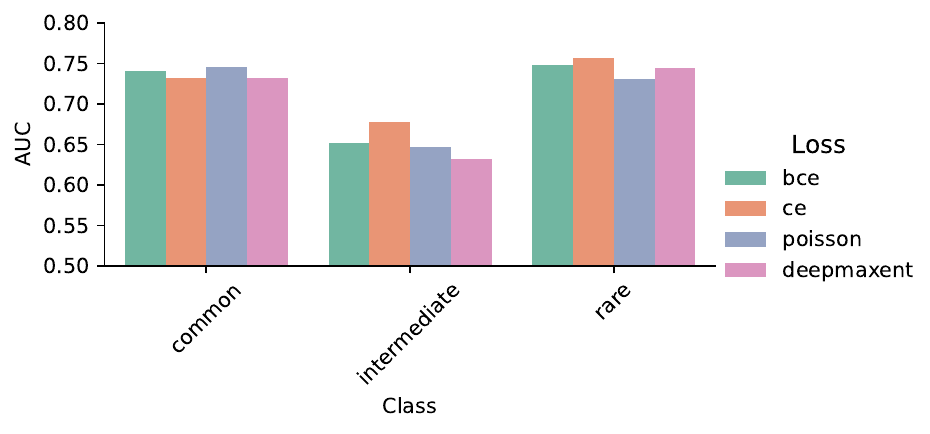}
    \caption{Comparison of average AUC values for AWT region by loss and abundance classes}
    \label{fig:barplot_AWT}
\end{figure}
\begin{figure}[h!]
    \centering
    \includegraphics[width=0.5\linewidth]{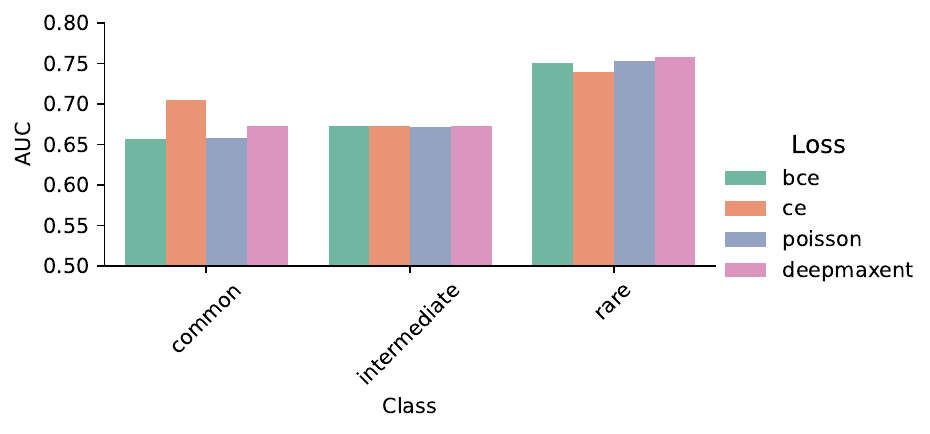}
    \caption{Comparison of average AUC values for CAN region by loss and abundance classes}
    \label{fig:barplot_CAN}
\end{figure}
\begin{figure}[h!]
    \centering
    \includegraphics[width=0.5\linewidth]{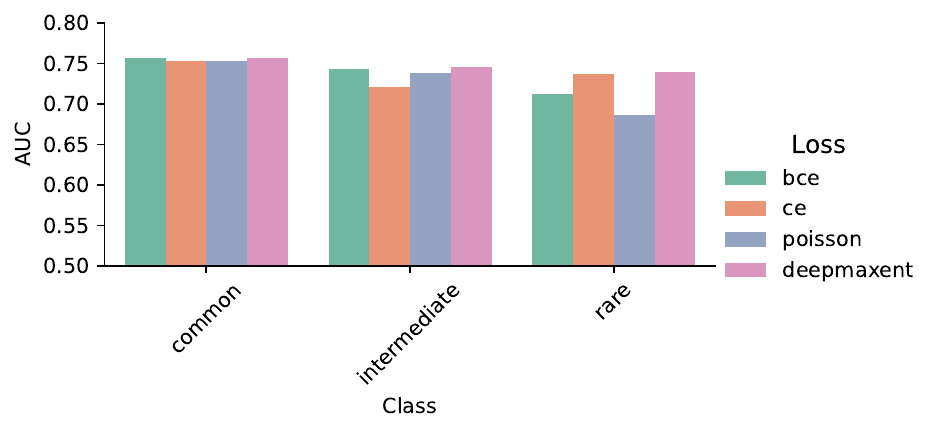}
    \caption{Comparison of average AUC values for NSW region by loss and abundance classes}
    \label{fig:barplot_NSW}
\end{figure}
\begin{figure}[h!]
    \centering
    \includegraphics[width=0.5\linewidth]{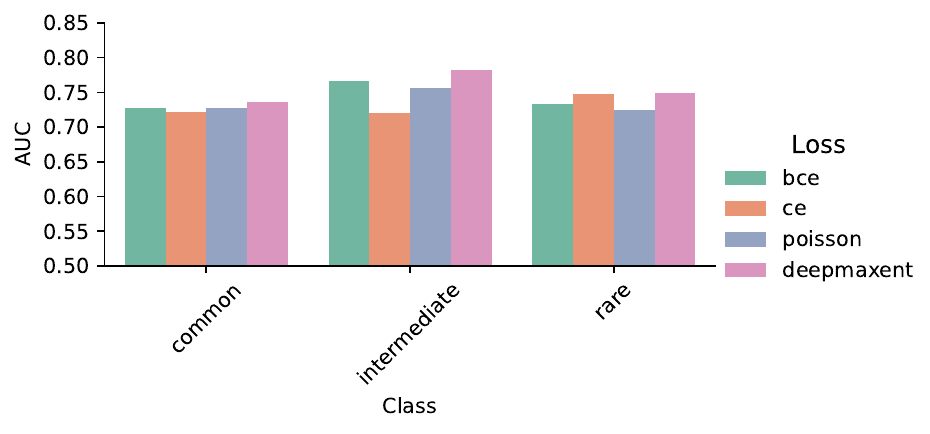}
    \caption{Comparison of average AUC values for NZ region by loss and abundance classes}
    \label{fig:barplot_NZ}
\end{figure}
\begin{figure}[h!]
    \centering
    \includegraphics[width=0.5\linewidth]{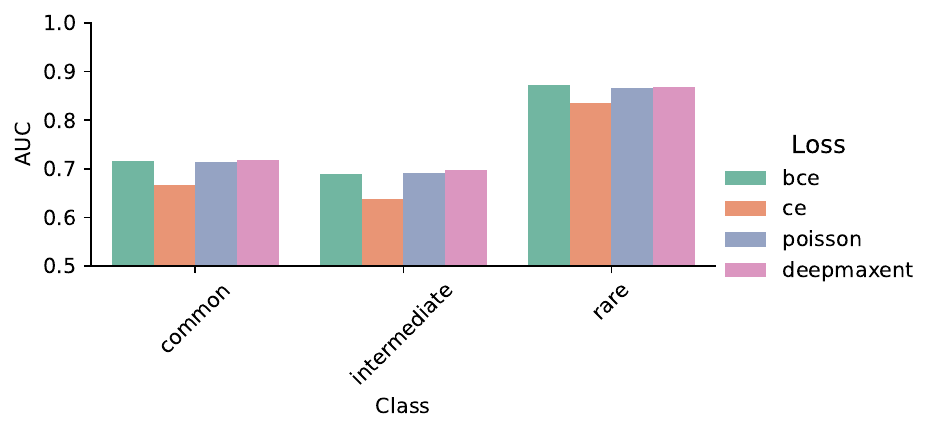}
    \caption{Comparison of average AUC values for SA region by loss and abundance classes}
    \label{fig:barplot_SA}
\end{figure}
\begin{figure}[h!]
    \centering
    \includegraphics[width=0.5\linewidth]{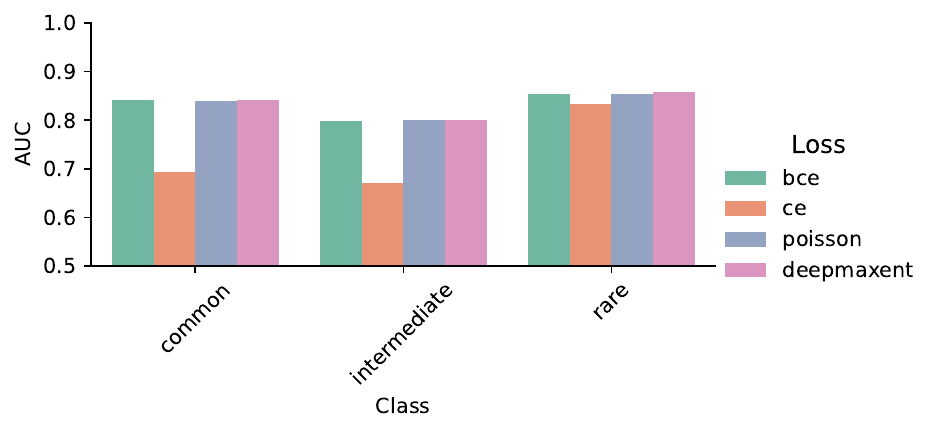}
    \caption{Comparison of average AUC values for SWI region by loss and abundance classes}
    \label{fig:barplot_SWI}
\end{figure}

Table \ref{annex:detailed_std_mean_by_species_class} shows the mean and standard deviation across all regions by species abundance class for each loss function. For common and intermediate species, BCE, Poisson, and DeepMaxent losses yield similar and relatively high values with low standard deviations, all outperforming the standard CE loss. For rare species, values are consistently higher across all losses, with DeepMaxent again achieving the best performance (0.786$\pm$0.001), suggesting it may be better suited for modelling the distributions of less frequent species. The CE loss displays the largest performance gap between common and rare species, highlighting its relative limitation in handling class imbalance.

\begin{table}[h!]
    \centering
        \caption{Mean value and standard deviation across all regions by species abundance class across all loss functions.}
    \label{annex:detailed_std_mean_by_species_class}
\begin{tabular}{lccc}
    \toprule
    & \multicolumn{3}{c}{species class}  \\
    \midrule
Loss & common & intermediate & rare \\
\midrule
CE (using TGB) & 0.712$\pm$0.003 & 0.683$\pm$0.003 & 0.775$\pm$0.002 \\
BCE (using TGB) & 0.740$\pm$0.003 & 0.720$\pm$0.003 & 0.778$\pm$0.002 \\
Poisson loss (using TGB) & 0.740$\pm$0.003 & 0.718$\pm$0.003 & 0.769$\pm$0.003 \\
DeepMaxent (using TGB) & 0.742$\pm$0.002 & 0.722$\pm$0.001 & 0.786$\pm$0.001 \\
    \bottomrule
  \end{tabular}
\end{table}

\section{Estimated probabilities maps by species with corresponding PA data, for CAN region}

\begin{figure}[h!]
    \centering
        \begin{subfigure}[b]{0.40\linewidth}
        \centering
        \includegraphics[width=\linewidth]{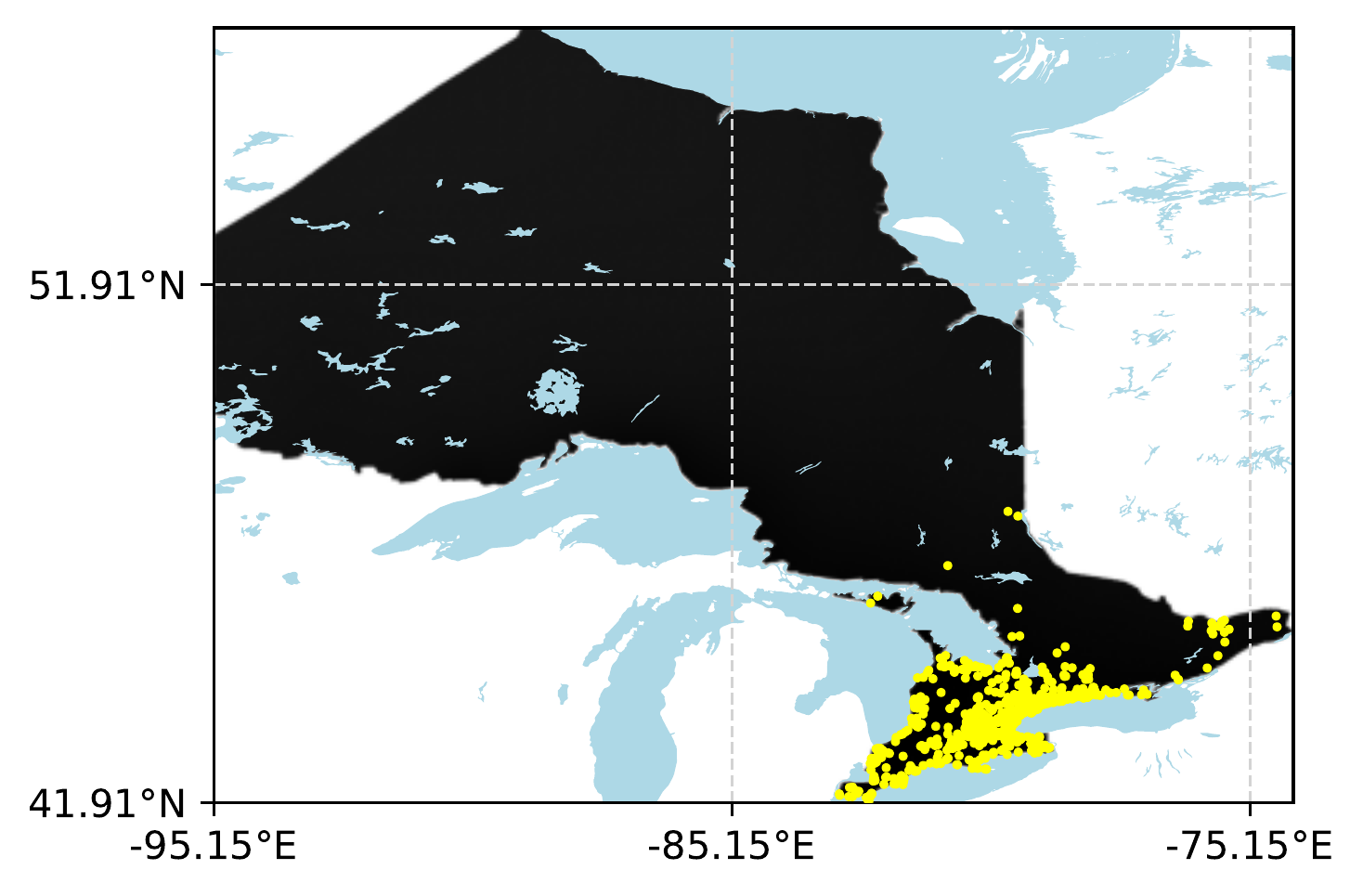}
        \caption{}
    \end{subfigure}
      \begin{subfigure}[b]{0.40\linewidth}
        \centering
        \includegraphics[width=\linewidth]{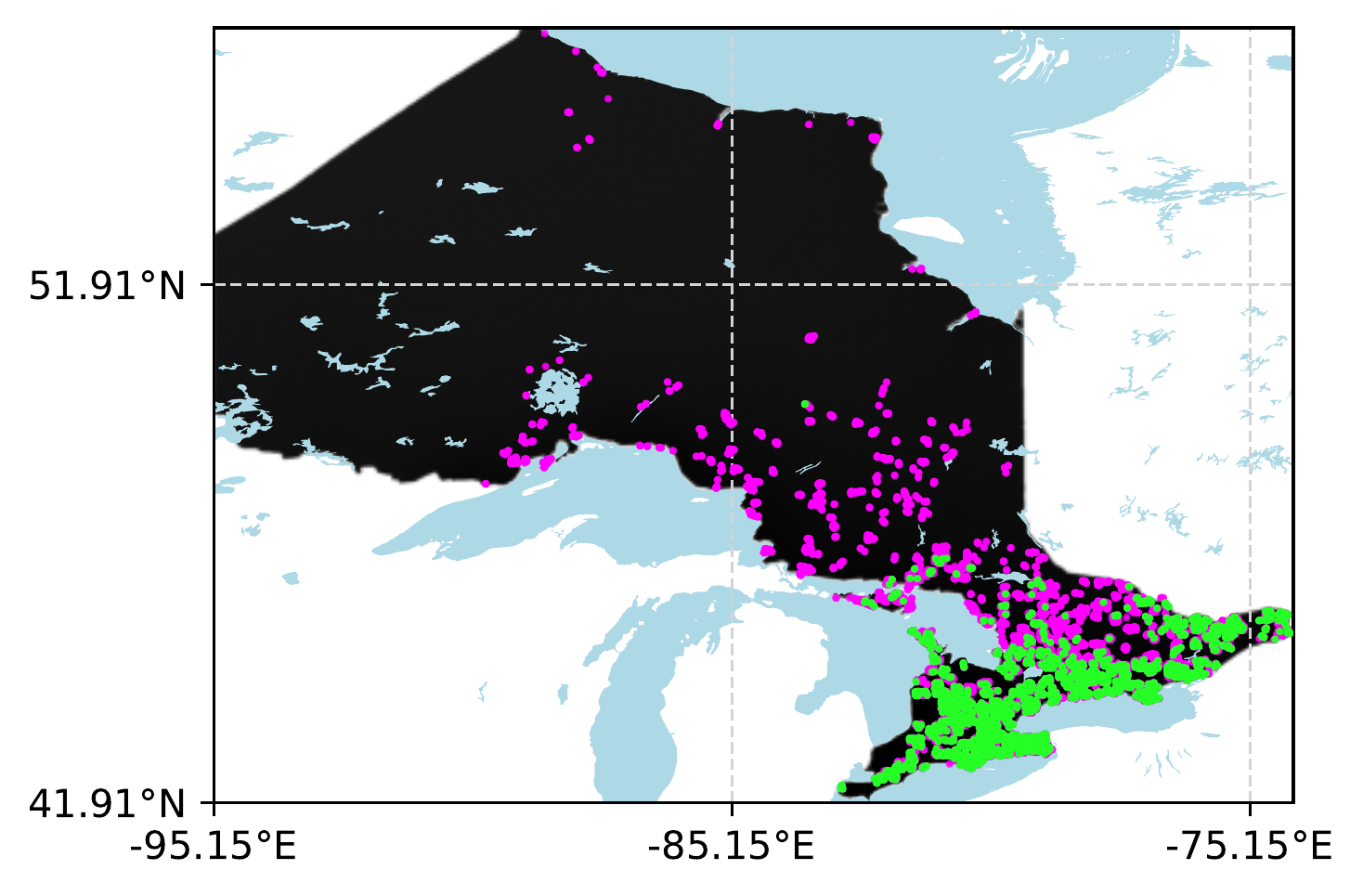}
        \caption{}
    \end{subfigure}
    \\
        \begin{subfigure}[b]{0.45\linewidth}
        \centering
        \includegraphics[width=\linewidth]{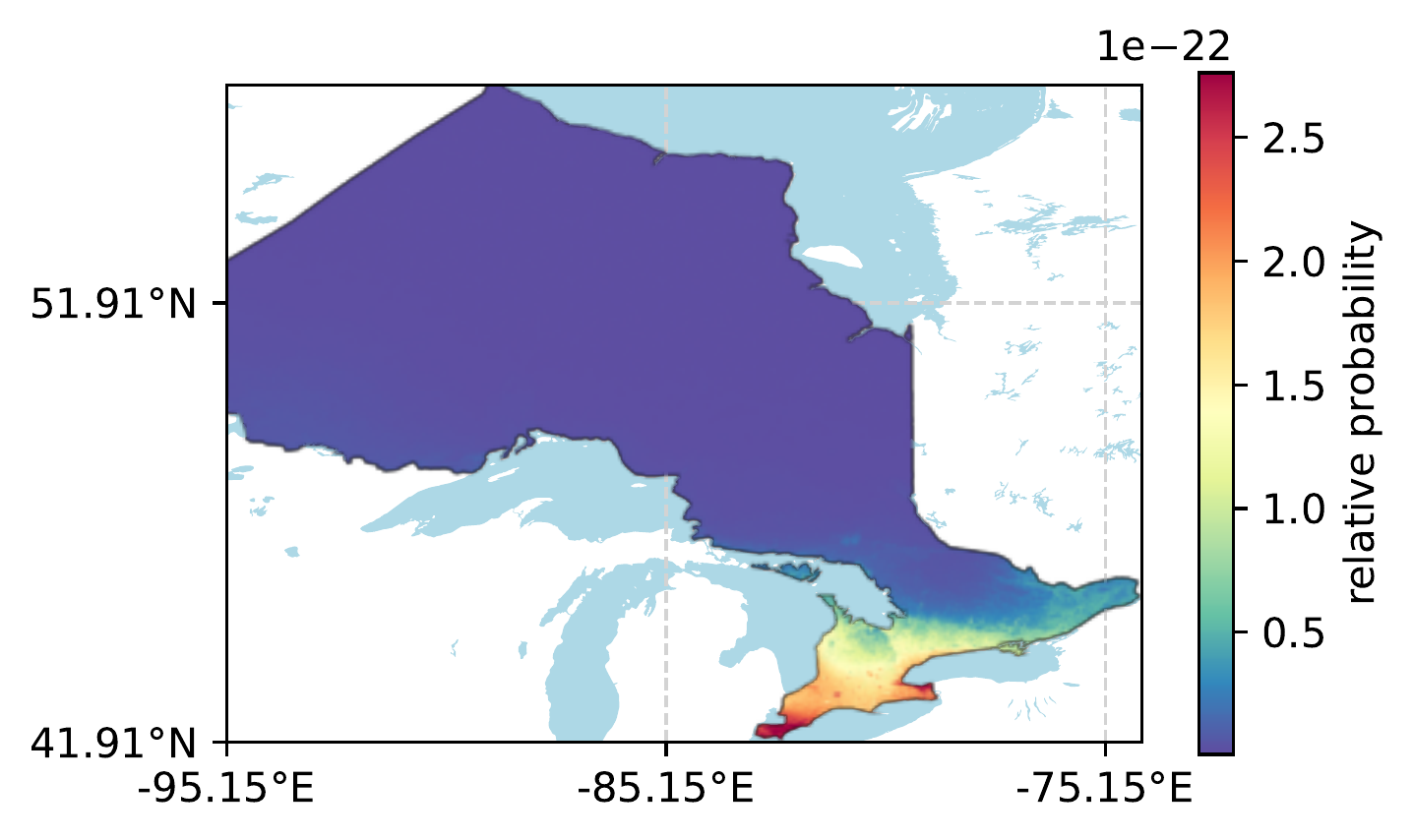}
        \caption{}
    \end{subfigure}
    \begin{subfigure}[b]{0.45\linewidth}
        \centering
        \includegraphics[width=\linewidth]{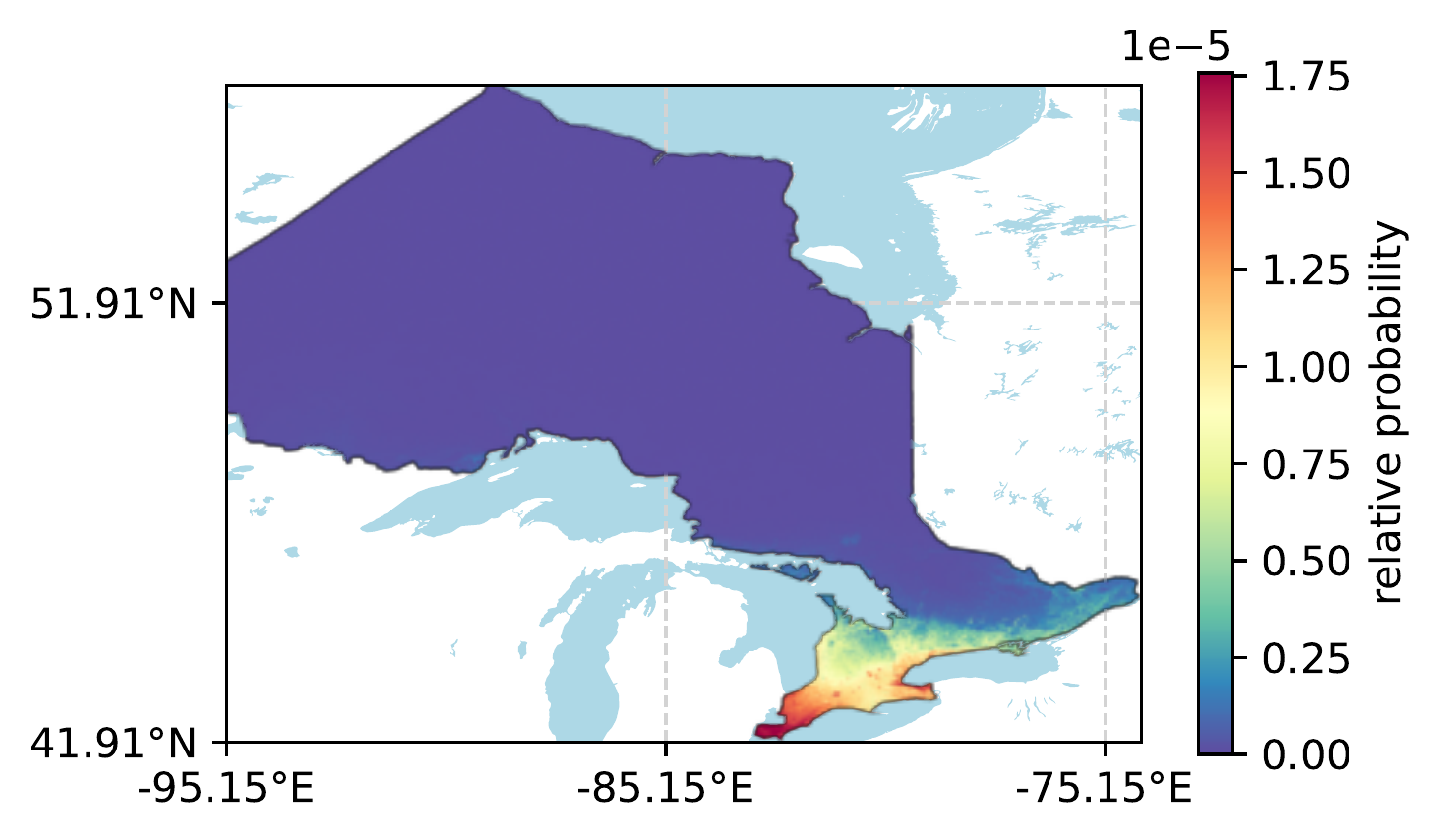}
        \caption{}
    \end{subfigure}
    \\
    \begin{subfigure}[b]{0.45\linewidth}
        \centering
        \includegraphics[width=\linewidth]{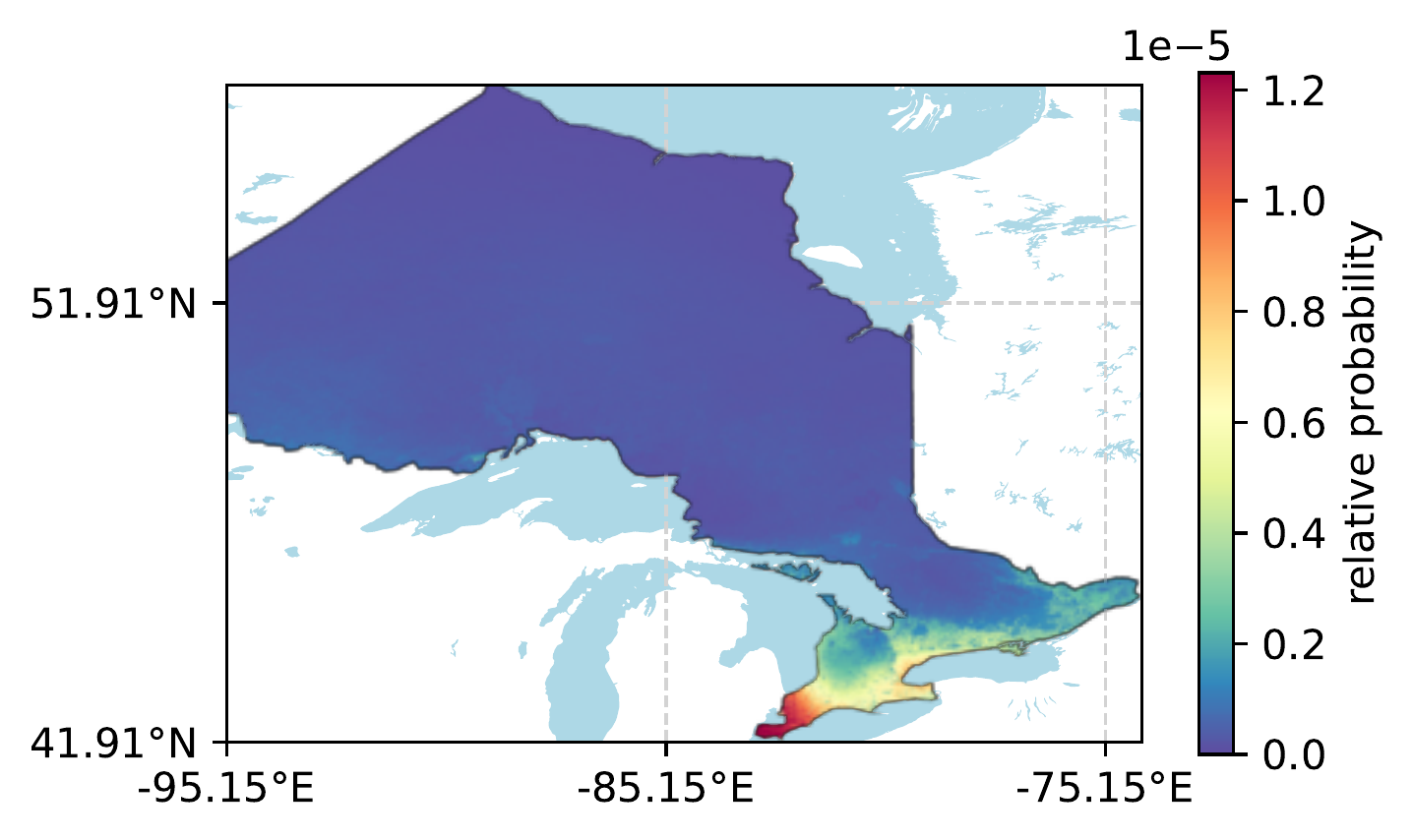}
        \caption{}
    \end{subfigure}
        \begin{subfigure}[b]{0.45\linewidth}
        \centering
        \includegraphics[width=\linewidth]{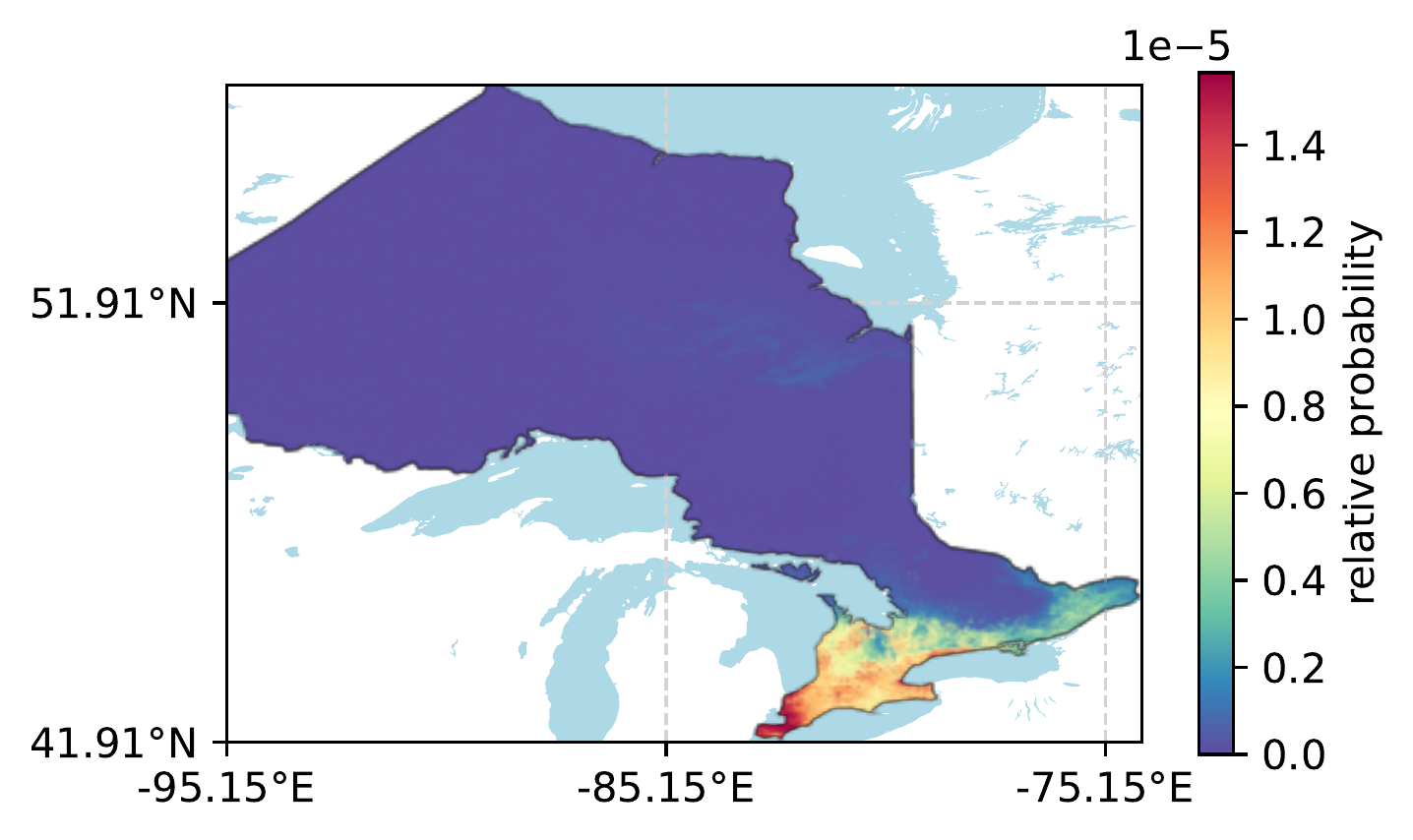}
        \caption{}
    \end{subfigure}
    \caption{Estimated relative probabilities for the species can02 (a common species):
(a) Presence-Only (PO) data,  {yellow points};
(b) Presence-Absence (PA) data  {where green corresponds to presences, and magenta to absences}.
(c–f) Estimated from different loss functions: (c) DeepMaxent, (d) Binary Cross-Entropy (BCE), (e) Cross-Entropy (CE), (f) Poisson loss.
    }
    \label{fig:allmaps_comparative_common_species}
\end{figure}

\begin{figure}[h!]
    \centering
    \renewcommand{\arraystretch}{0.1}
    \begin{tabular}{>{\centering}m{1cm} m{6cm} m{6cm}}
        & \textbf{Presence/Absence} & \textbf{Estimated relative probability} \\
        \midrule
        \textbf{can01} \newline 
        & \includegraphics[width=1.00\linewidth]{figures/MAPS_WITH_PA/CAN_bird_CE_0_MEAN_only_PA_1.png}
        & \includegraphics[width=1.00\linewidth]{figures/MAPS_WITH_PA/CAN_bird_CE_0_MEAN_1.png} \\
        
        \textbf{can02} \newline 
        & \includegraphics[width=1.00\linewidth]{figures/MAPS_WITH_PA/CAN_bird_CE_1_MEAN_only_PA_1.png}
        & \includegraphics[width=1.00\linewidth]{figures/MAPS_WITH_PA/CAN_bird_CE_1_MEAN_1.png} \\
        
        \textbf{can03} \newline 
        & \includegraphics[width =1.00\linewidth]{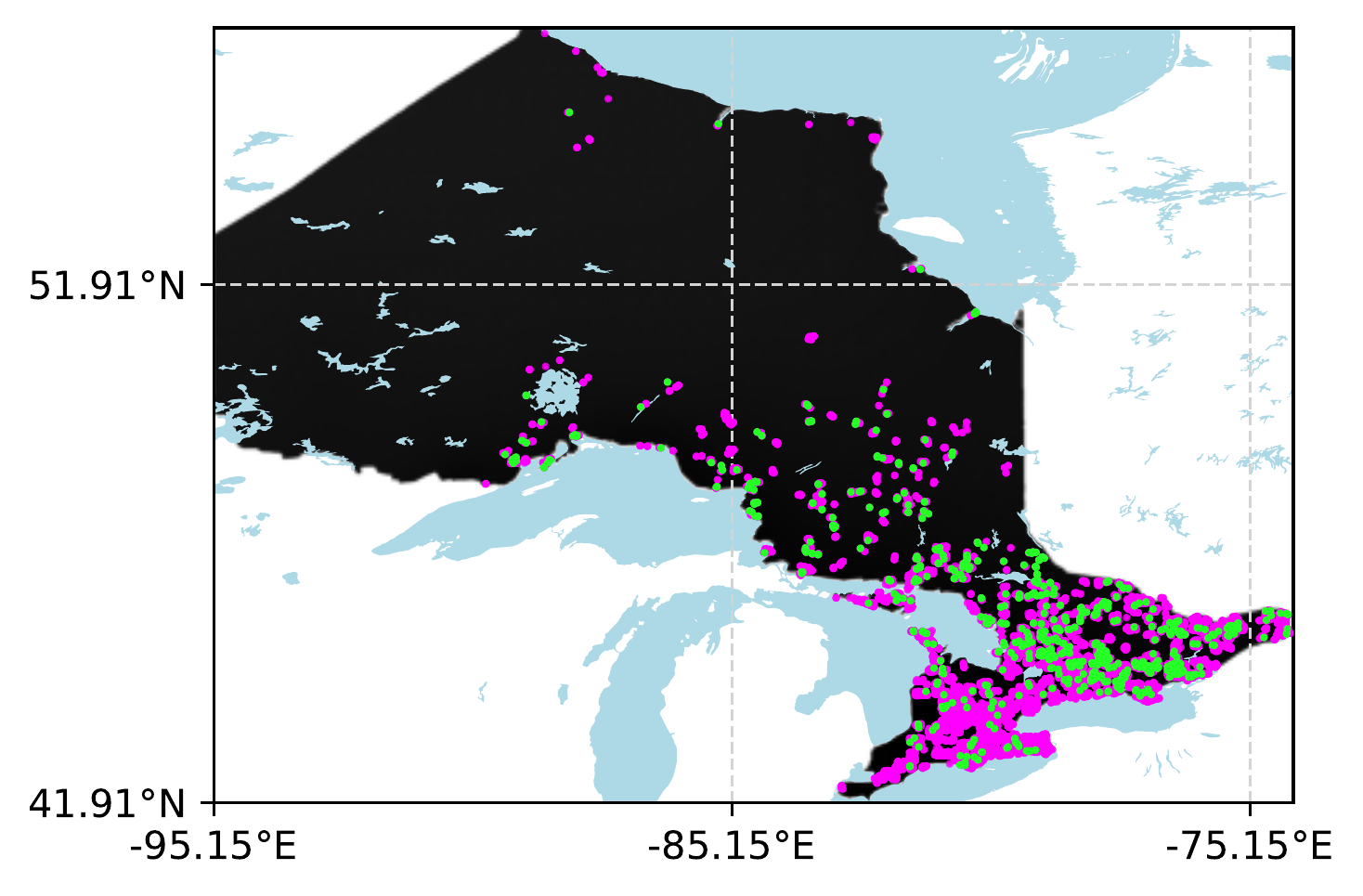}
        & \includegraphics[width=1.00\linewidth]{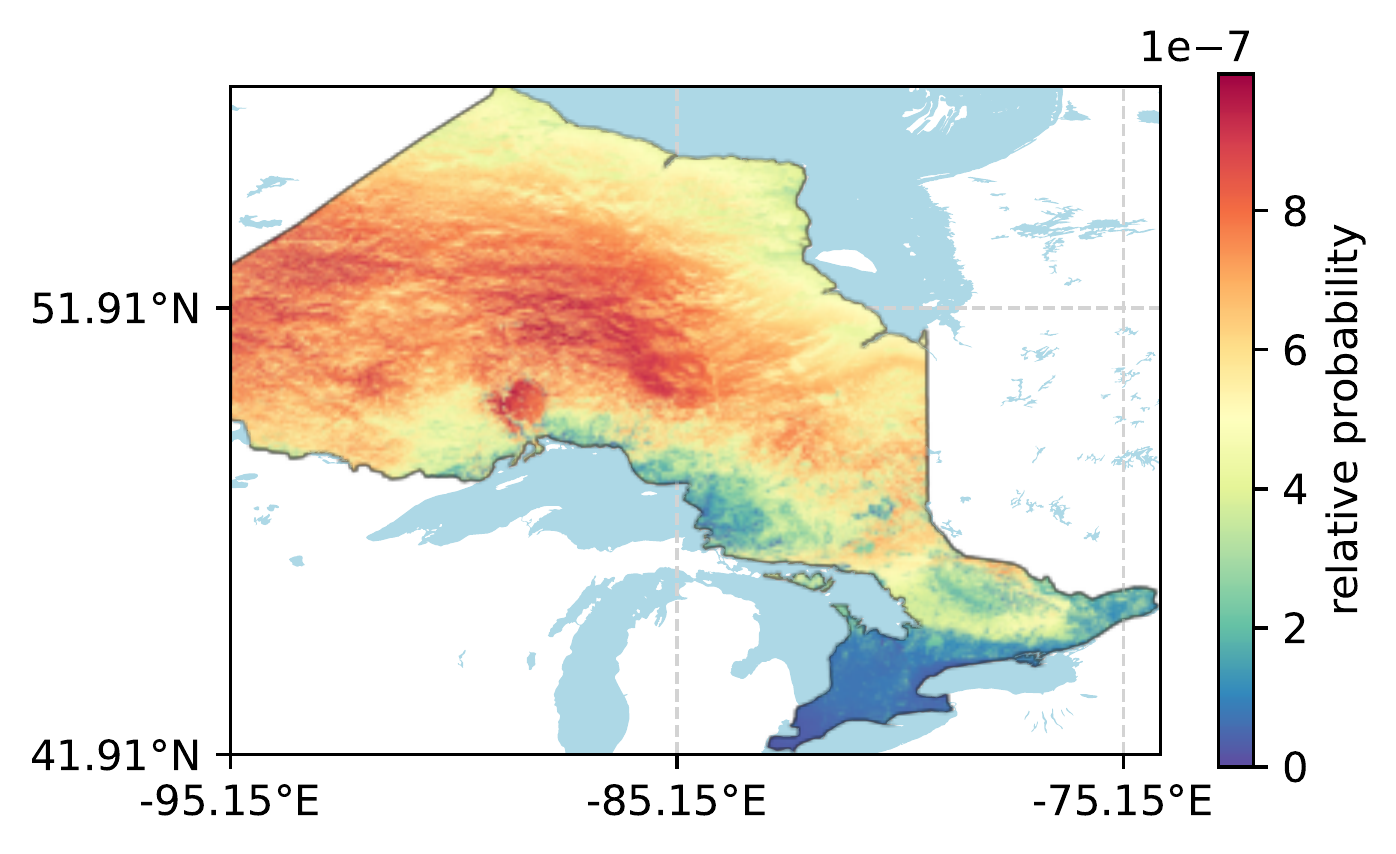} \\
        
        \textbf{can04} \newline
        & \includegraphics[width=1.00\linewidth]{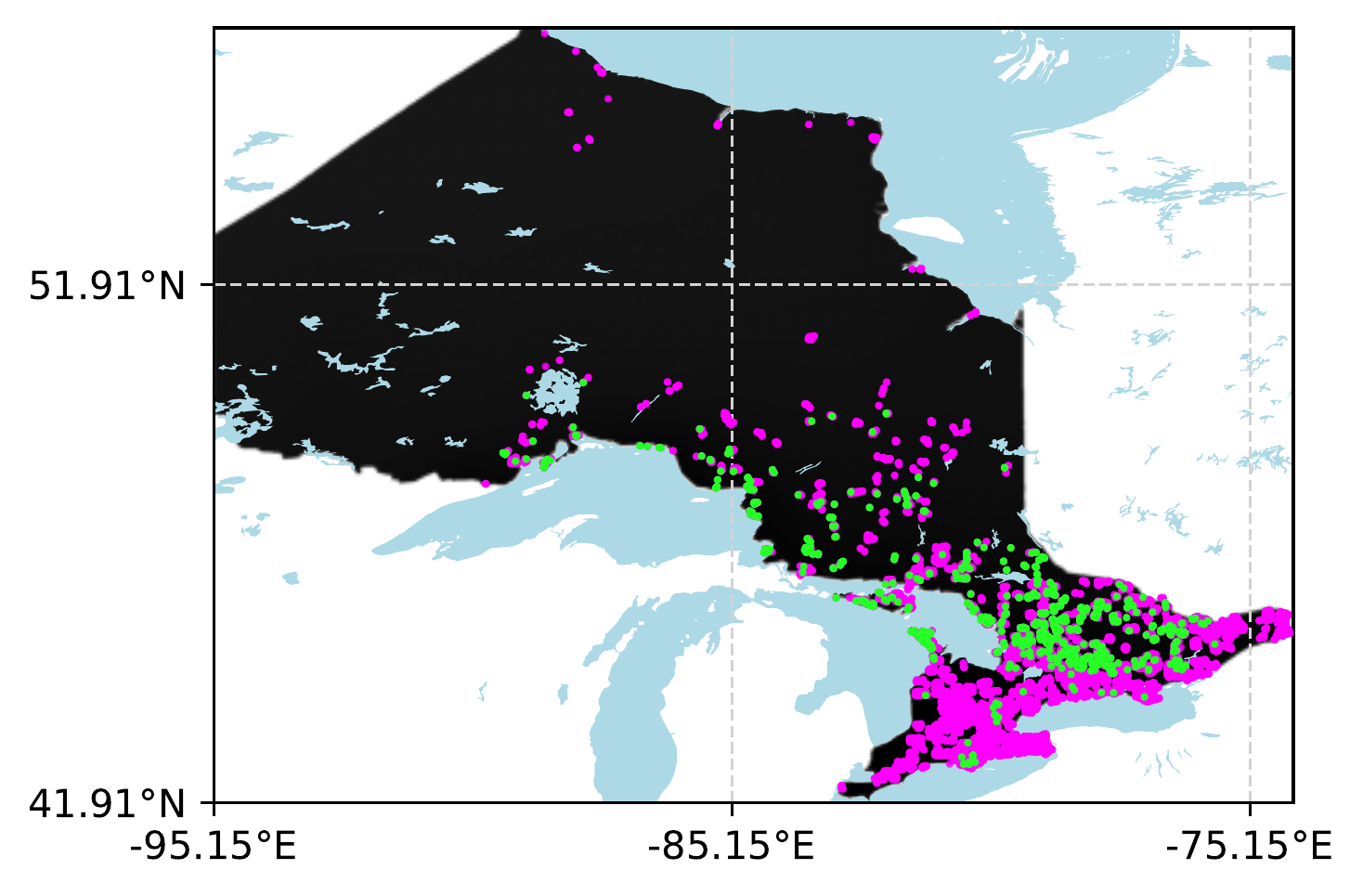}
        & \includegraphics[width=1.00\linewidth]{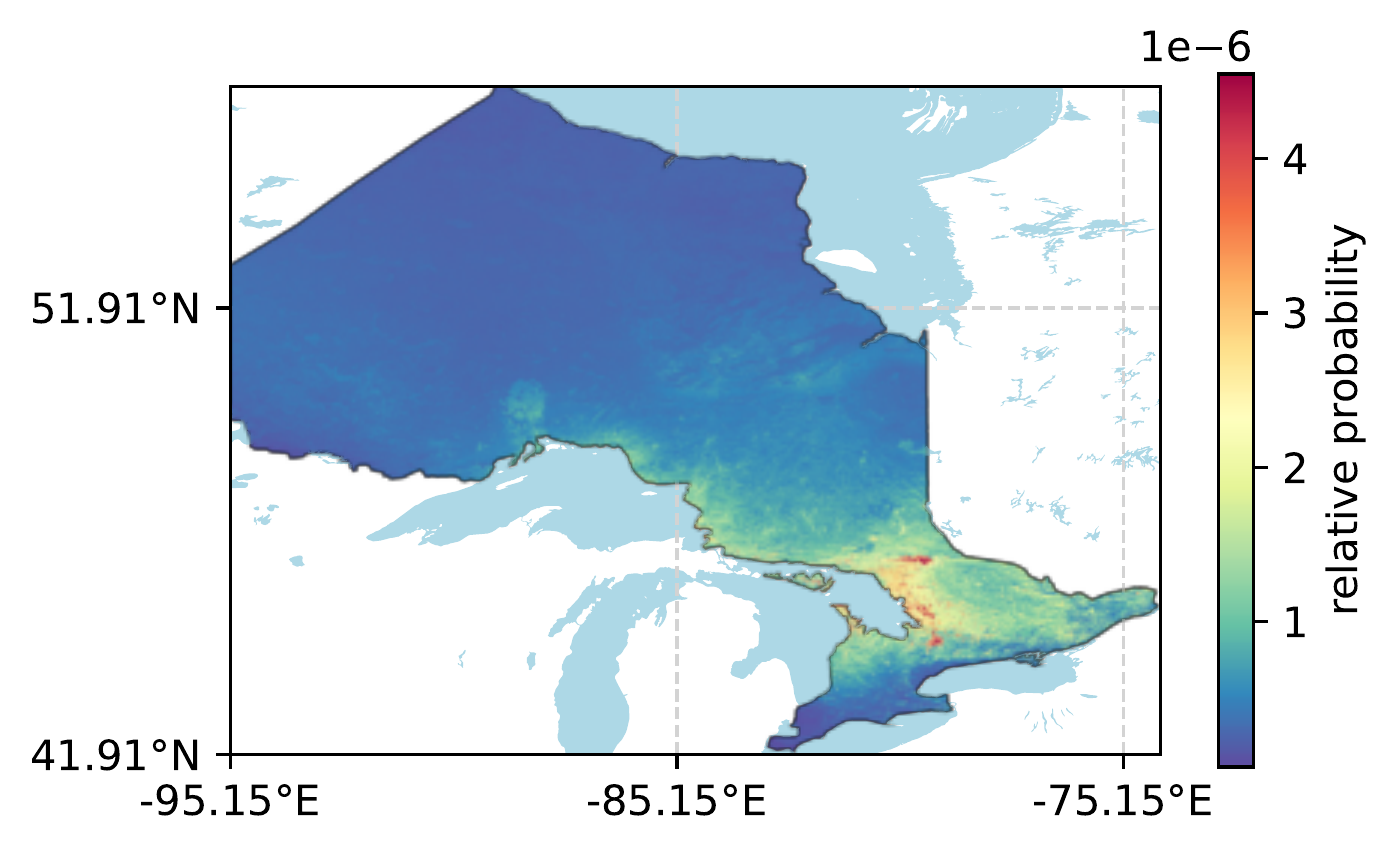} \\
        
        \textbf{can05} \newline
        & \includegraphics[width=1.00\linewidth]{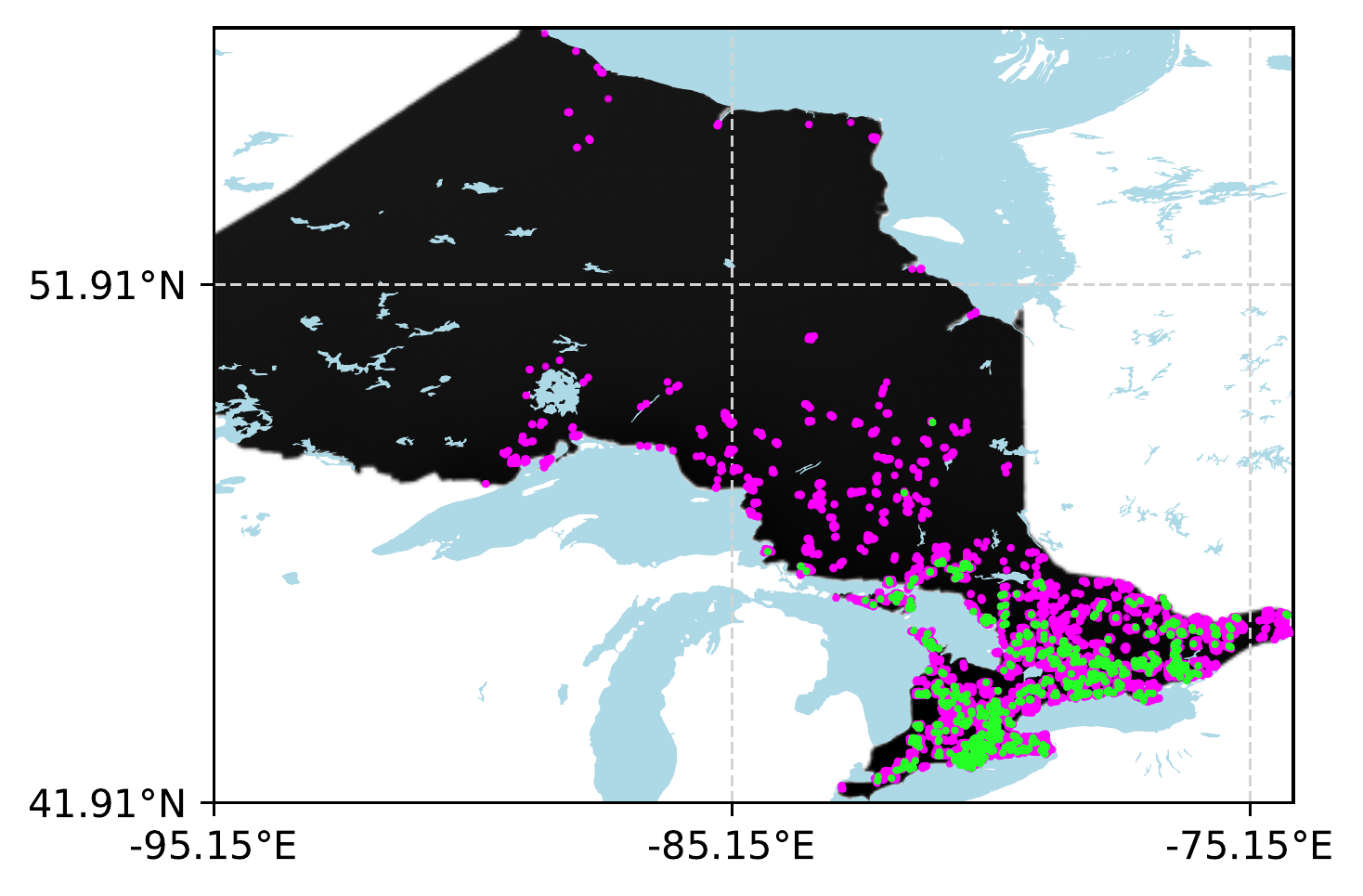}
        & \includegraphics[width=1.00\linewidth]{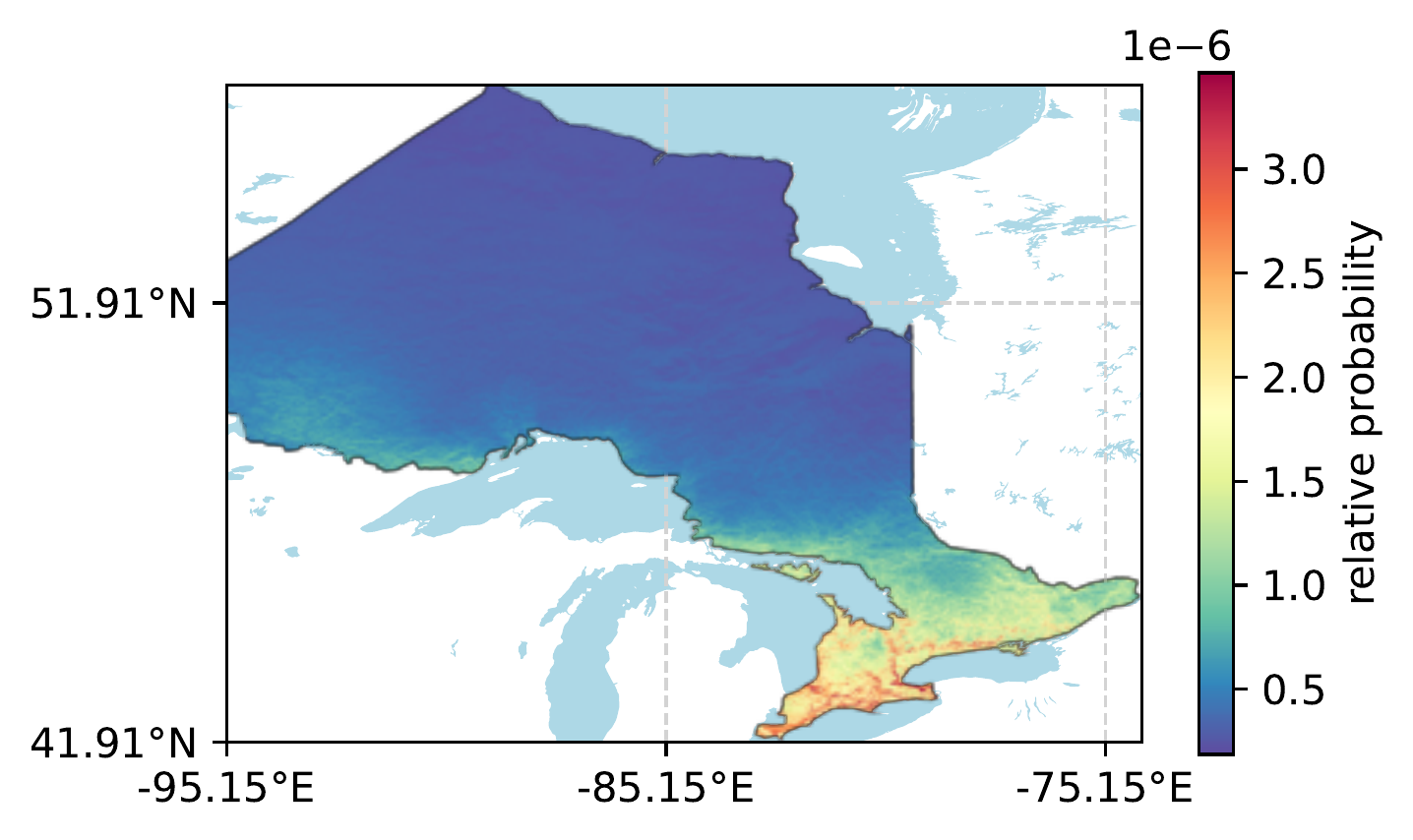} \\
    \end{tabular}
    \caption{Locations of species presences ({green}) and absences ({magenta}) based on PA observations, alongside predicted values from the reference DeepMaxent model across species can01 to can05}
    \label{fig:allmaps_withPA_1}
\end{figure}

\begin{figure}[h!]
    \centering
    \renewcommand{\arraystretch}{0.1}
    \begin{tabular}{>{\centering}m{1cm} m{6cm} m{6cm}}
        & \textbf{Presence/Absence} & \textbf{Estimated relative probability} \\
        \midrule
        \textbf{can06} \newline 
        & \includegraphics[width=1.00\linewidth]{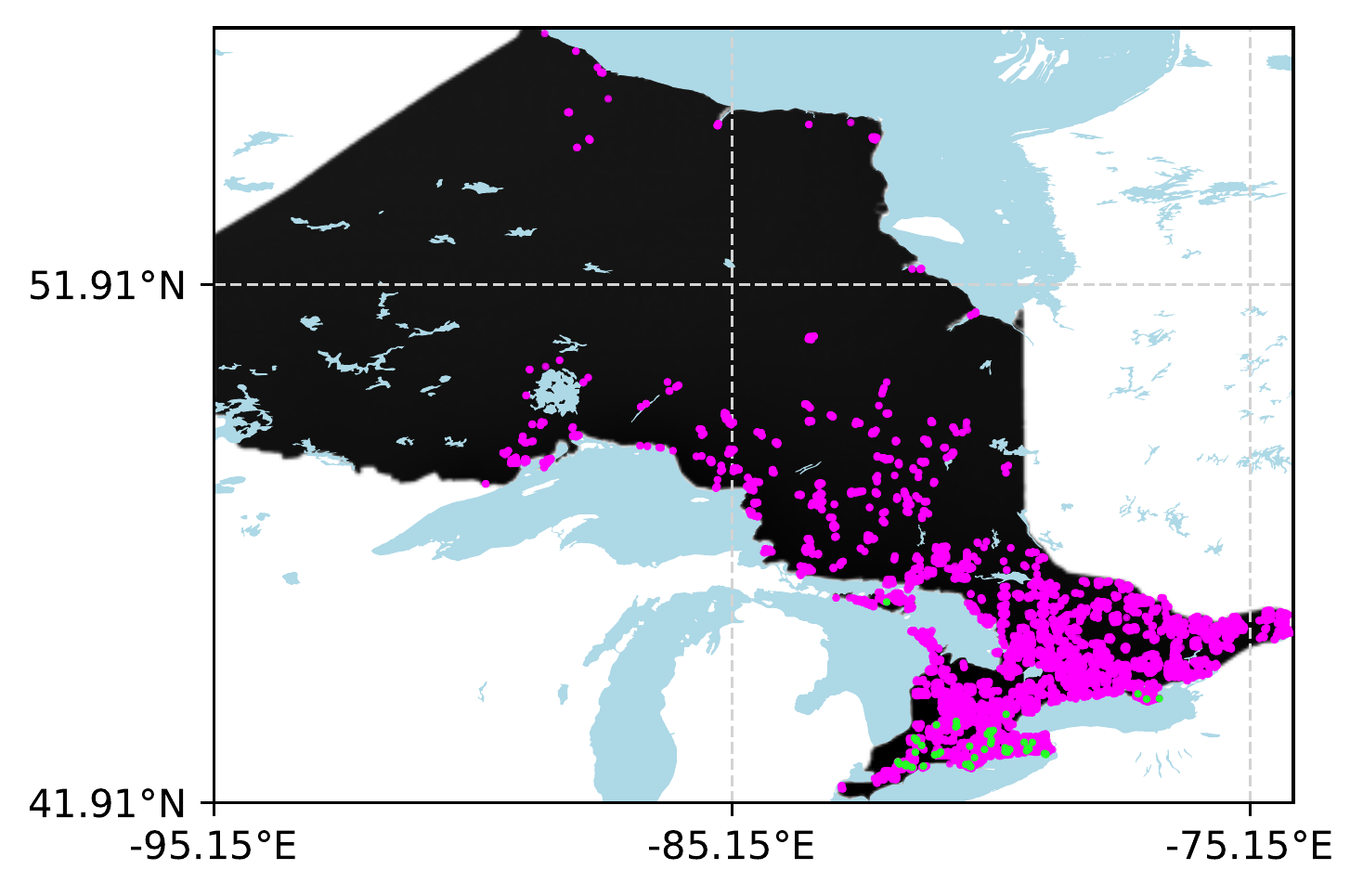}
        & \includegraphics[width=1.00\linewidth]{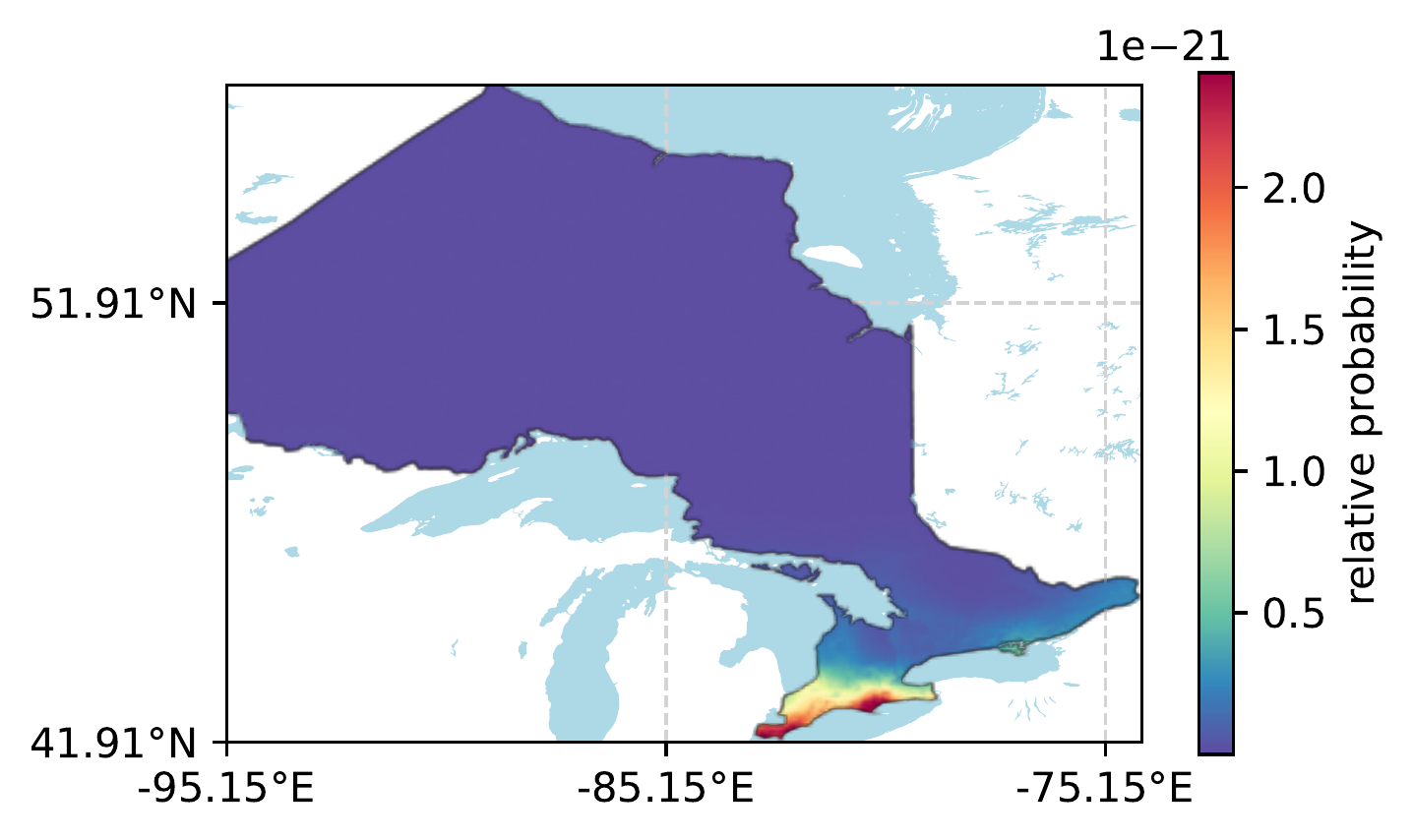} \\
        
        \textbf{can07} \newline 
        & \includegraphics[width=1.00\linewidth]{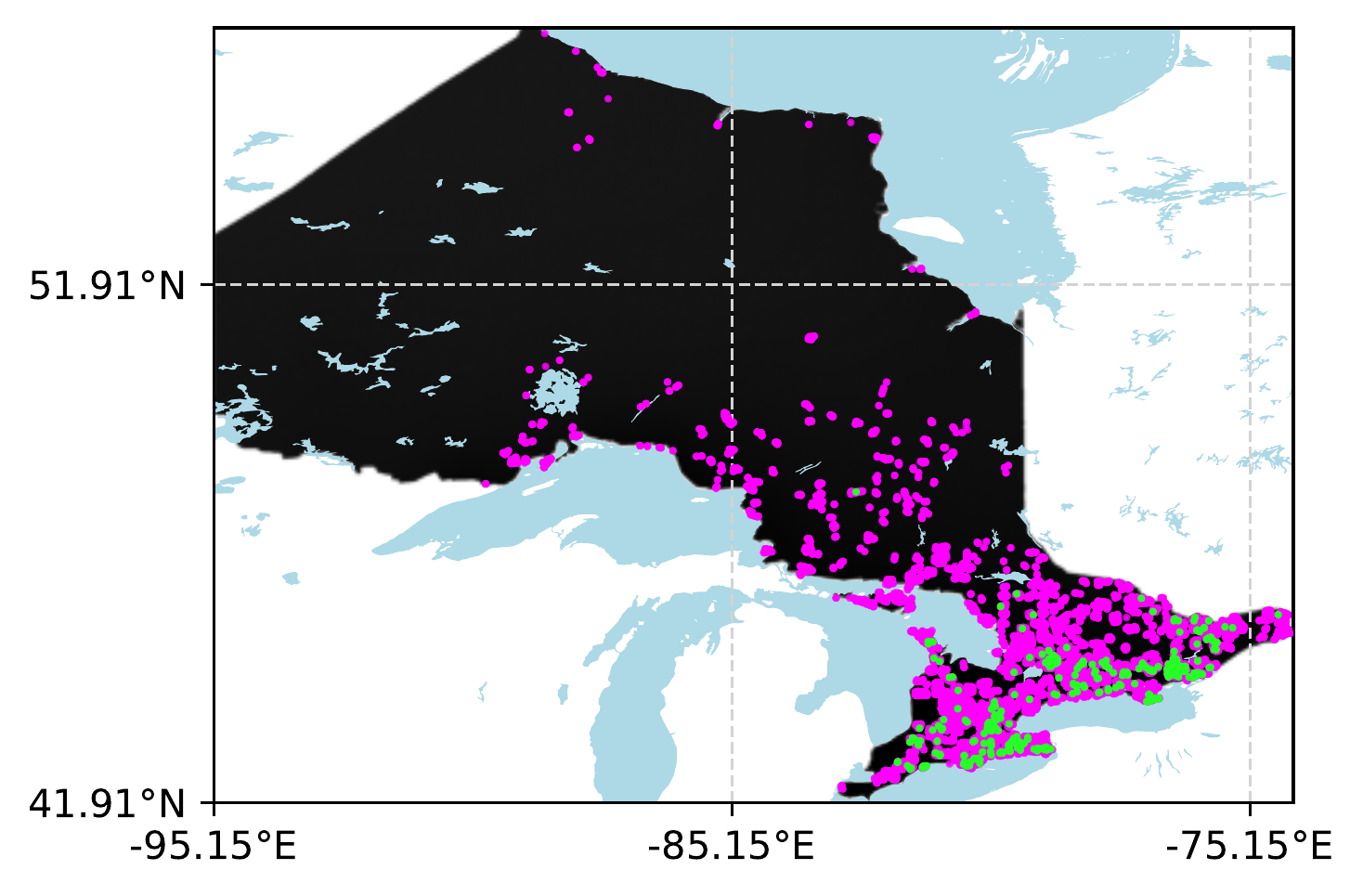}
        & \includegraphics[width=1.00\linewidth]{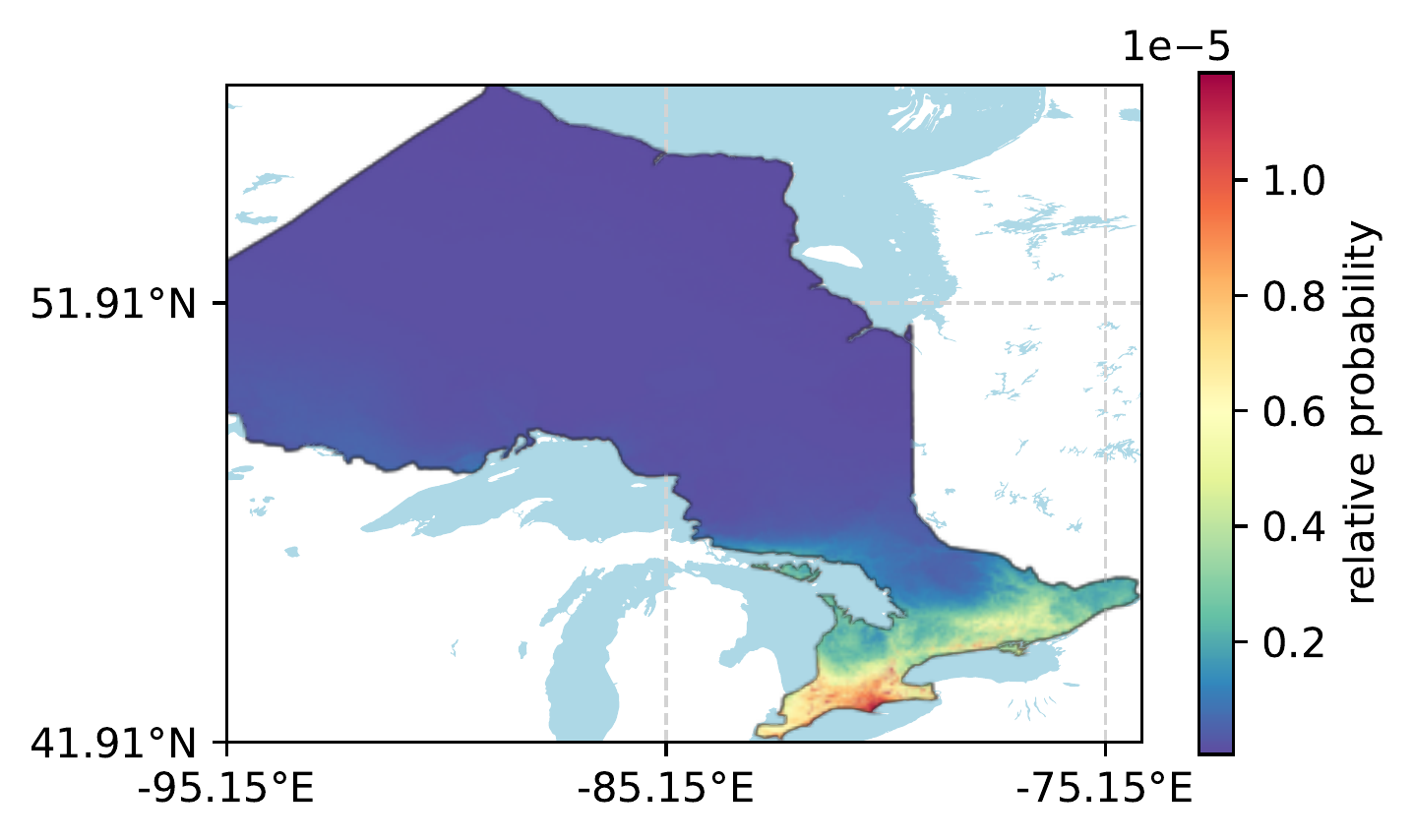} \\
        
        \textbf{can08} \newline 
        & \includegraphics[width=1.00\linewidth]{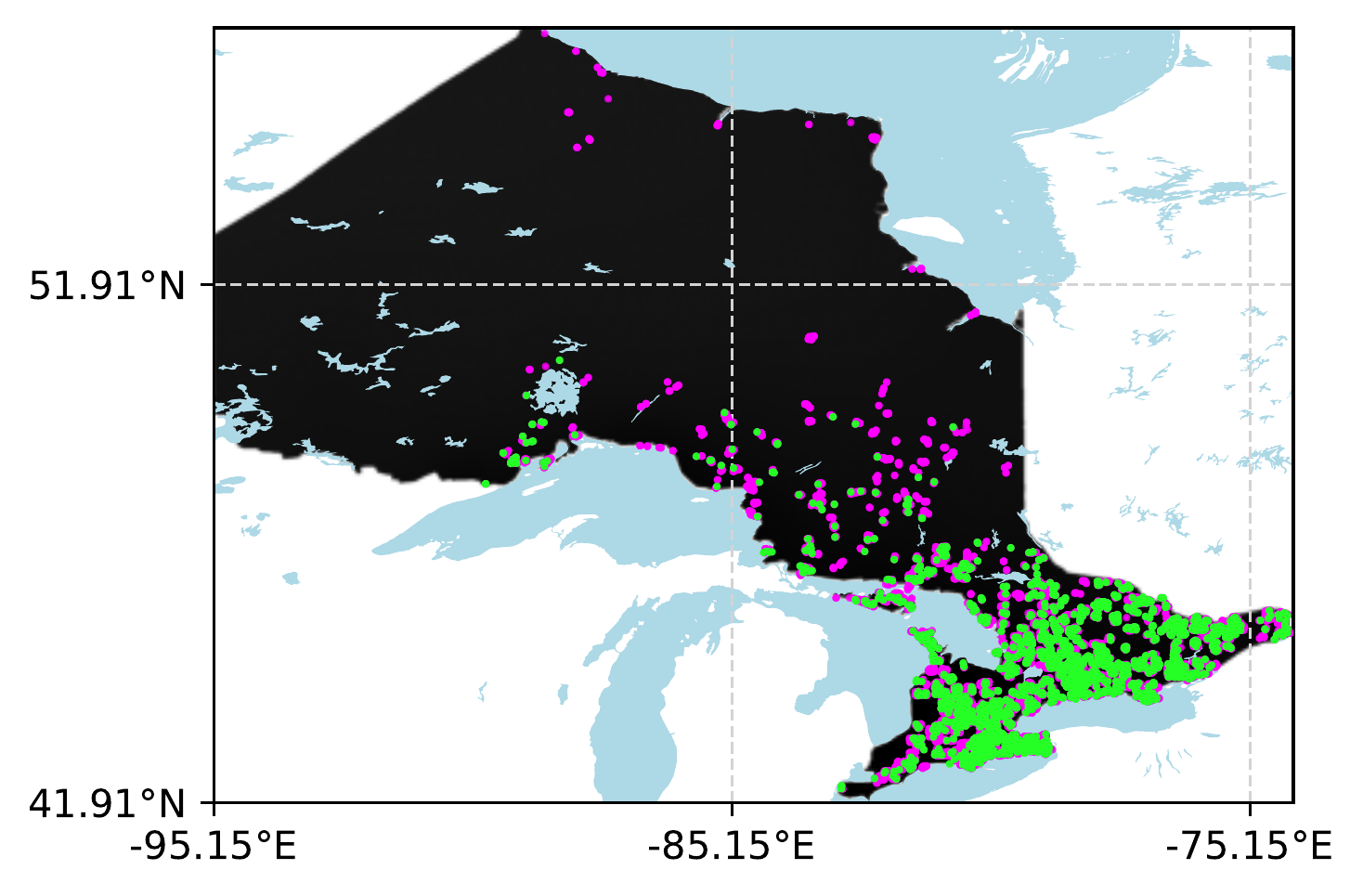}
        & \includegraphics[width=1.00\linewidth]{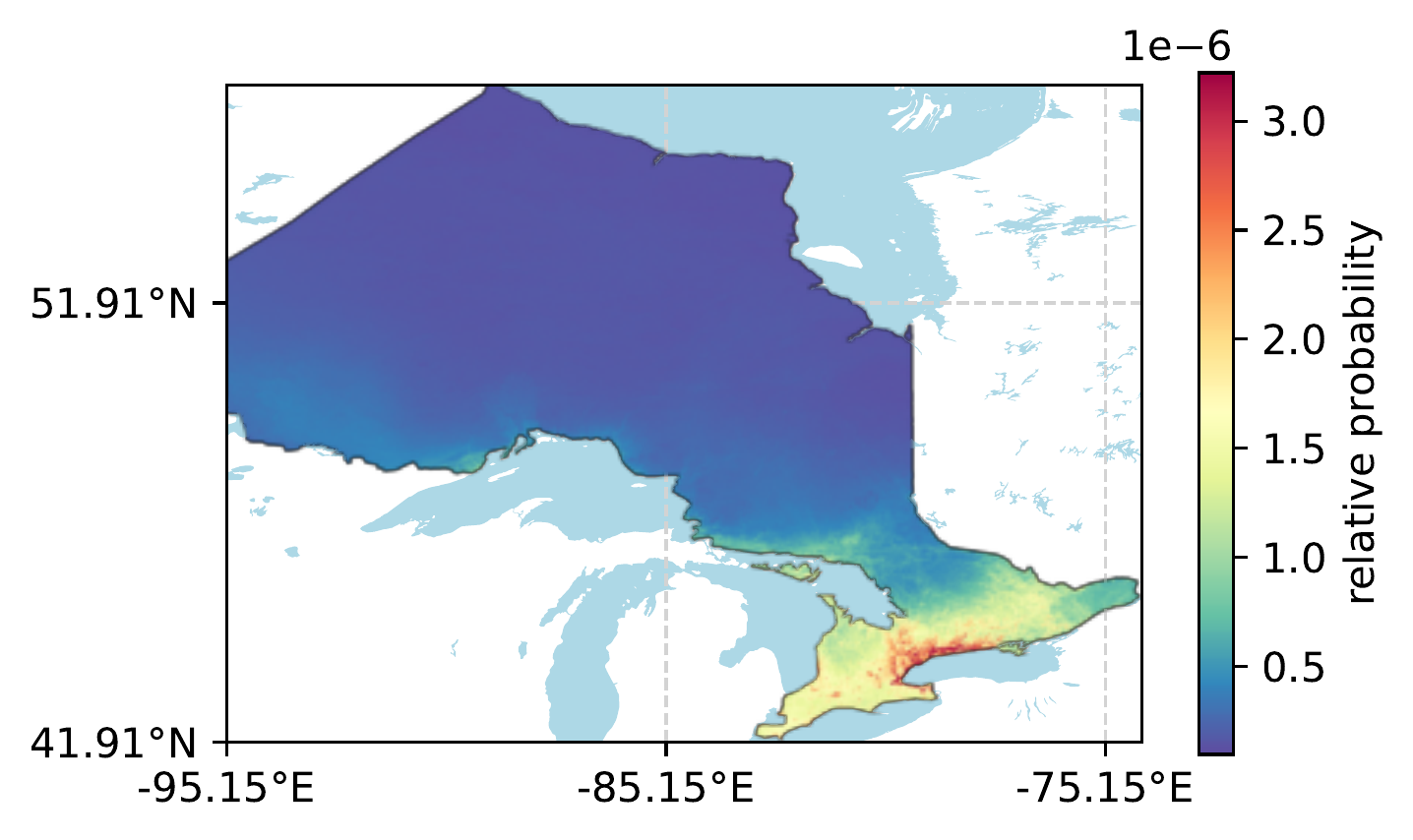} \\
        
        \textbf{can09} \newline
        & \includegraphics[width=1.00\linewidth]{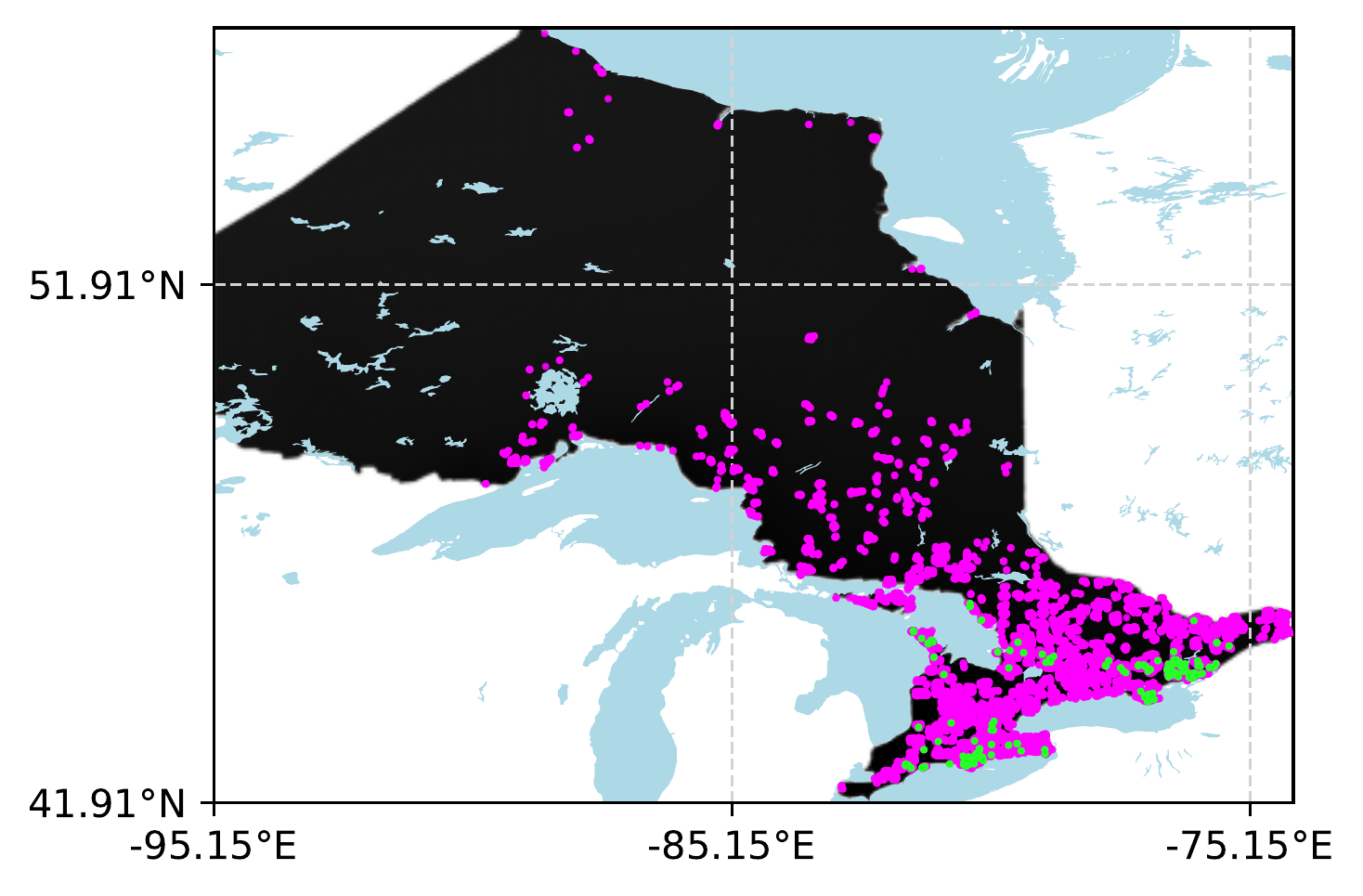}
        & \includegraphics[width=1.00\linewidth]{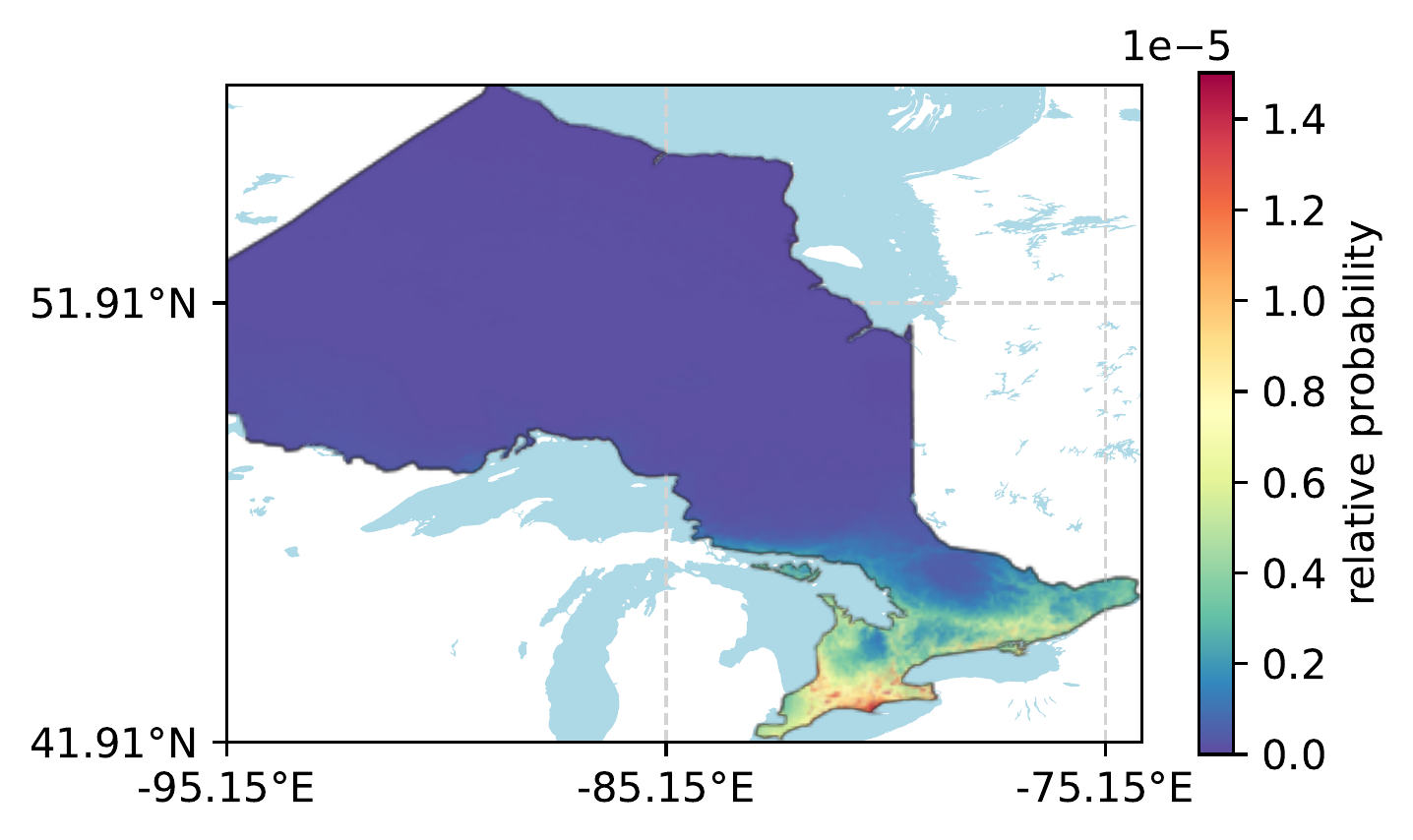} \\
        
        \textbf{can10} \newline
        & \includegraphics[width=1.00\linewidth]{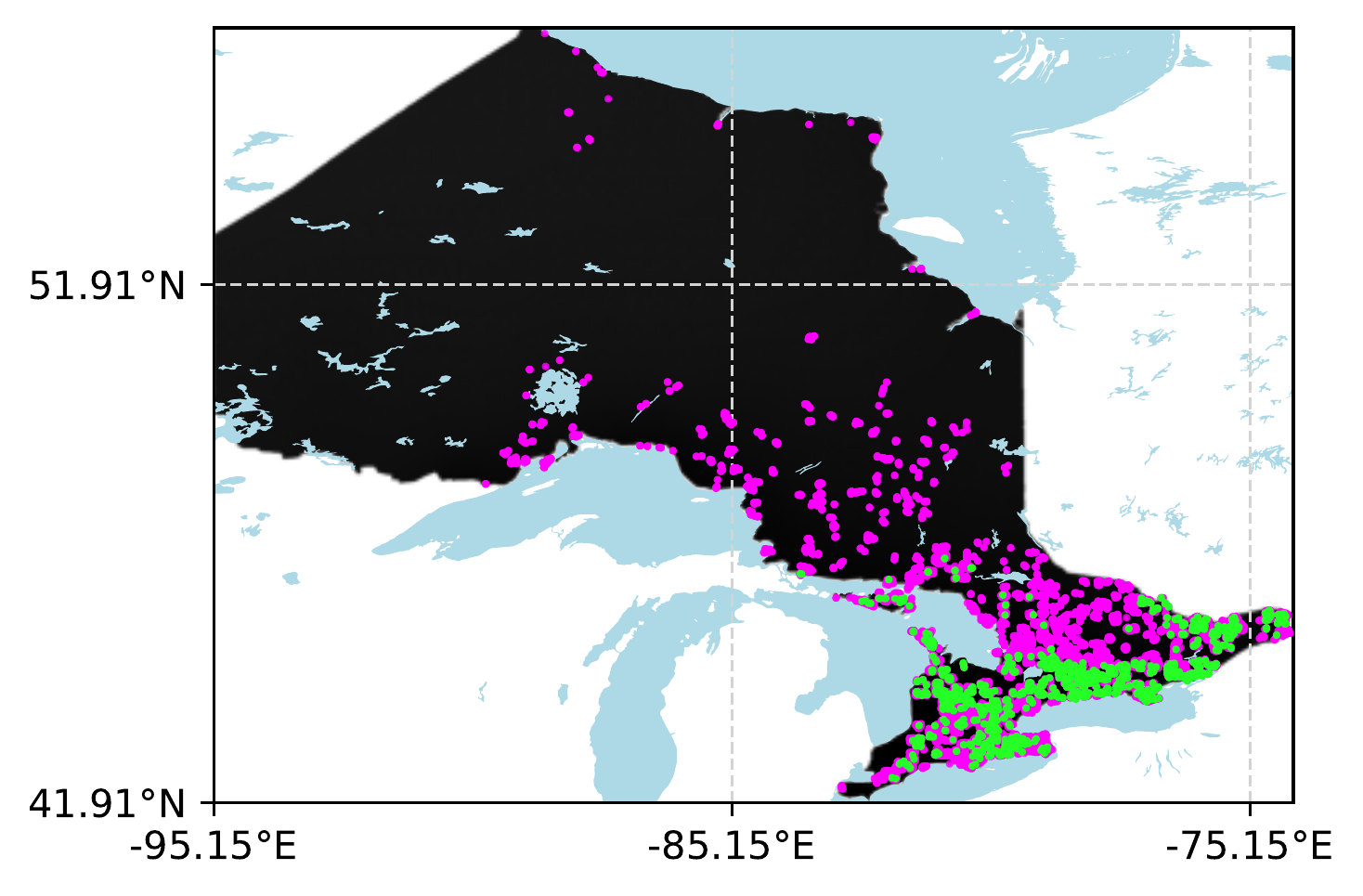}
        & \includegraphics[width=1.00\linewidth]{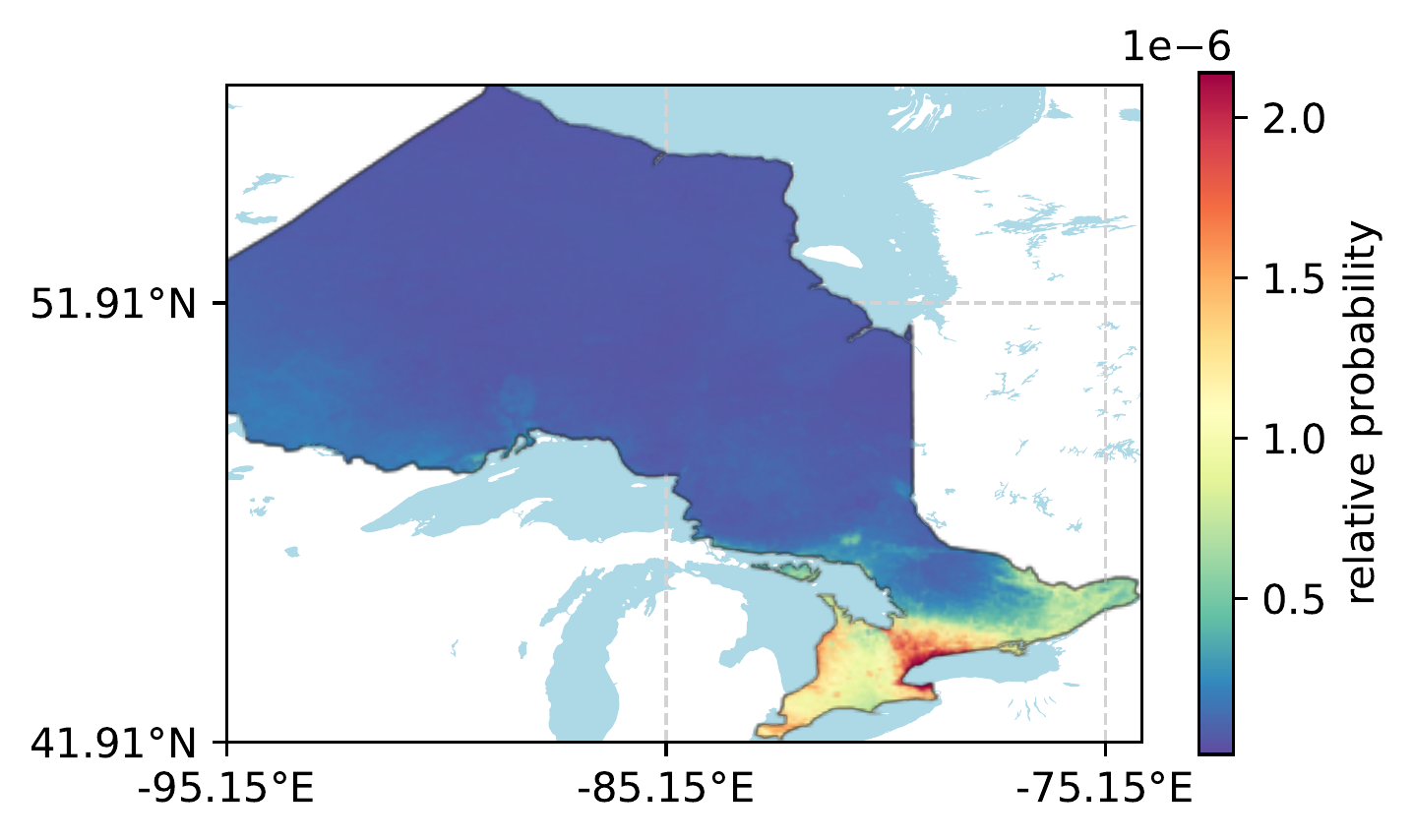} \\
    \end{tabular}
    \caption{Locations of species presences ({green}) and absences ({magenta}) based on PA observations, alongside predicted values from the reference DeepMaxent model across species can06 to can10}
    \label{fig:allmaps_withPA_2}
\end{figure}

\begin{figure}[h!]
    \centering
    \renewcommand{\arraystretch}{0.1}
    \begin{tabular}{>{\centering}m{1cm} m{6cm} m{6cm}}
        & \textbf{Presence/Absence} & \textbf{Estimated relative probability} \\
        \midrule
        \textbf{can11} \newline 
        & \includegraphics[width=1.00\linewidth]{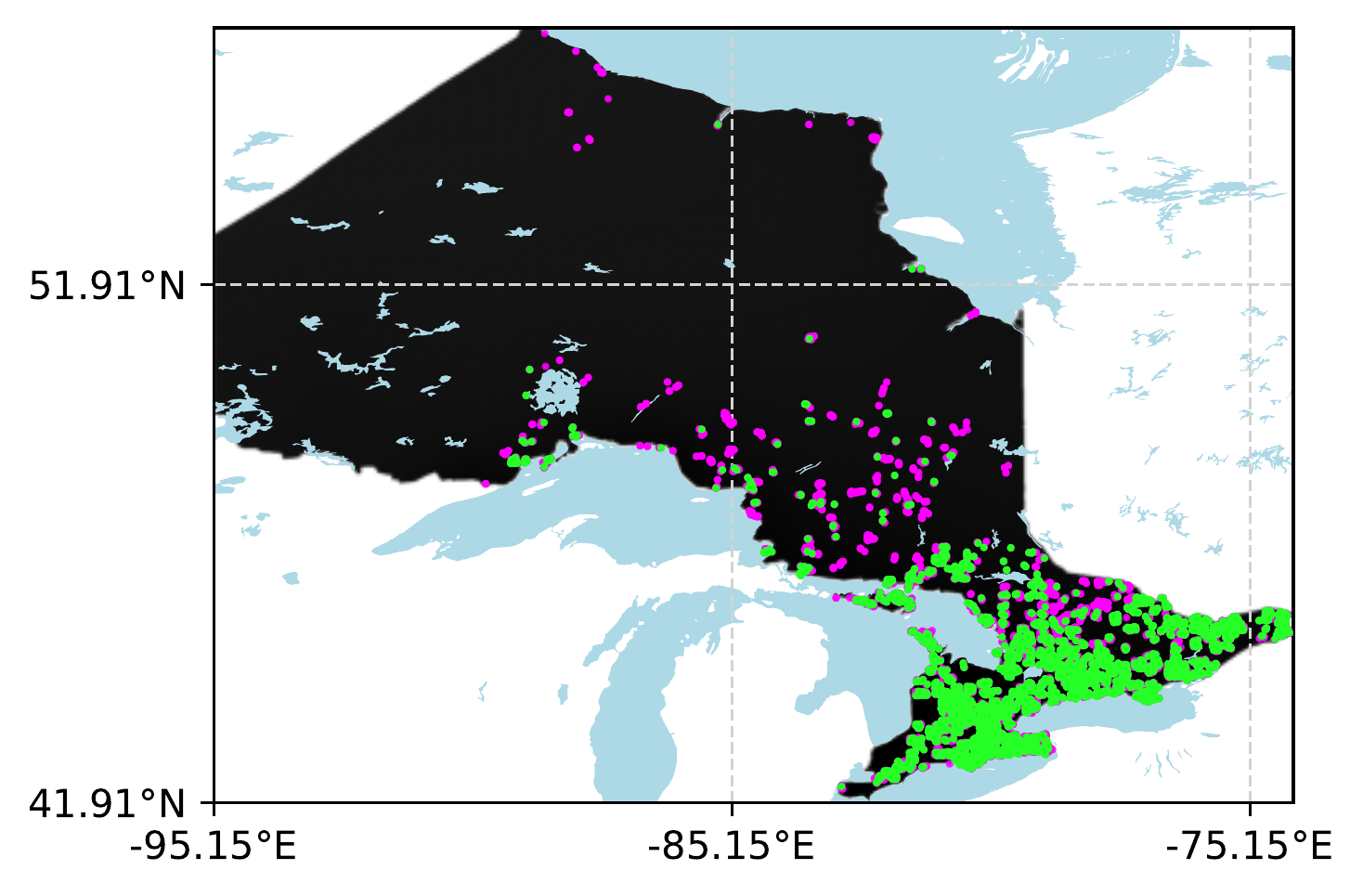}
        & \includegraphics[width=1.00\linewidth]{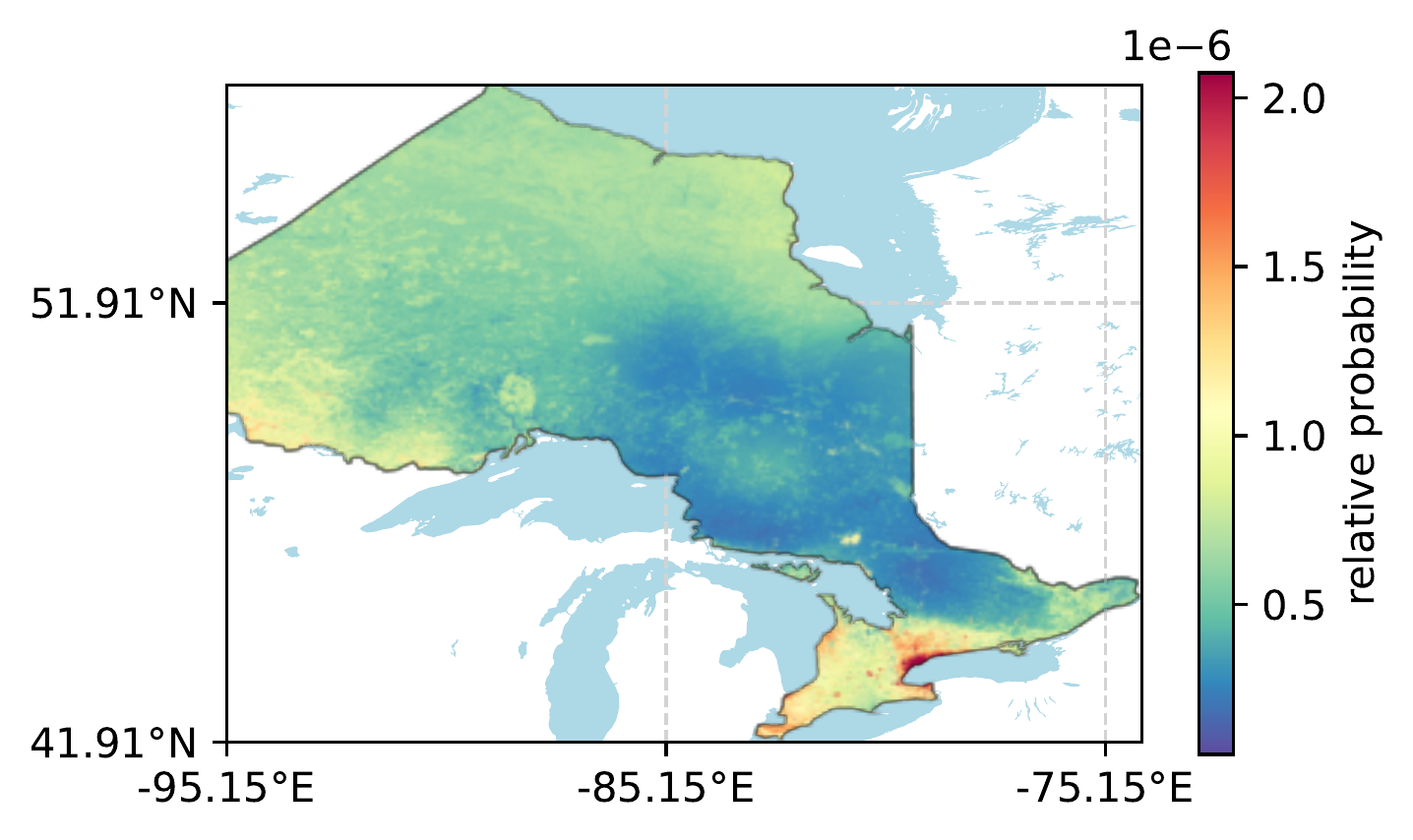} \\
        
        \textbf{can12} \newline 
        & \includegraphics[width=1.00\linewidth]{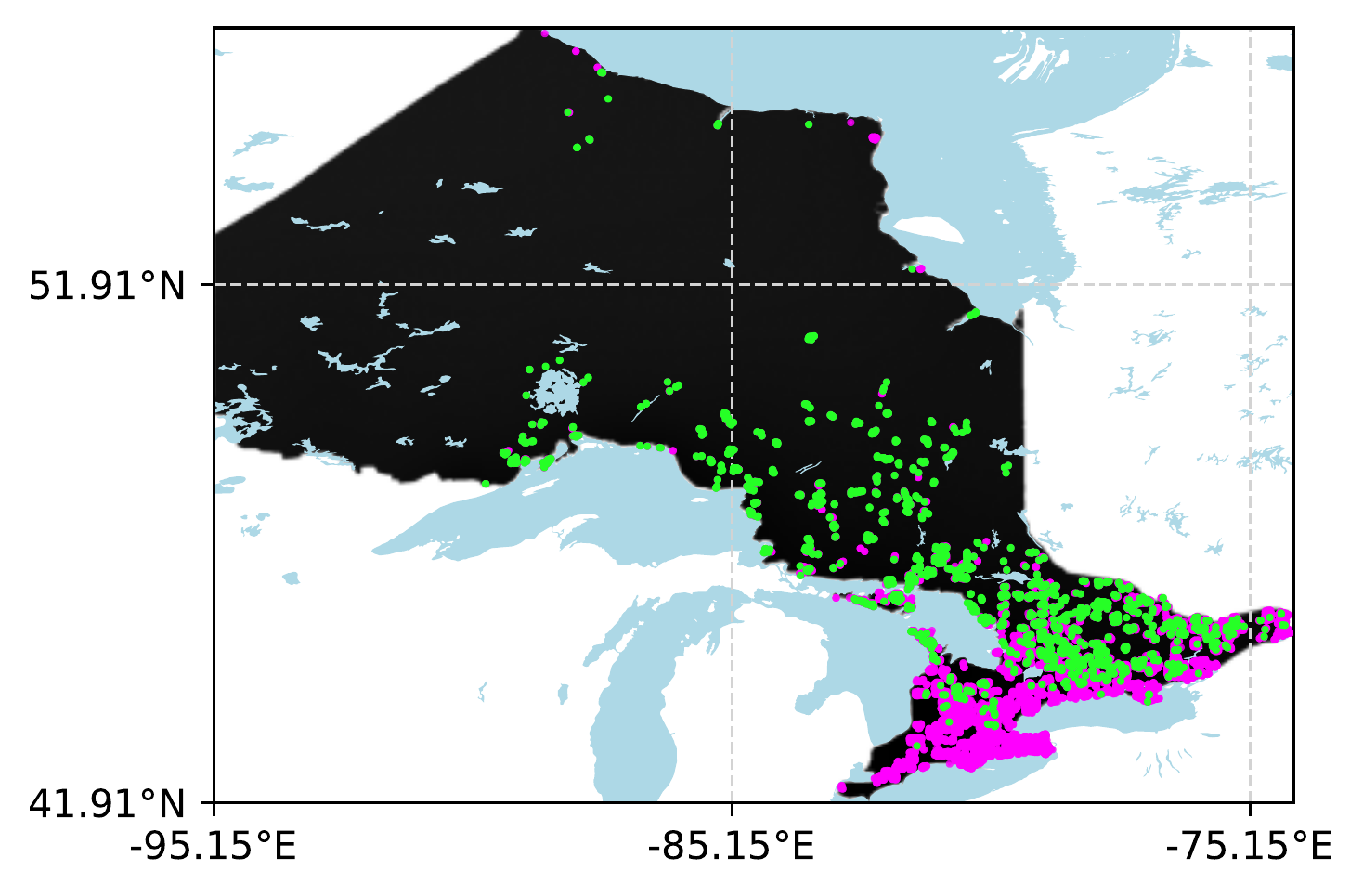}
        & \includegraphics[width=1.00\linewidth]{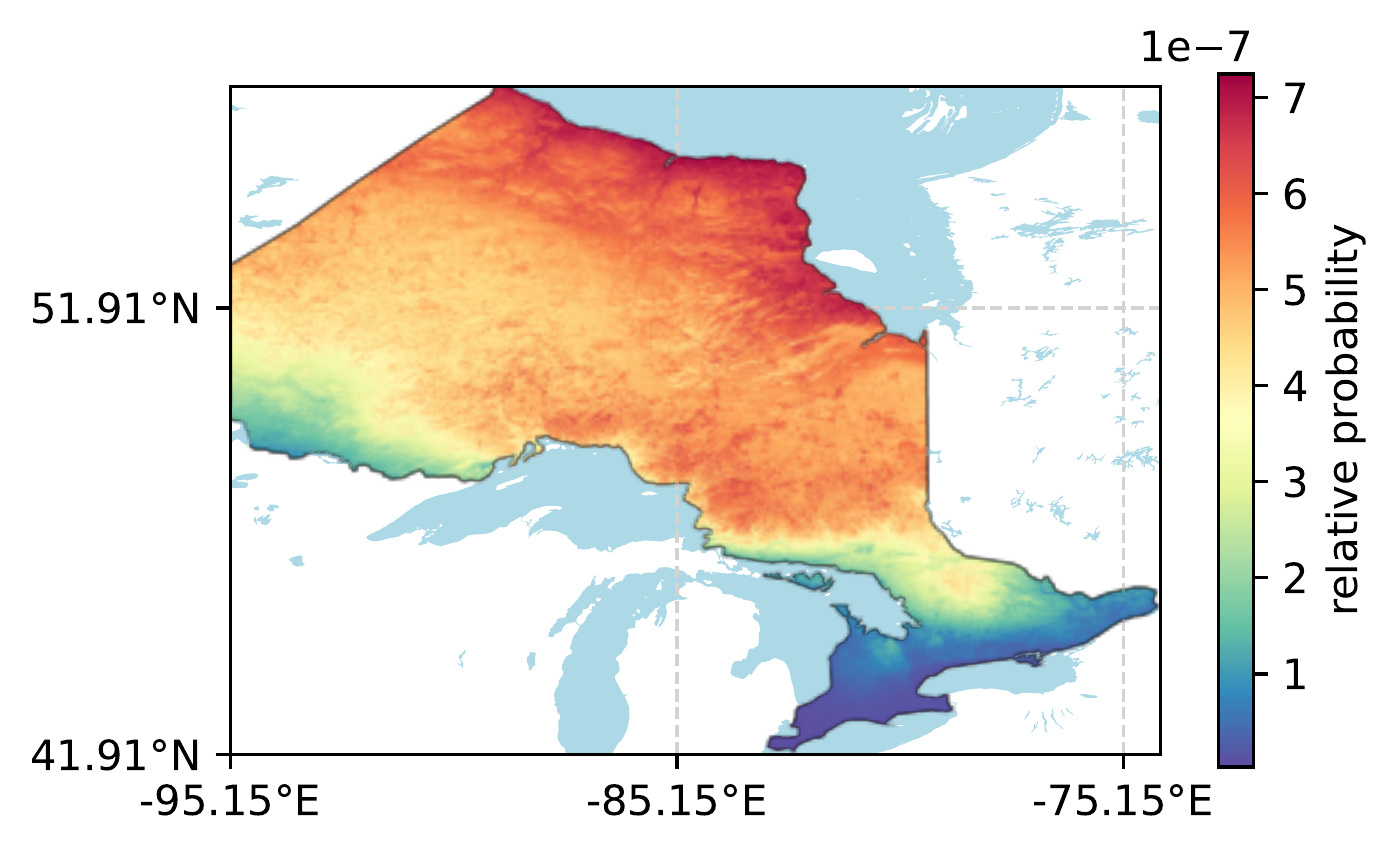} \\
        
        \textbf{can13} \newline 
        & \includegraphics[width=1.00\linewidth]{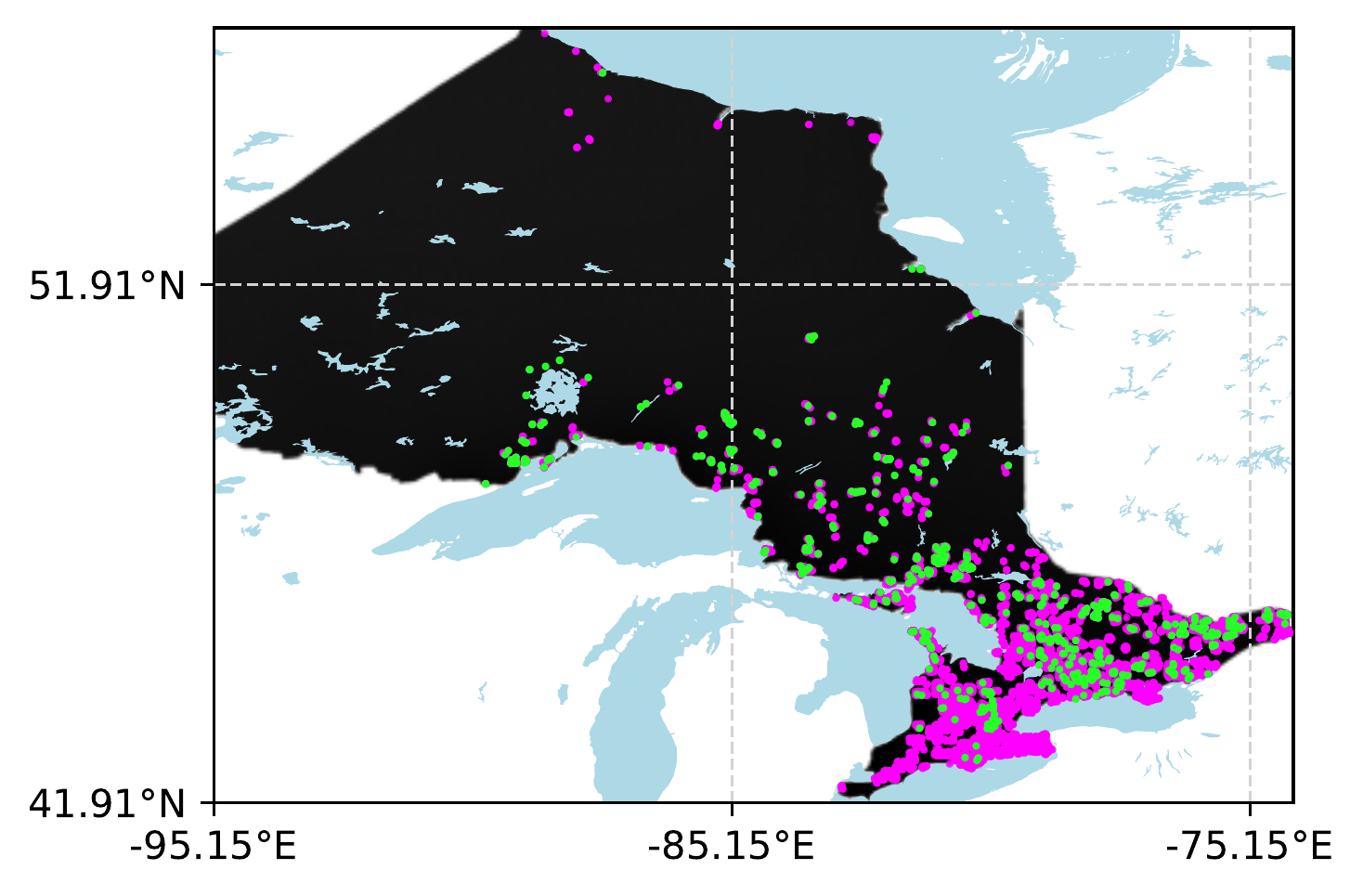}
        & \includegraphics[width=1.00\linewidth]{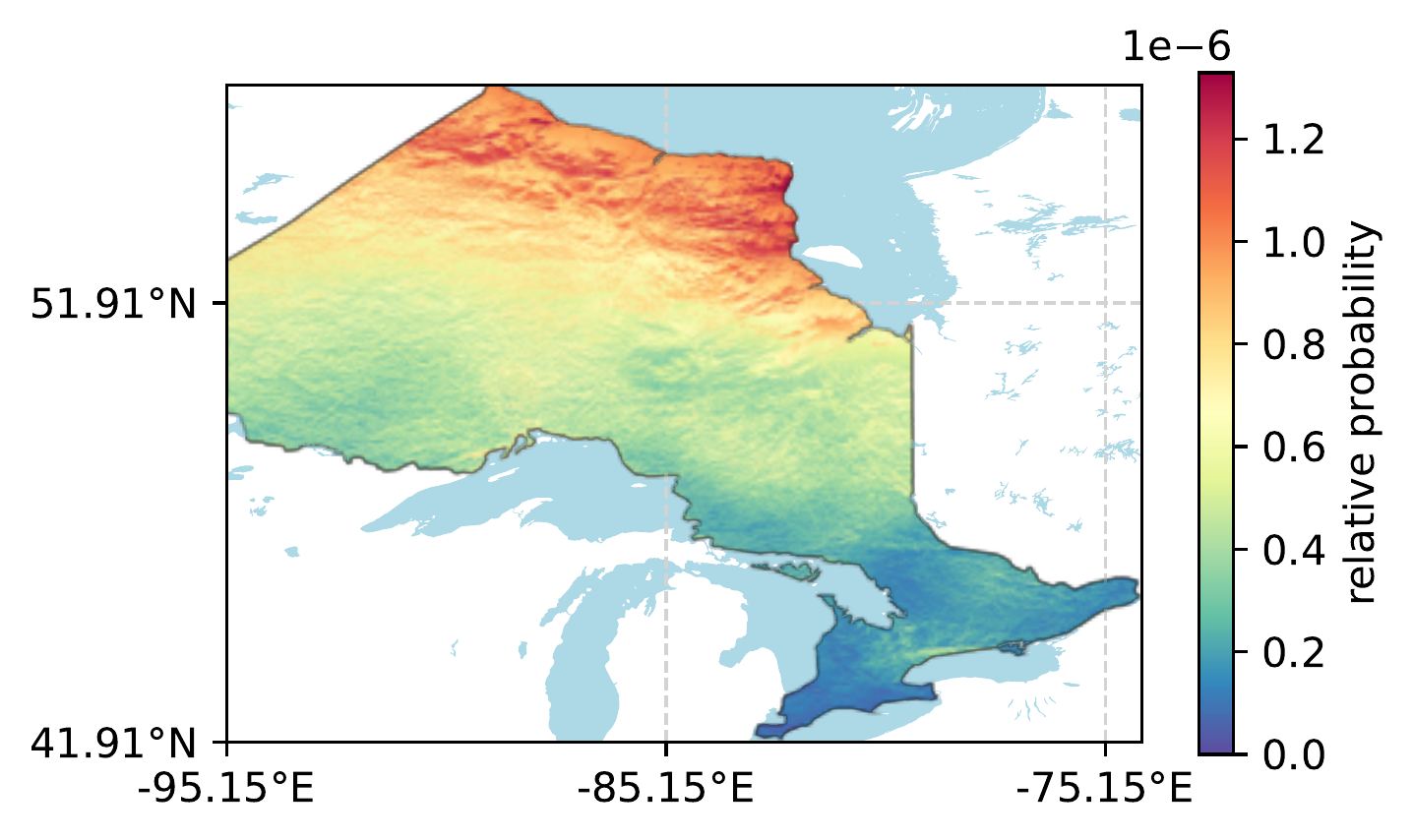} \\
        
        \textbf{can14} \newline
        & \includegraphics[width=1.00\linewidth]{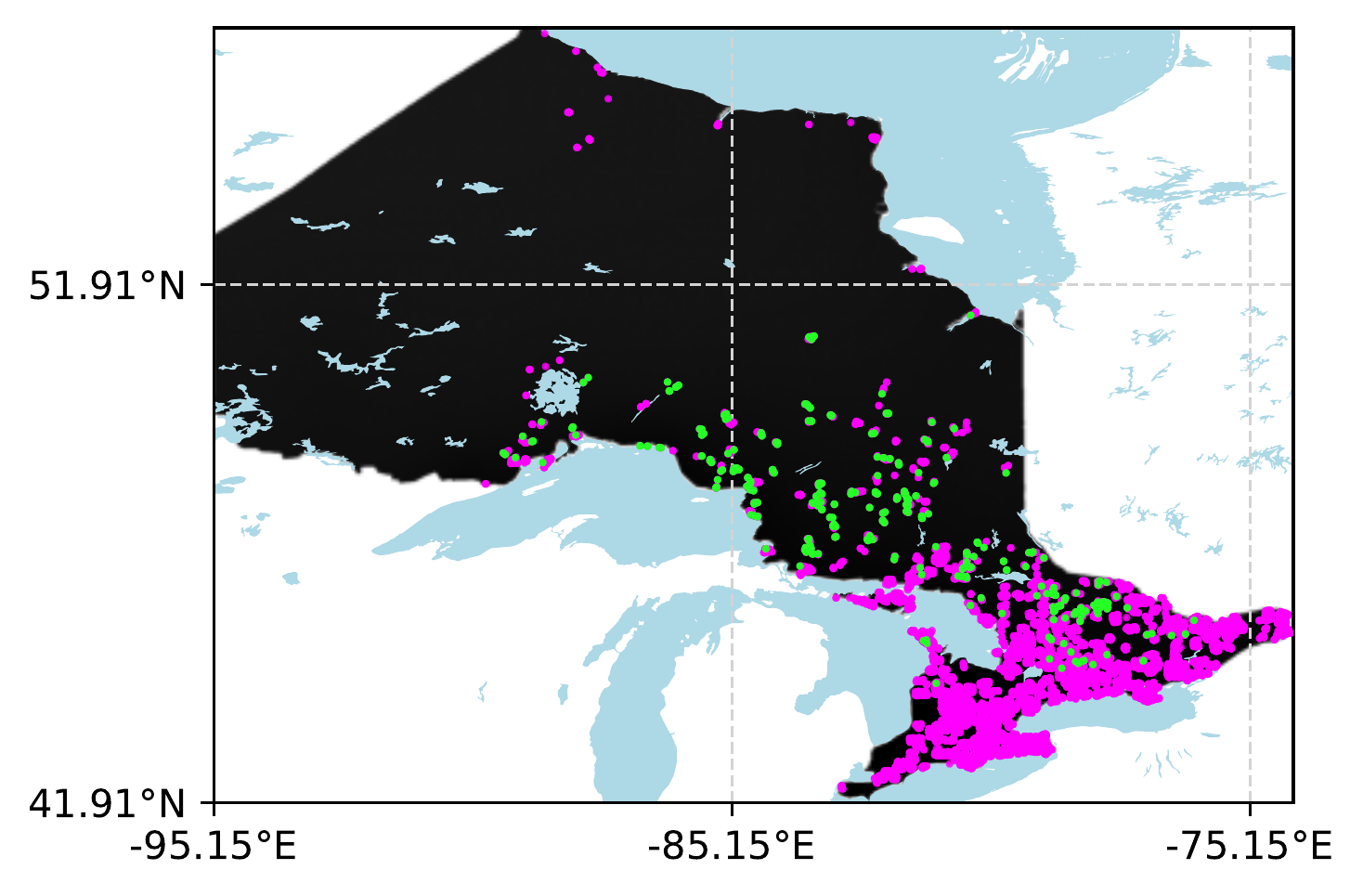}
        & \includegraphics[width=1.00\linewidth]{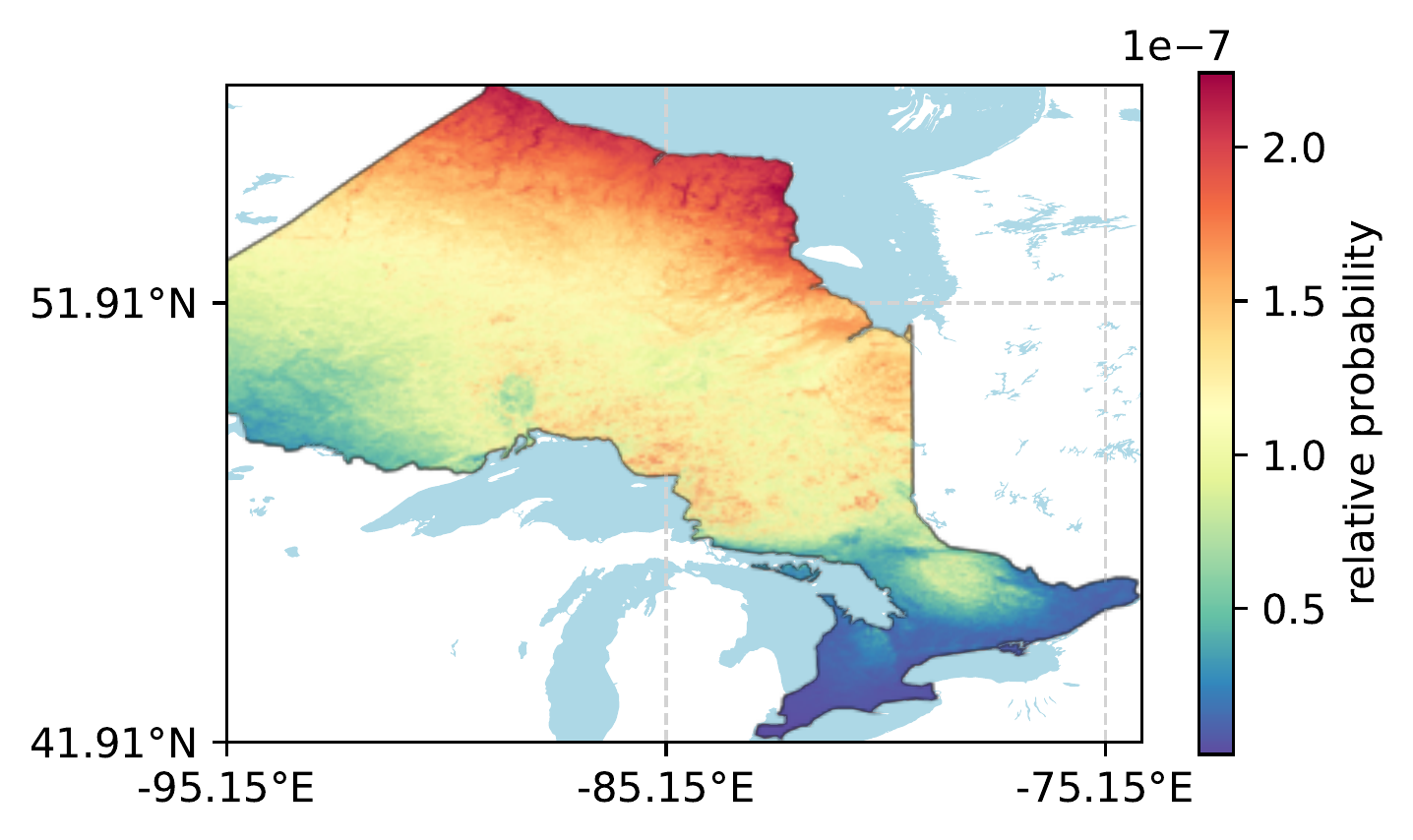} \\
        
        \textbf{can15} \newline
        & \includegraphics[width=1.00\linewidth]{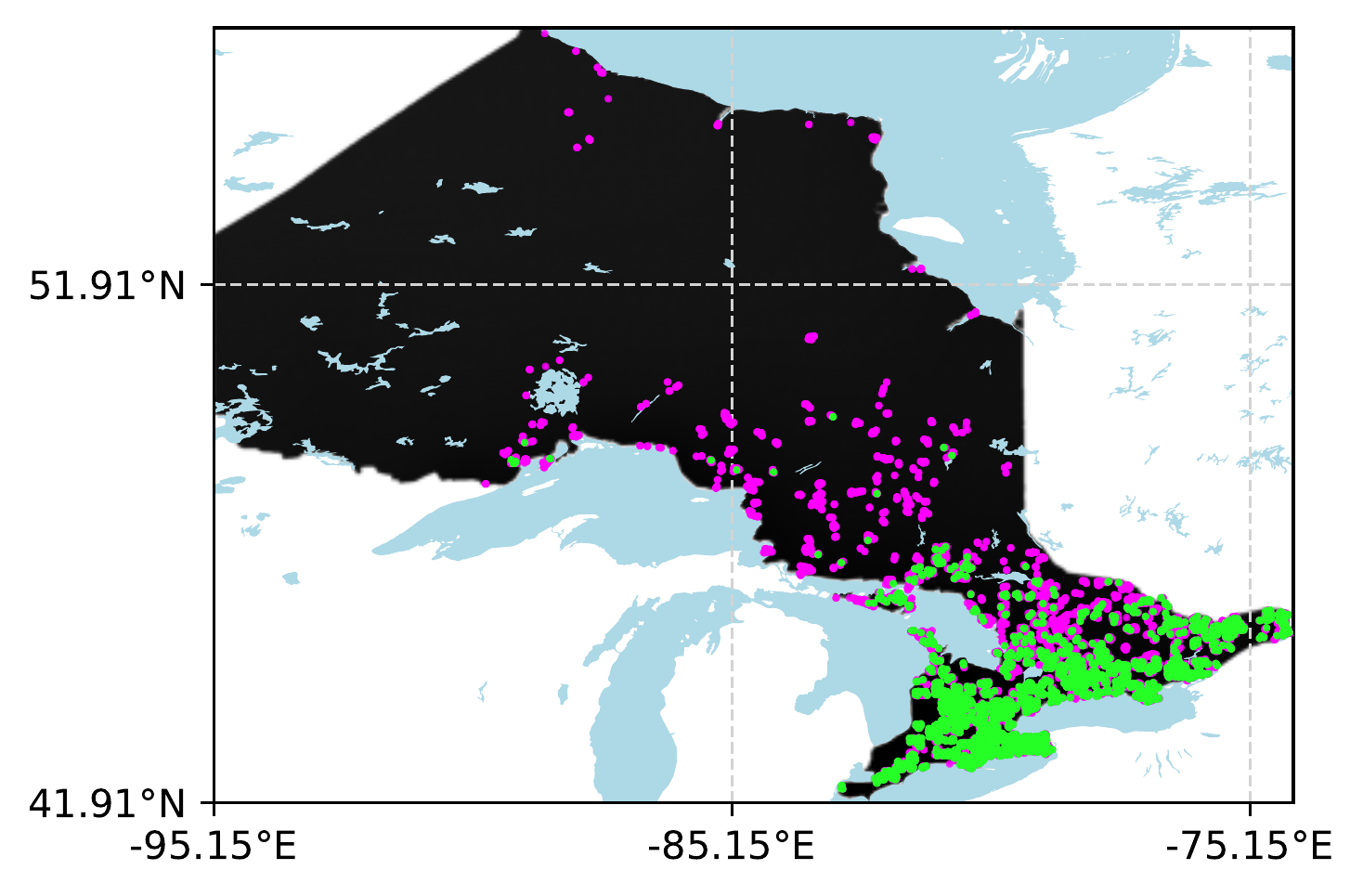}
        & \includegraphics[width=1.00\linewidth]{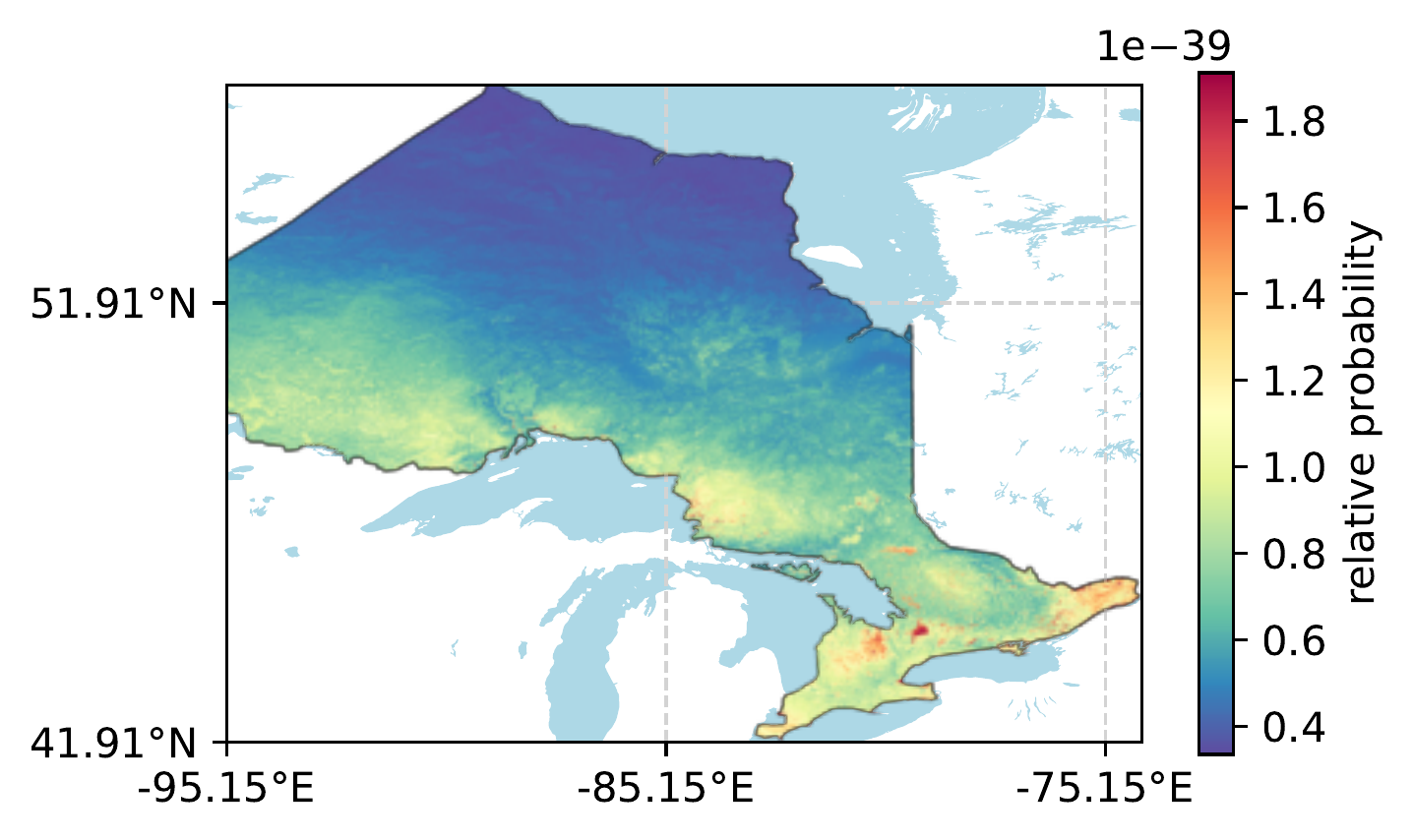} \\
    \end{tabular}
    \caption{Locations of species presences ({green}) and absences ({magenta}) based on PA observations, alongside predicted values from the reference DeepMaxent model across species can11 to can15}
    \label{fig:allmaps_withPA_3}
\end{figure}

\begin{figure}[h!]
    \centering
    \renewcommand{\arraystretch}{0.1}
    \begin{tabular}{>{\centering}m{4cm} m{6cm} m{6cm}}
        & \textbf{Presence/Absence} & \textbf{Estimated relative probability} \\
        \midrule
        \textbf{can16} \newline 
        & \includegraphics[width=1.00\linewidth]{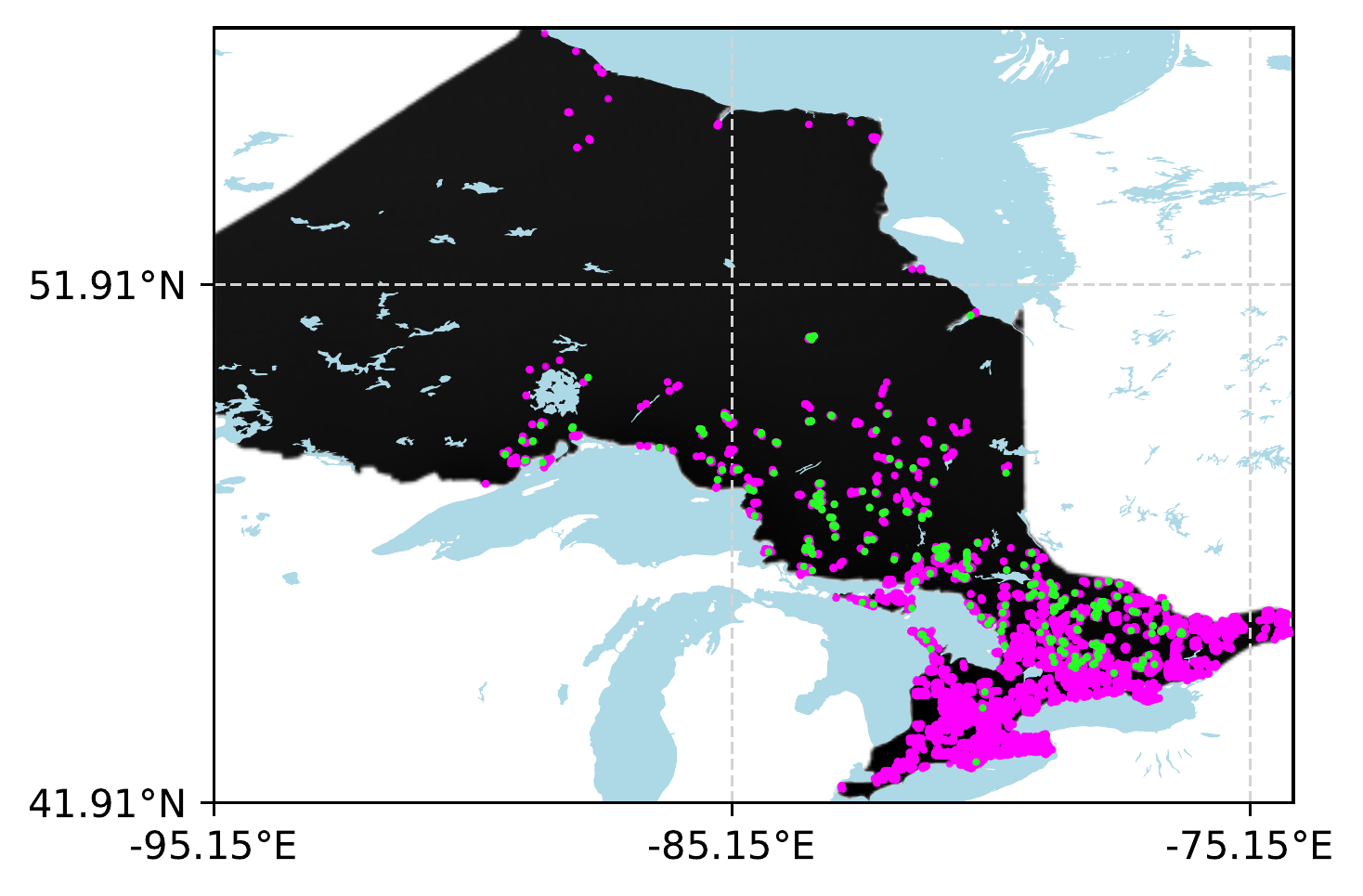}
        & \includegraphics[width=1.00\linewidth]{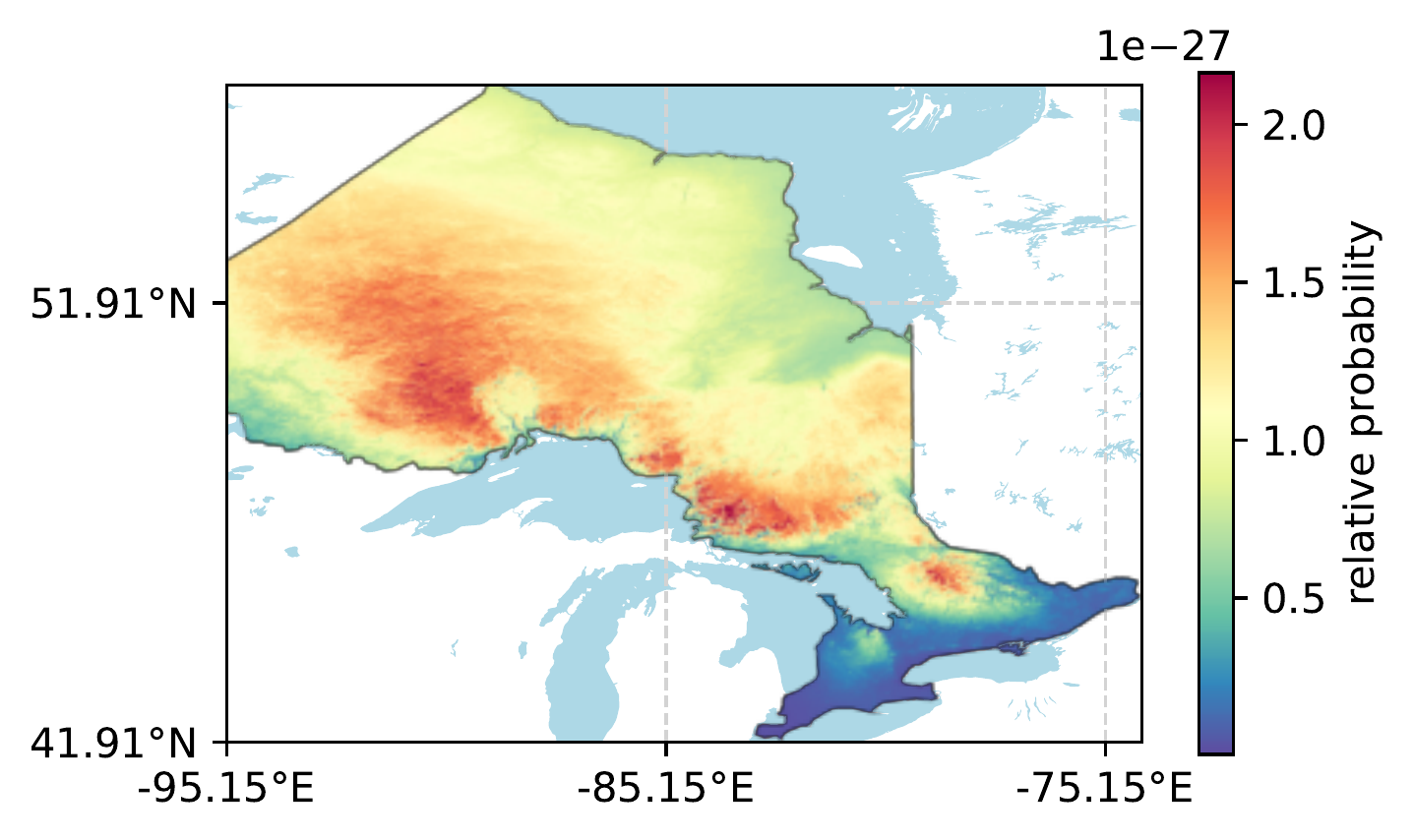} \\
        
        \textbf{can17} \newline 
        & \includegraphics[width=1.00\linewidth]{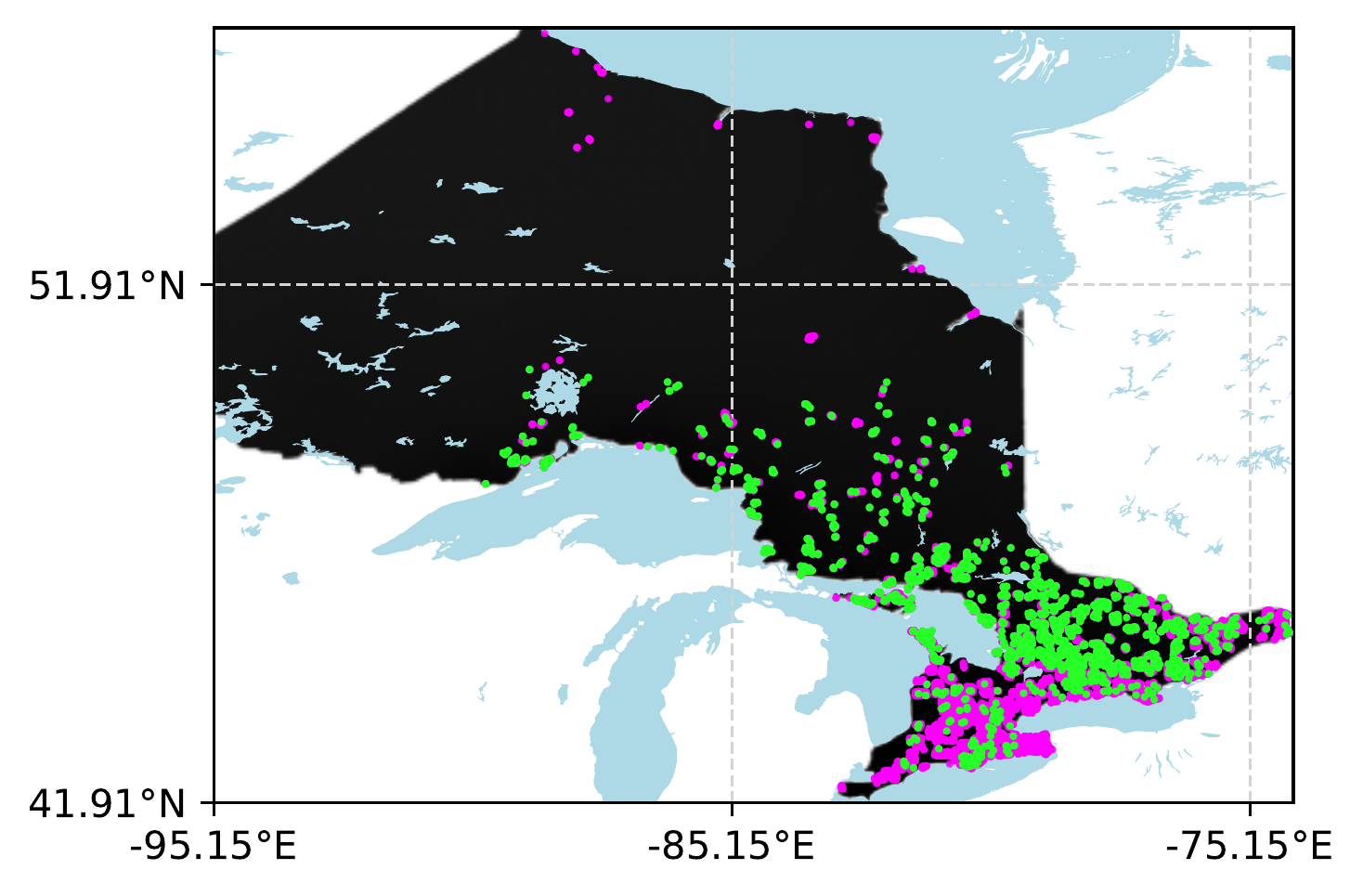}
        & \includegraphics[width=1.00\linewidth]{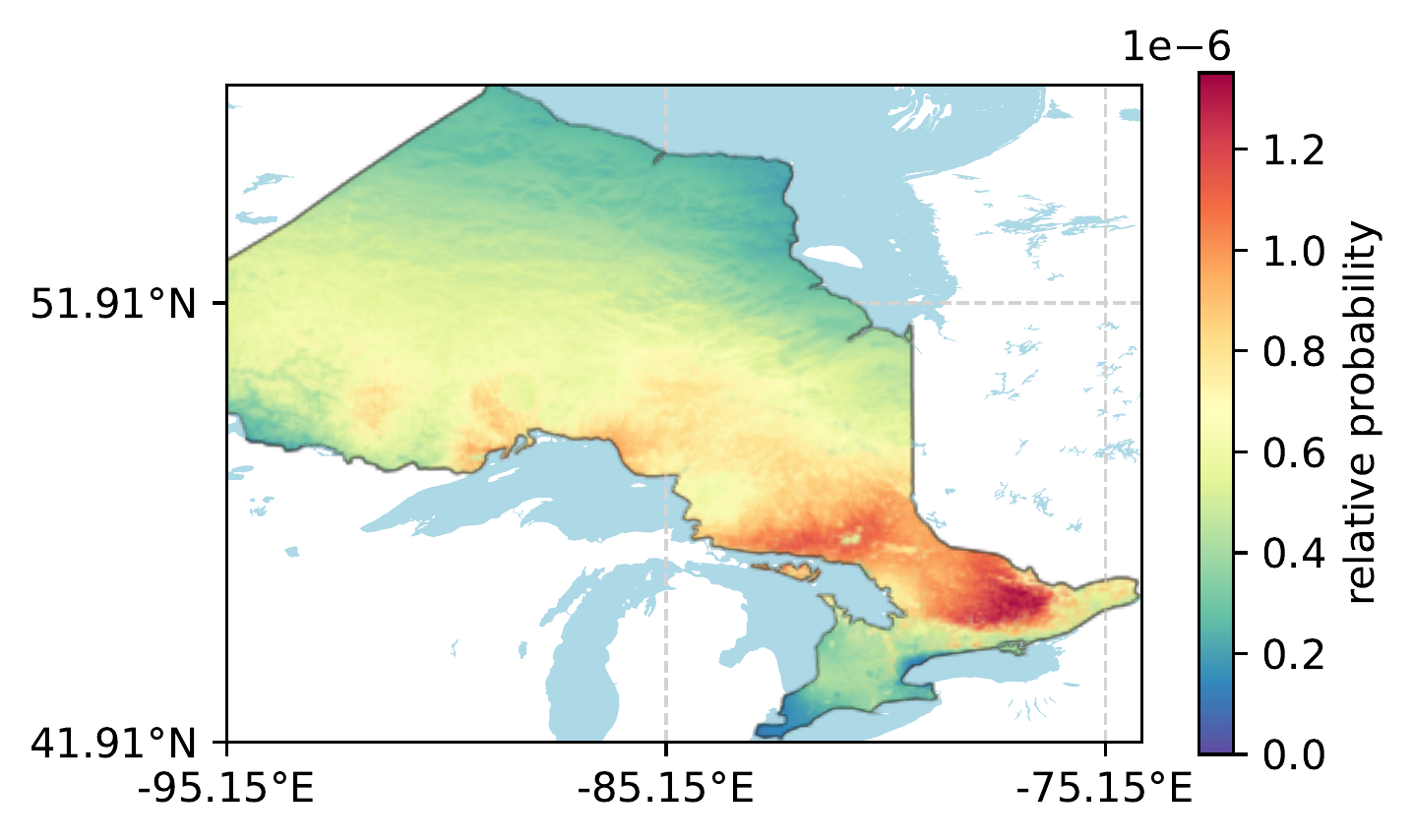} \\
        
        \textbf{can18} \newline 
        & \includegraphics[width=1.00\linewidth]{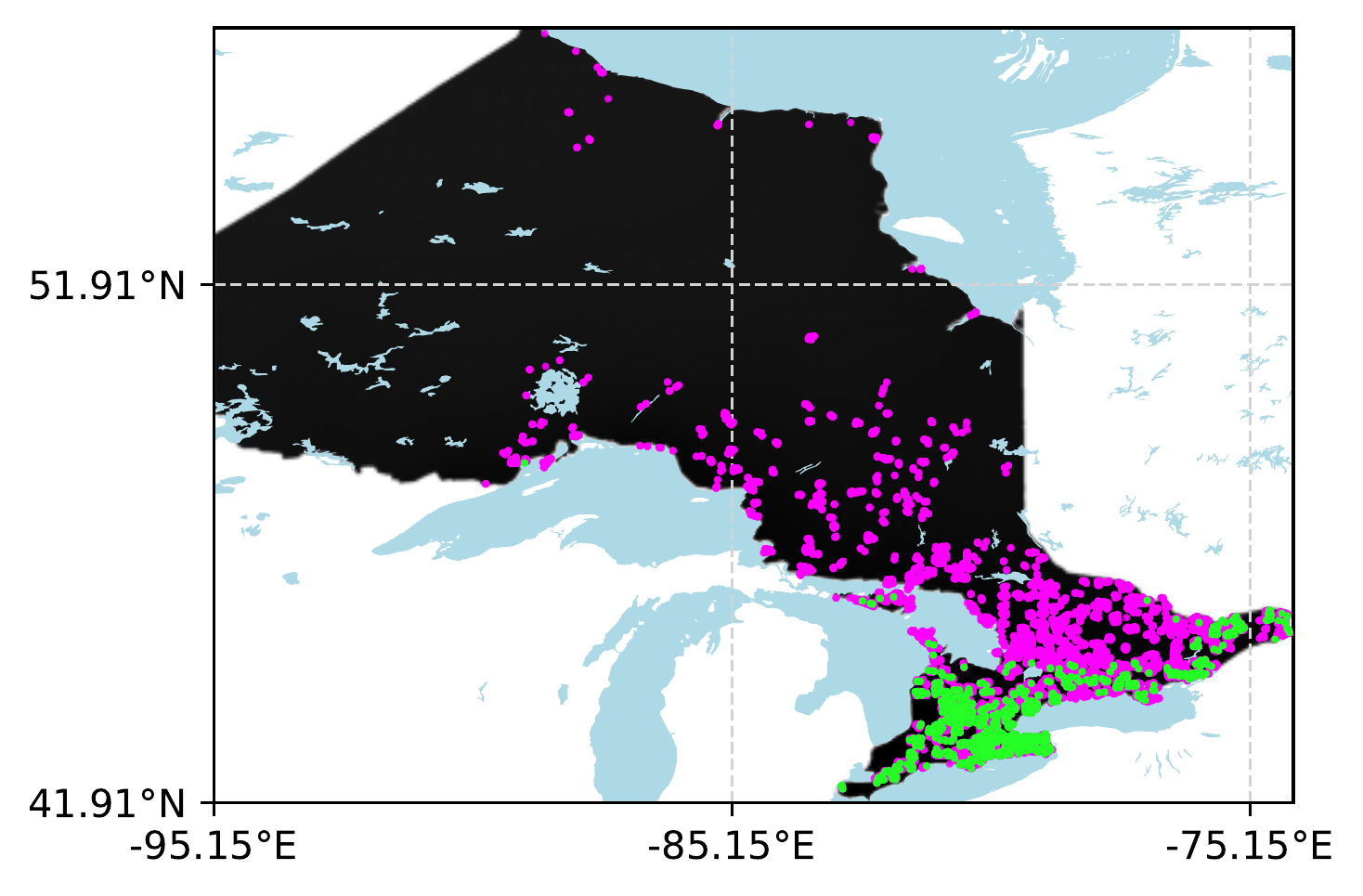}
        & \includegraphics[width=1.00\linewidth]{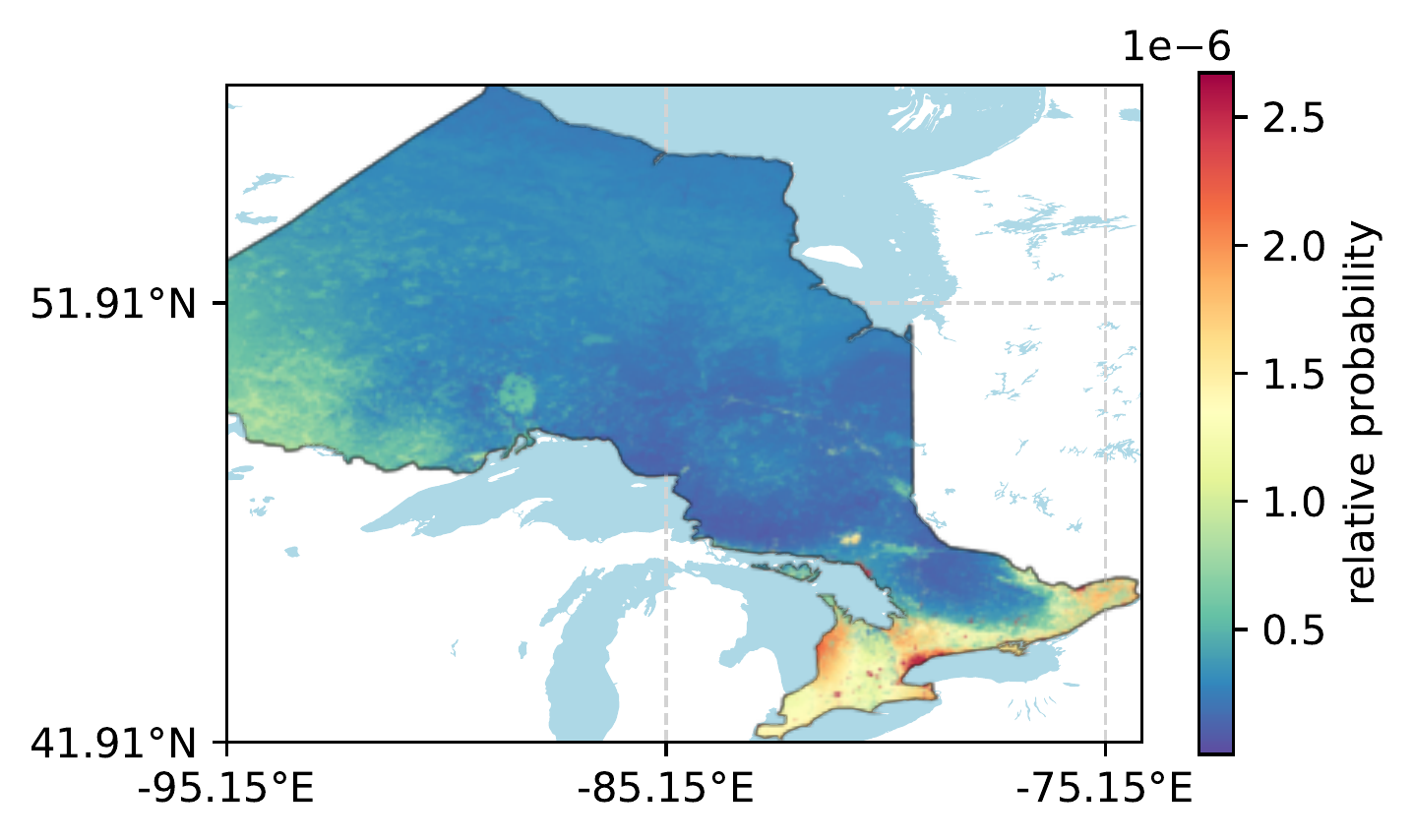} \\
        
        \textbf{can19} \newline
        & \includegraphics[width=1.00\linewidth]{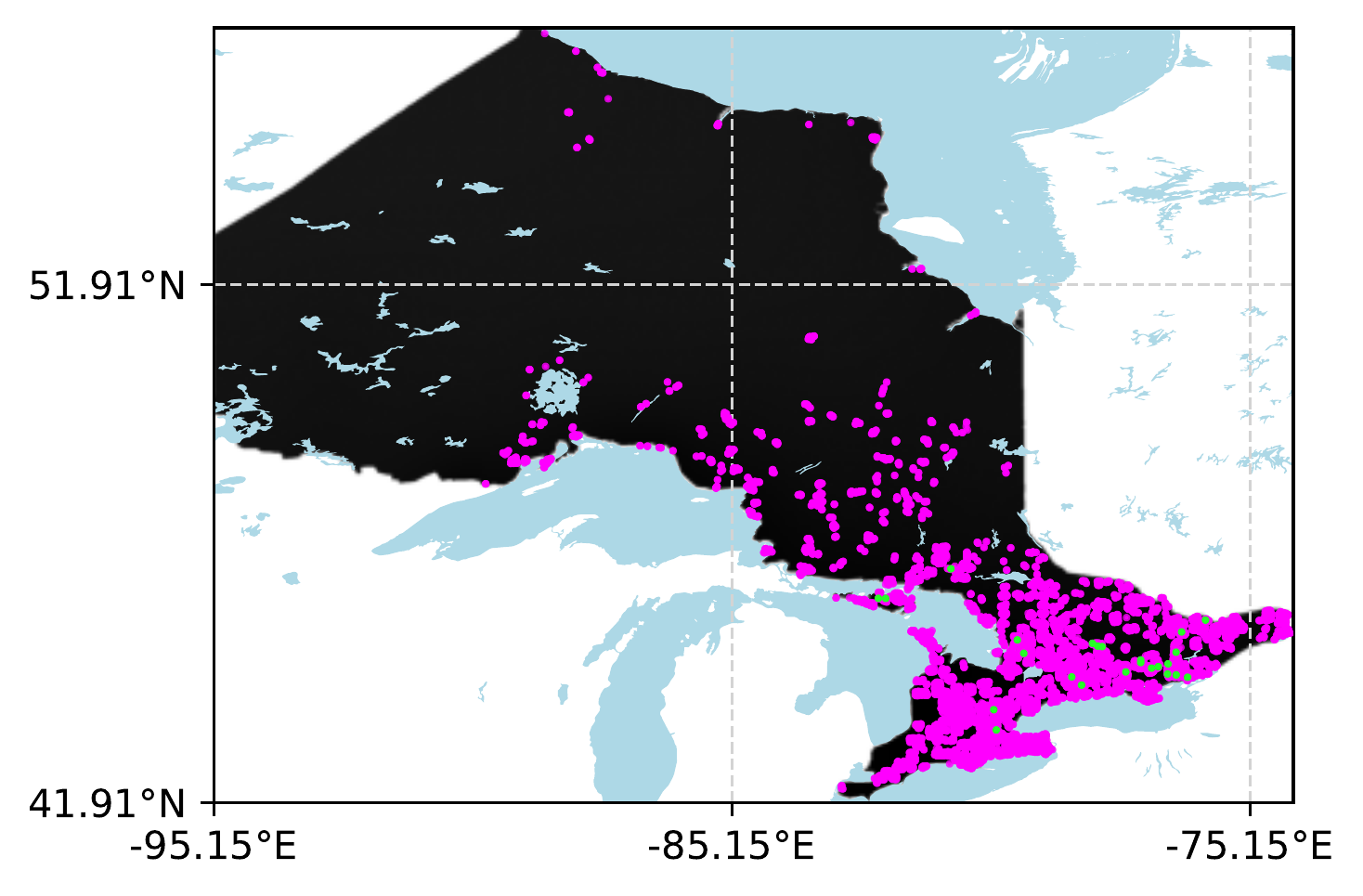}
        & \includegraphics[width=1.00\linewidth]{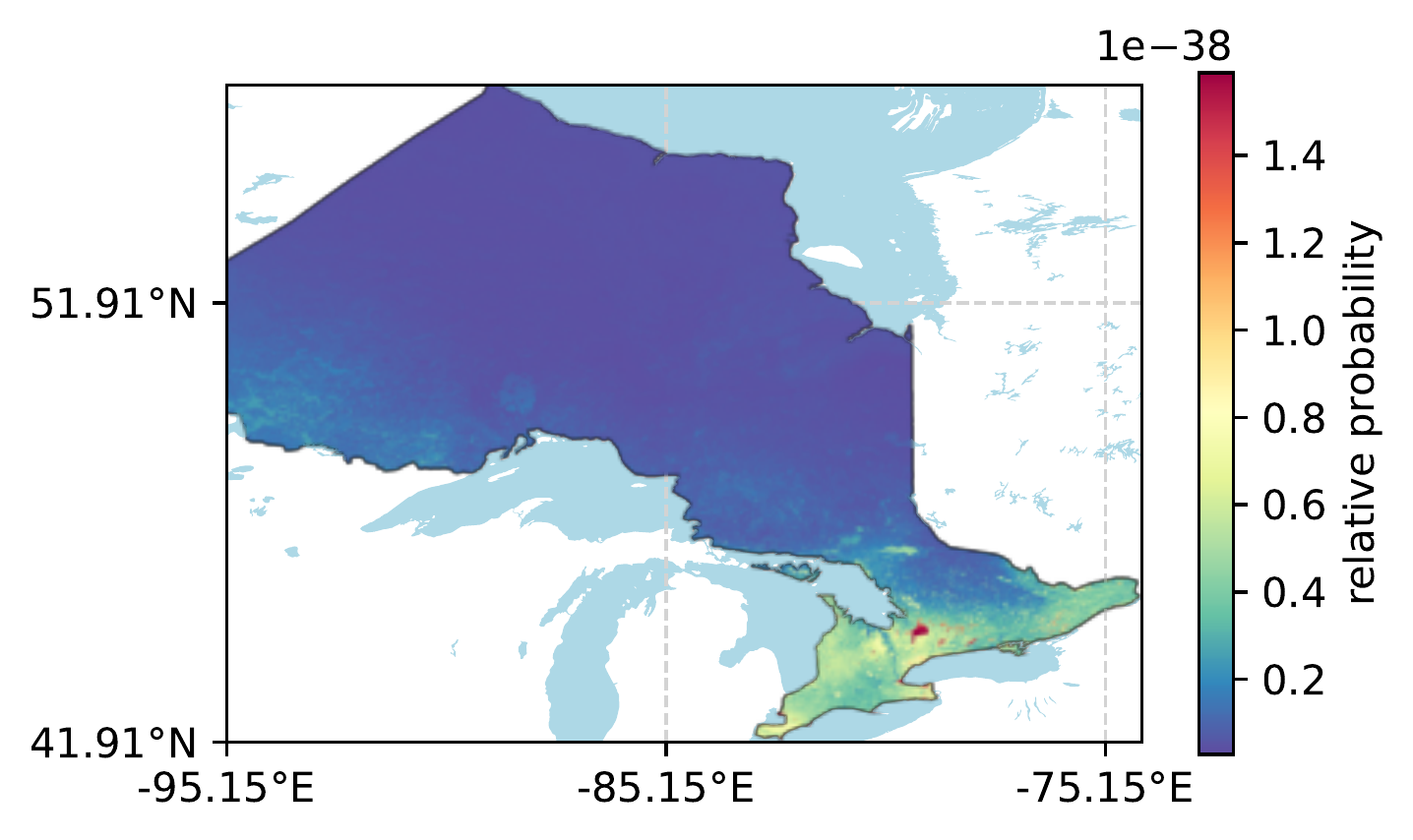} \\
        
        \textbf{can20} \newline
        & \includegraphics[width=1.00\linewidth]{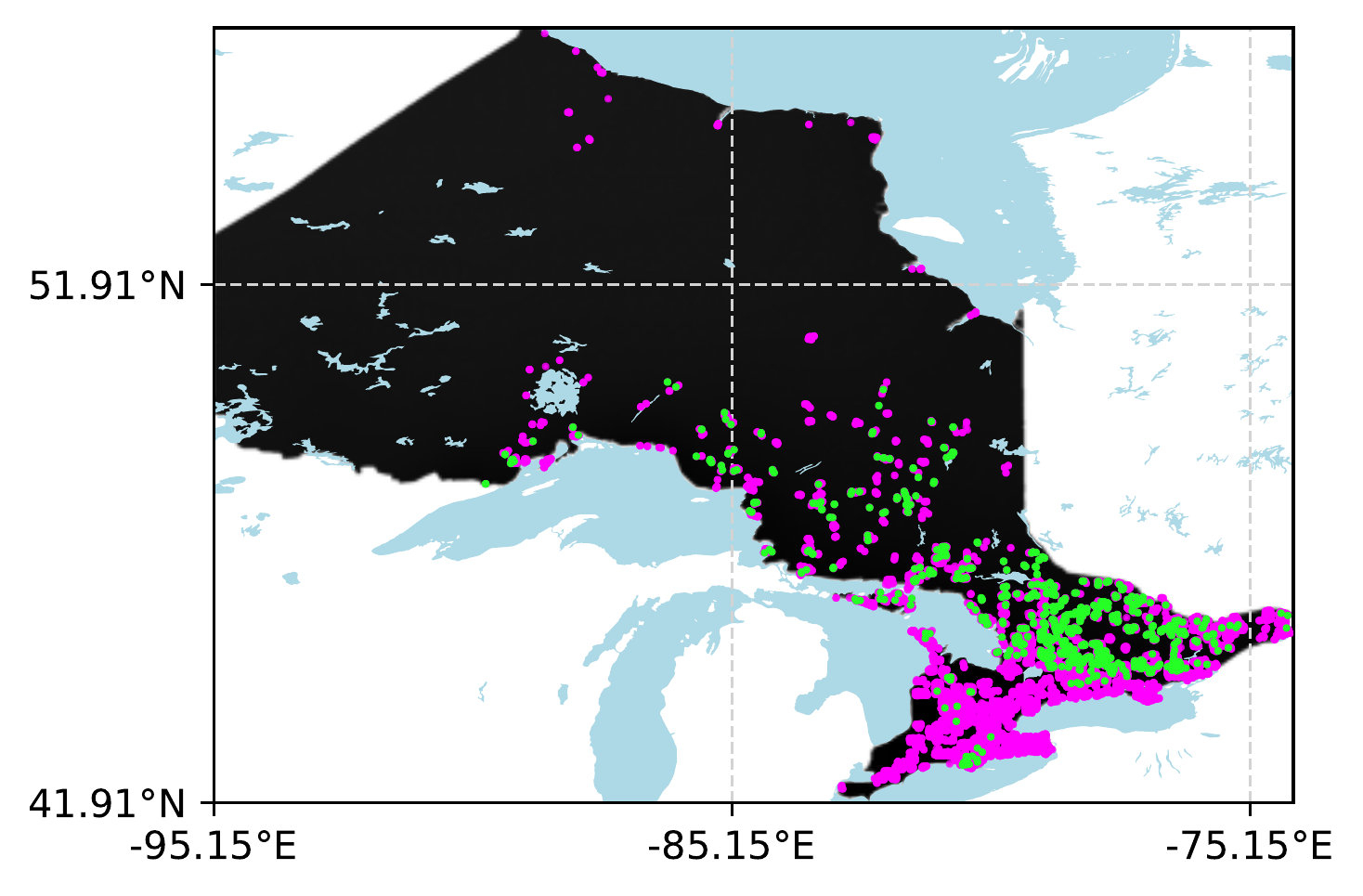}
        & \includegraphics[width=1.00\linewidth]{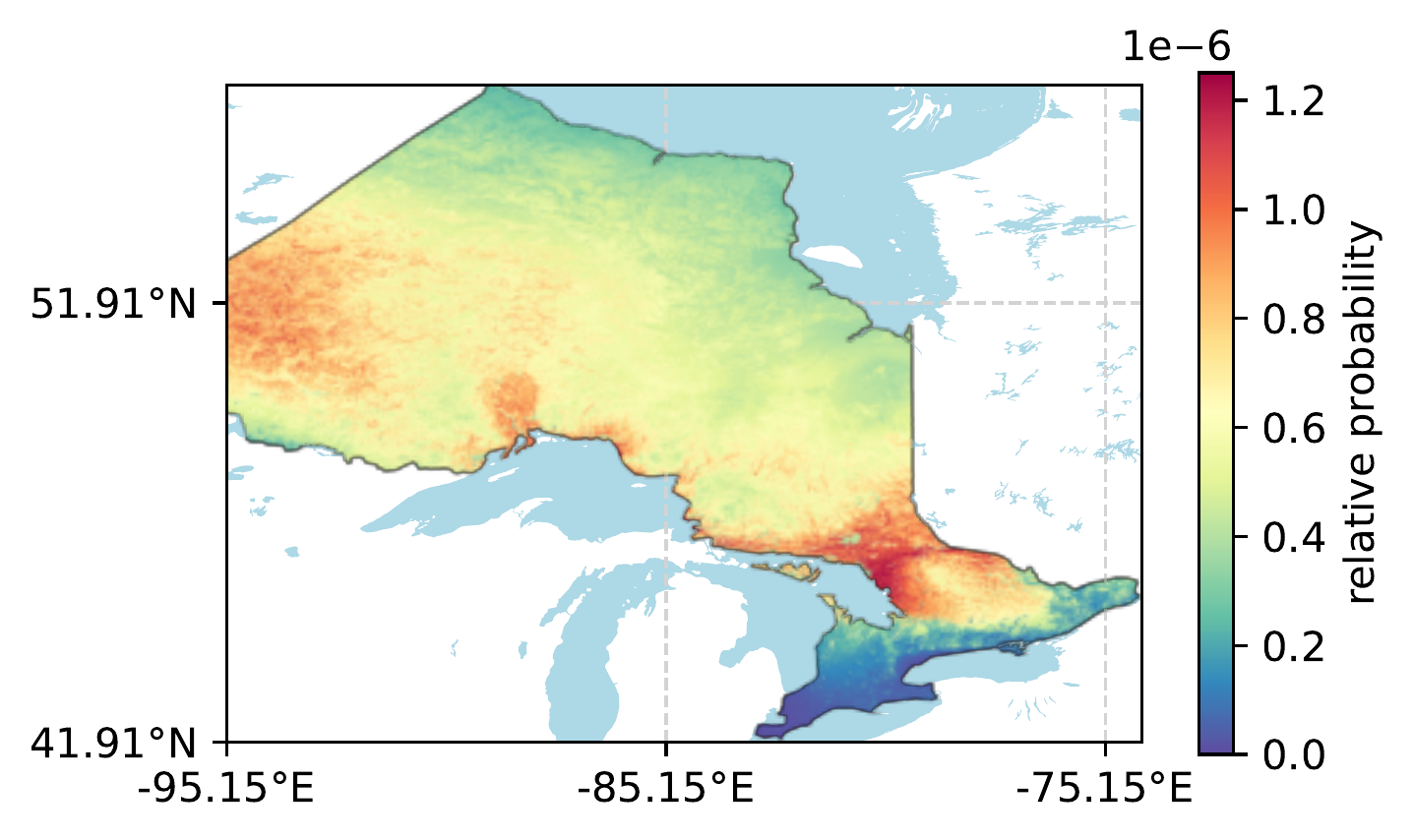} \\
    \end{tabular}
    \caption{Locations of species presences ({green}) and absences ({magenta}) based on PA observations, alongside predicted values from the reference DeepMaxent model across species can16 to can20}
    \label{fig:allmaps_withPA_4}
\end{figure}

\begin{figure}[h!]
    \centering
    \renewcommand{\arraystretch}{0.1}
    \begin{tabular}{>{\centering}m{4cm} m{6cm} m{6cm}}
        & \textbf{Mean} & \textbf{Std} \\
        \midrule
        \textbf{Reference} \newline 
        & \includegraphics[width=1.00\linewidth]{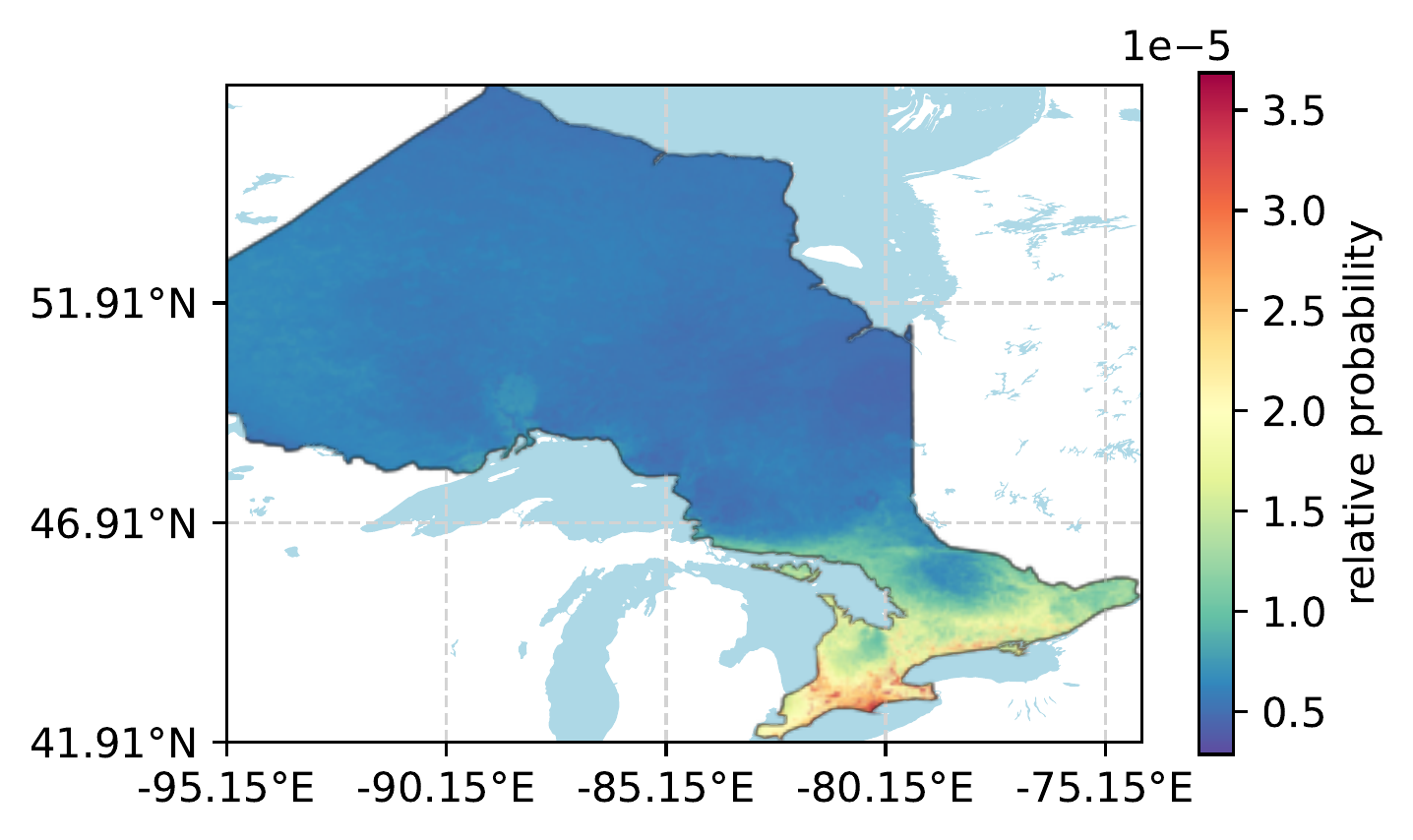}
        & \includegraphics[width=1.00\linewidth]{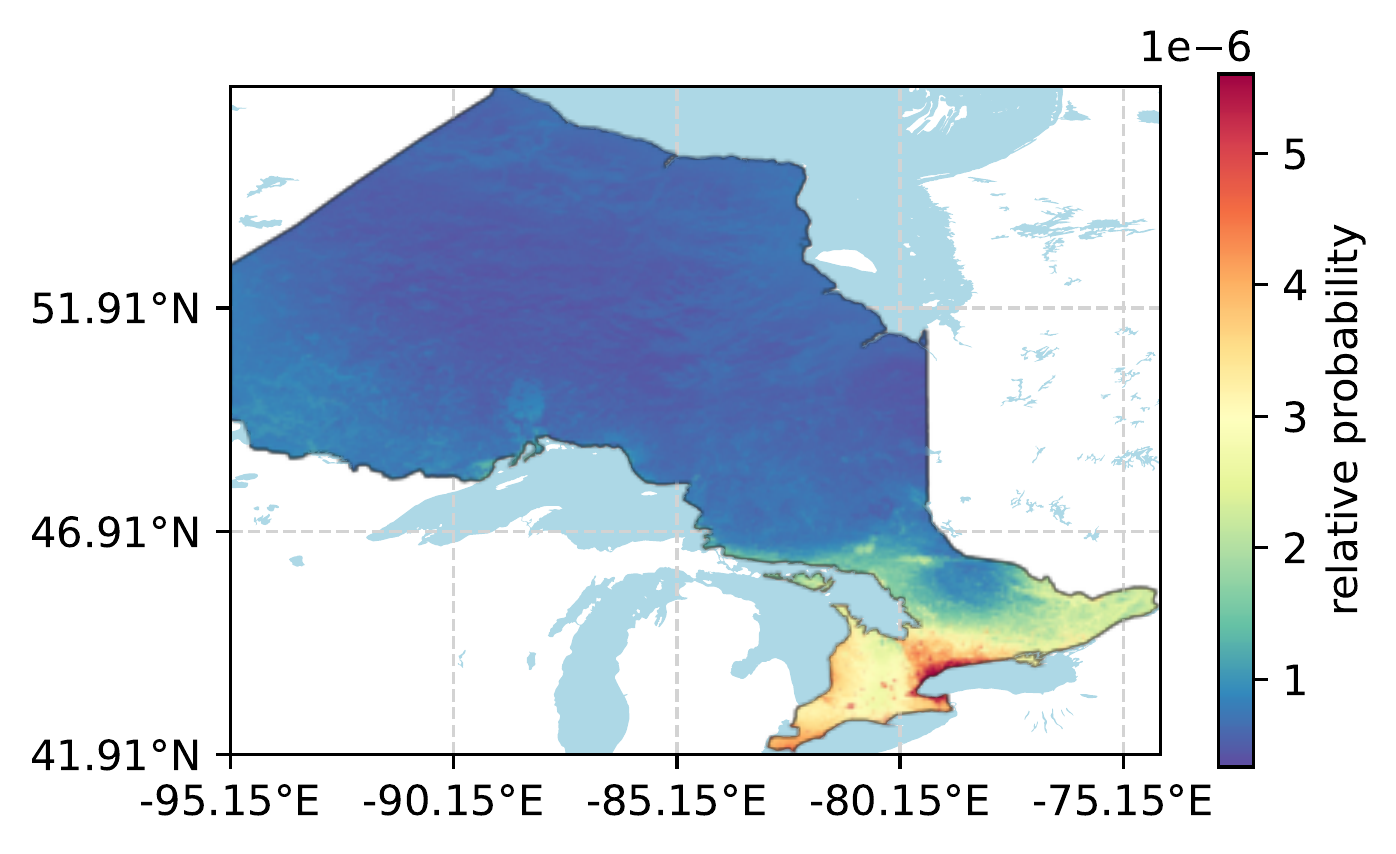} \\
        
        \textbf{Batch size} \newline \textit{Smaller value} 
        & \includegraphics[width=1.00\linewidth]{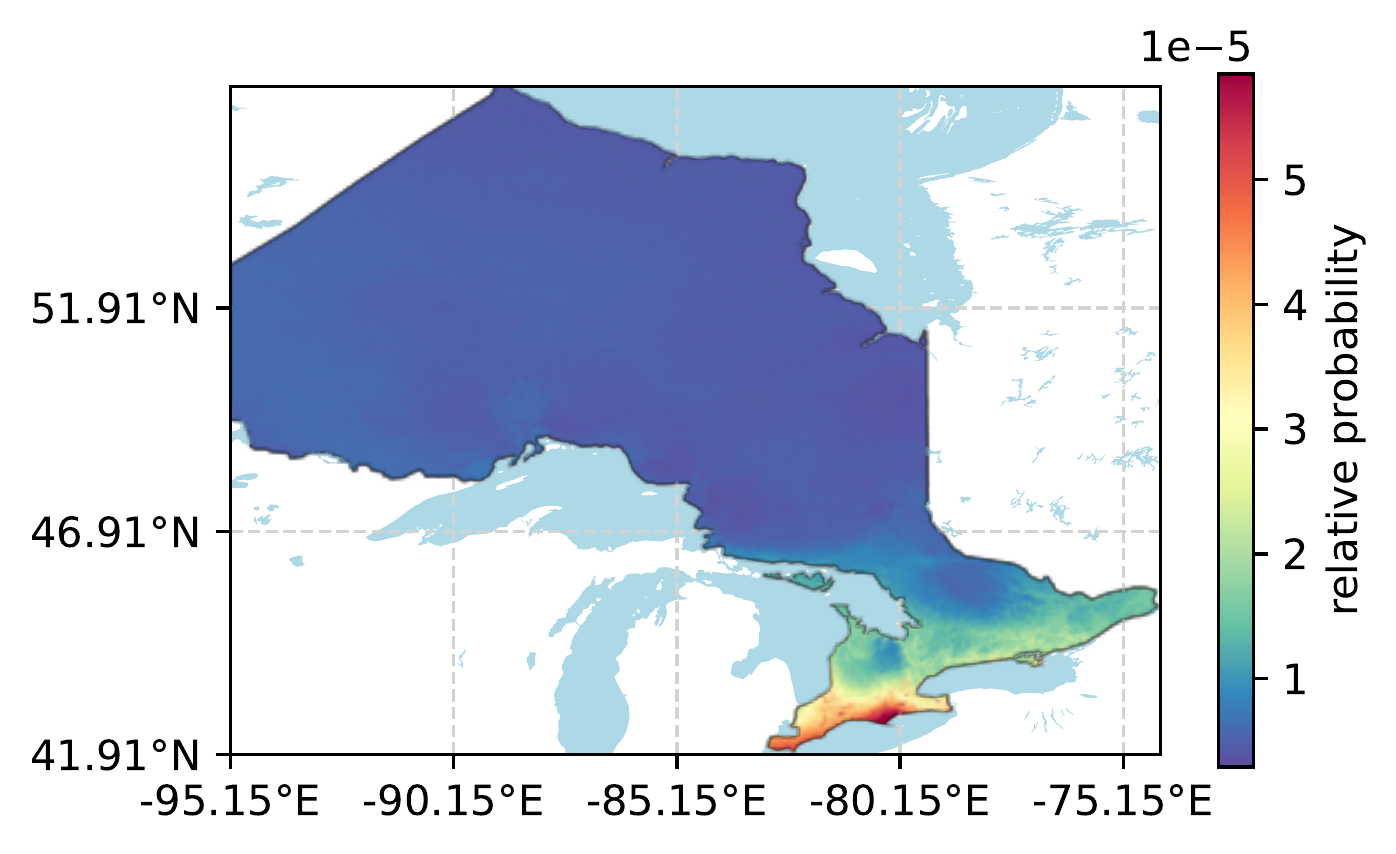}
        & \includegraphics[width=1.00\linewidth]{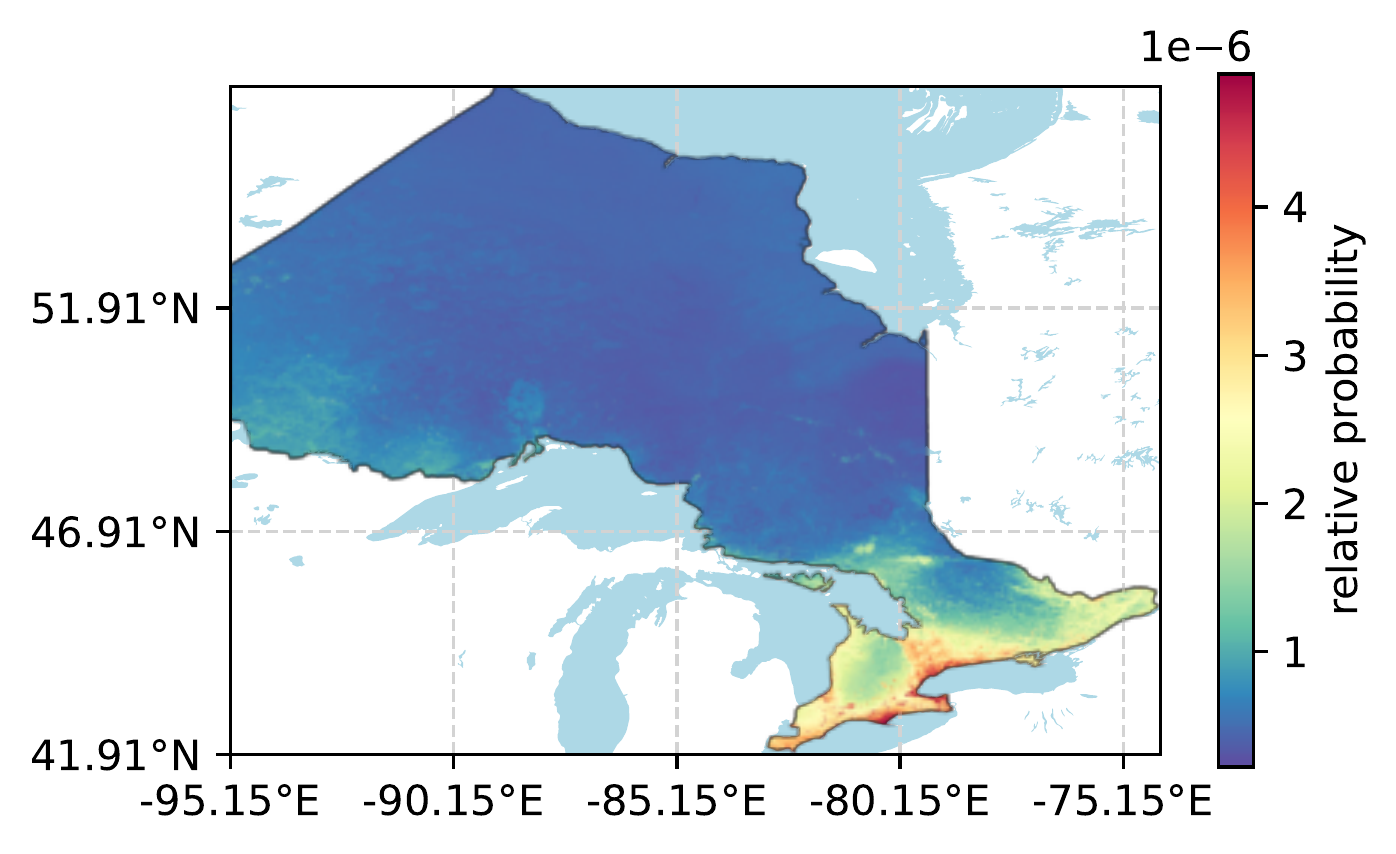} \\
        
        \textbf{Batch size} \newline \textit{Higher value}
        & \includegraphics[width=1.00\linewidth]{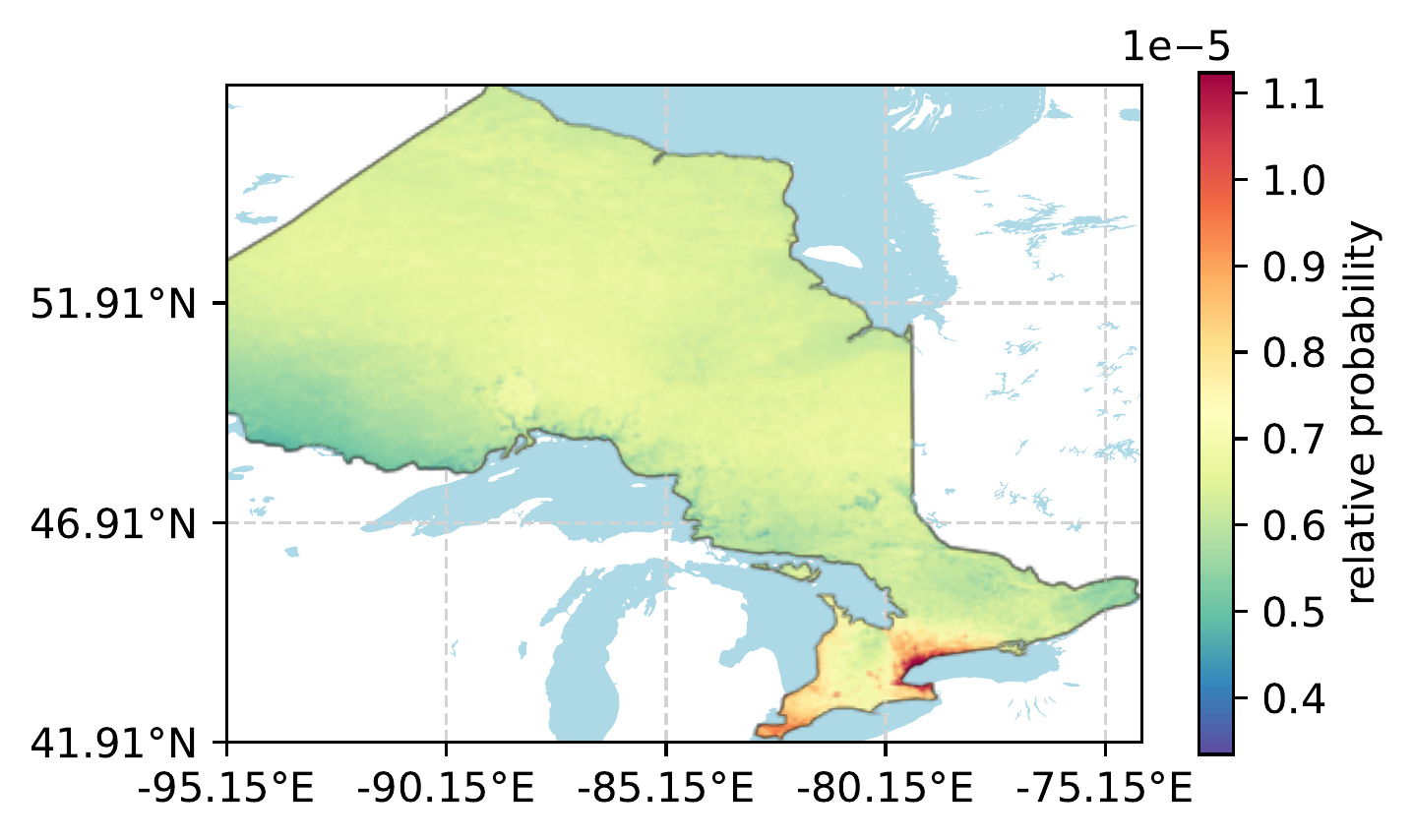}
        & \includegraphics[width=1.00\linewidth]{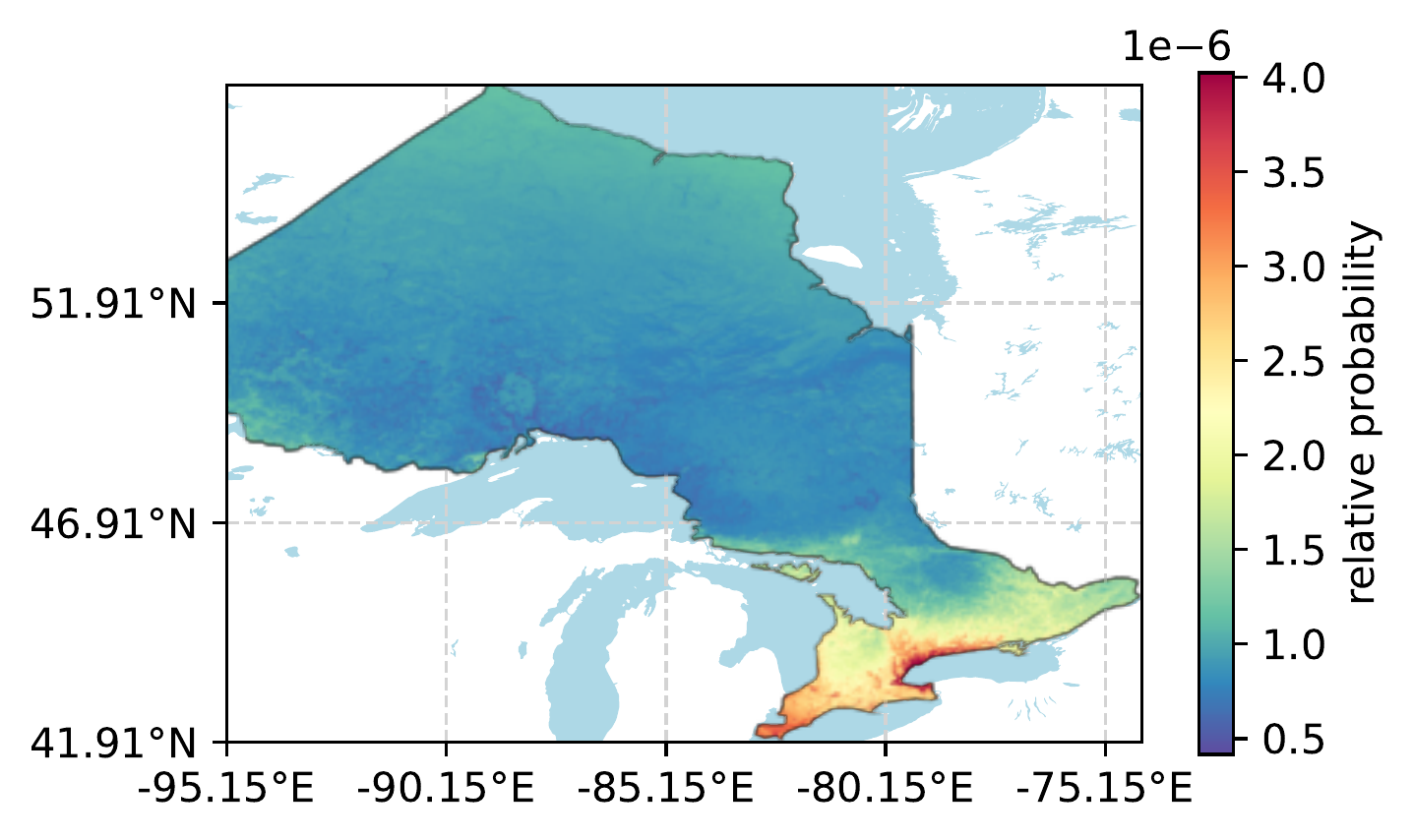} \\
        
        \textbf{Weight decay} \newline \textit{Smaller value}
        & \includegraphics[width=1.00\linewidth]{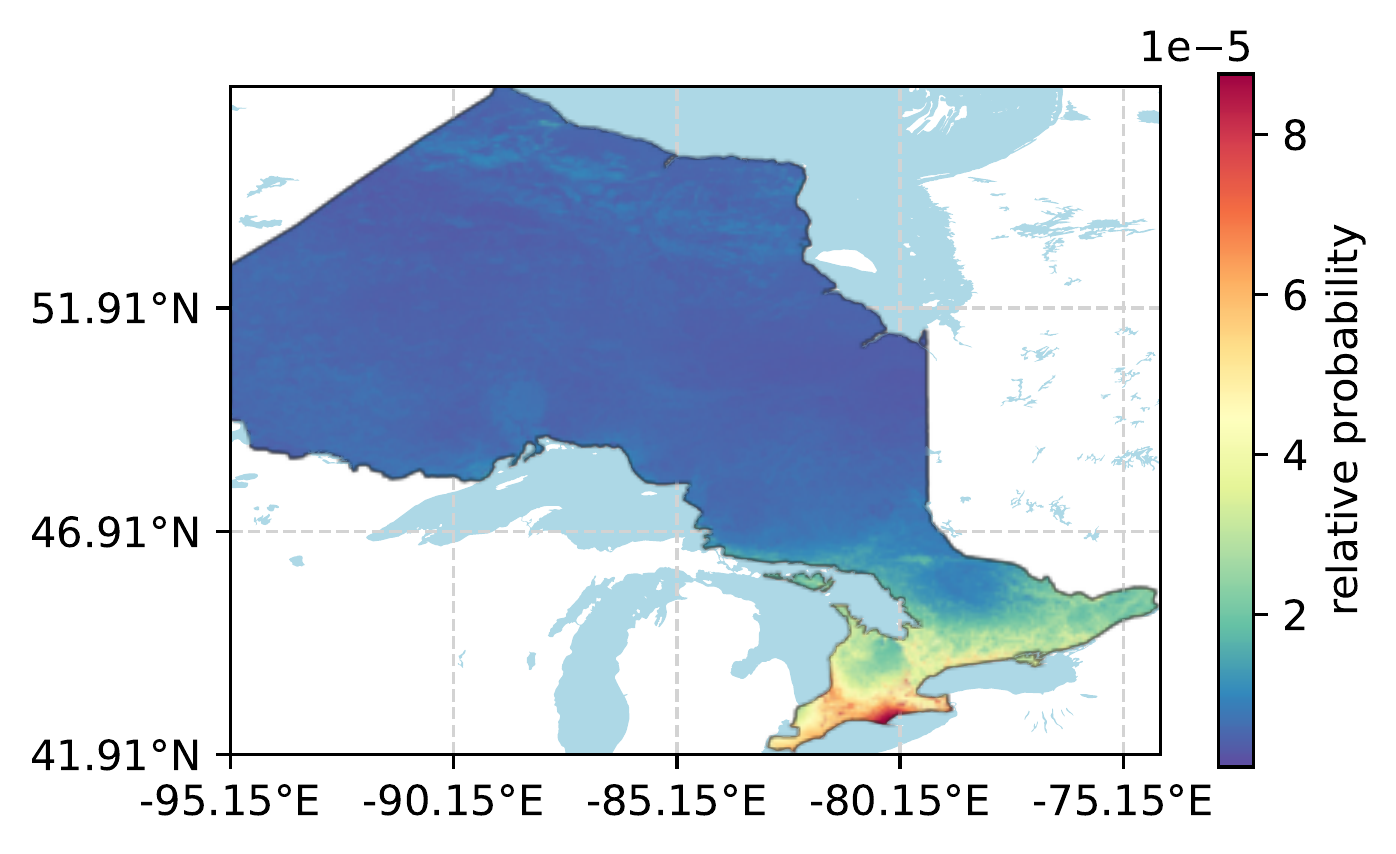}
        & \includegraphics[width=1.00\linewidth]{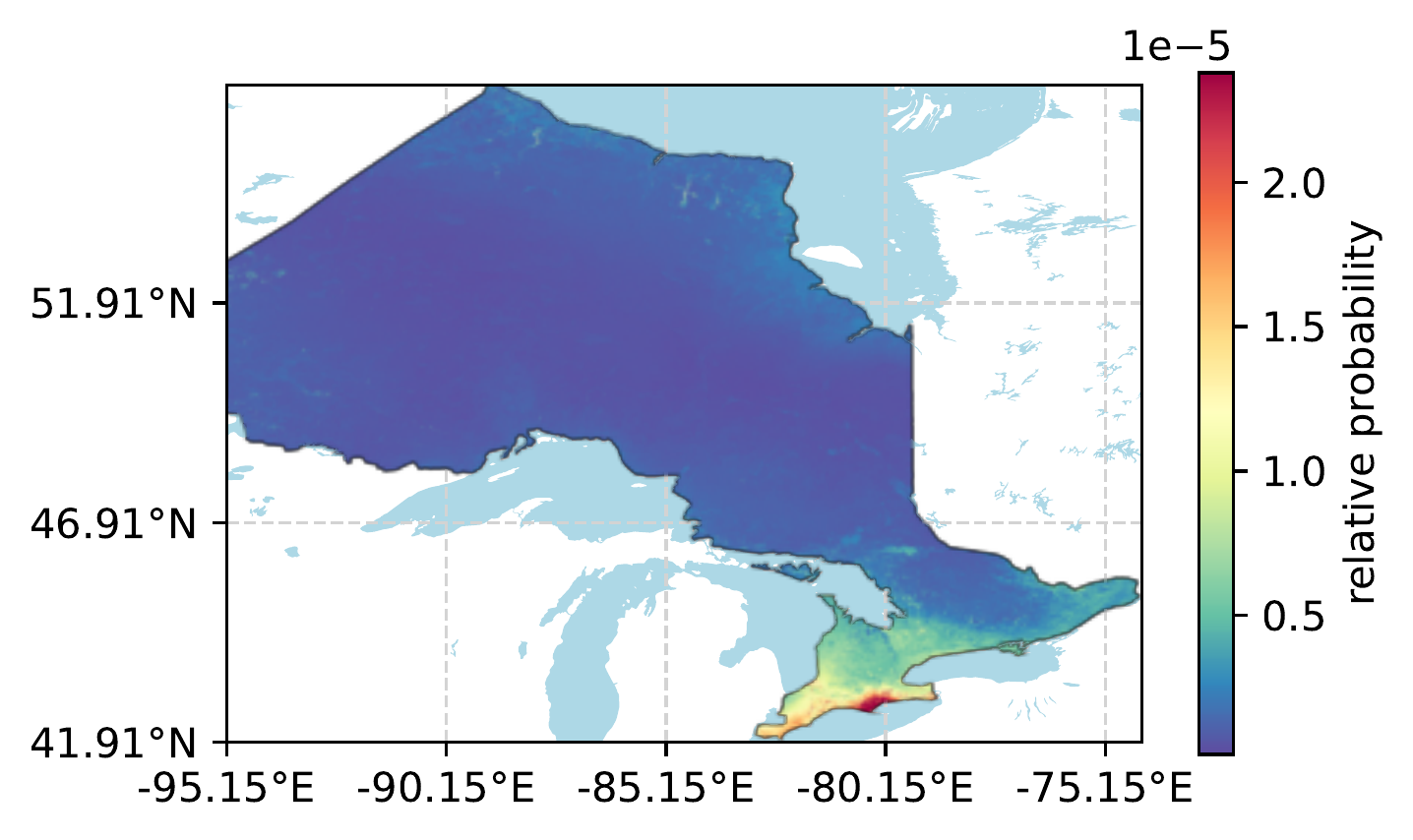} \\
        
        \textbf{Weight decay} \newline \textit{Higher value}
        & \includegraphics[width=1.00\linewidth]{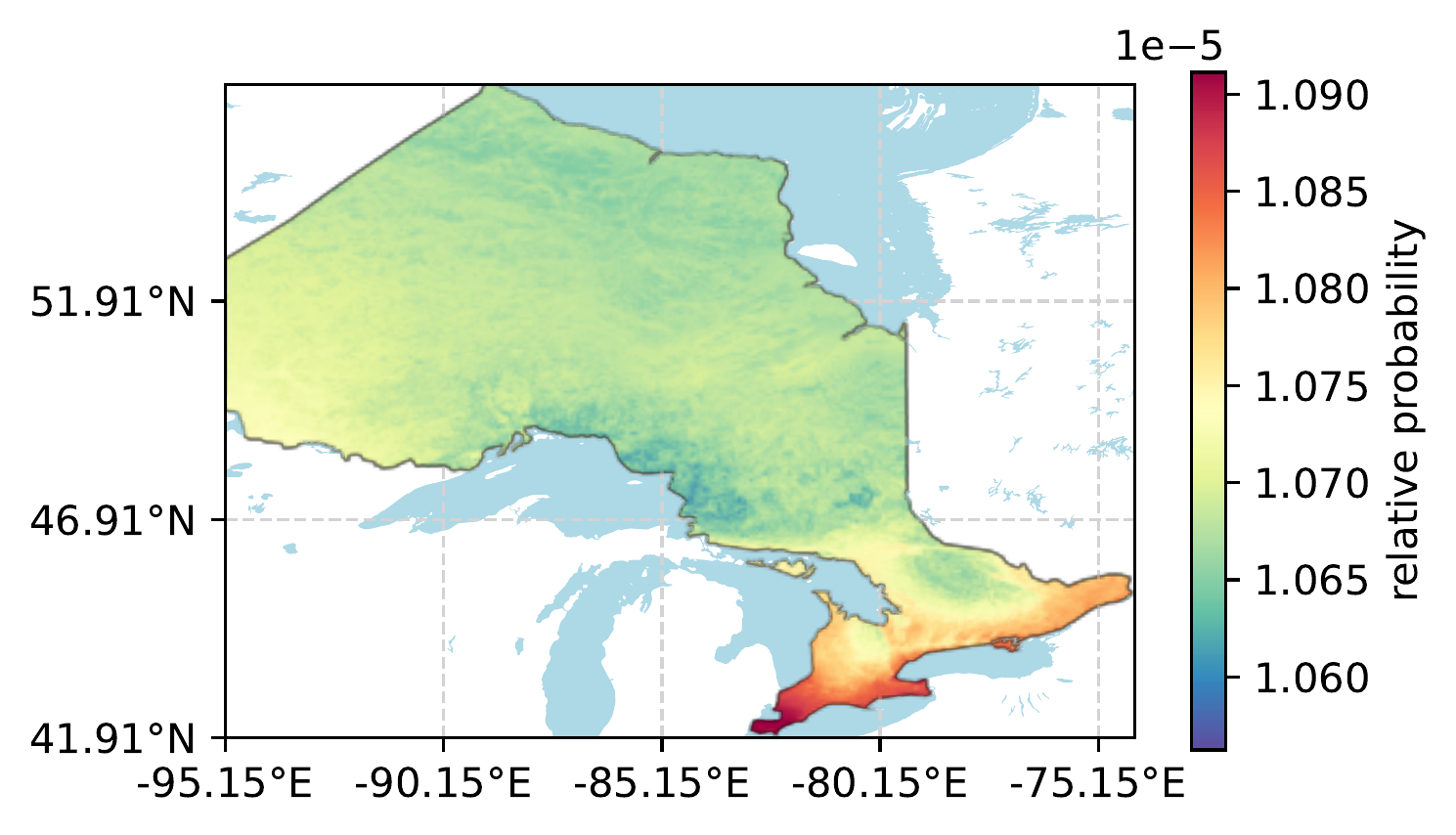}
        & \includegraphics[width=1.00\linewidth]{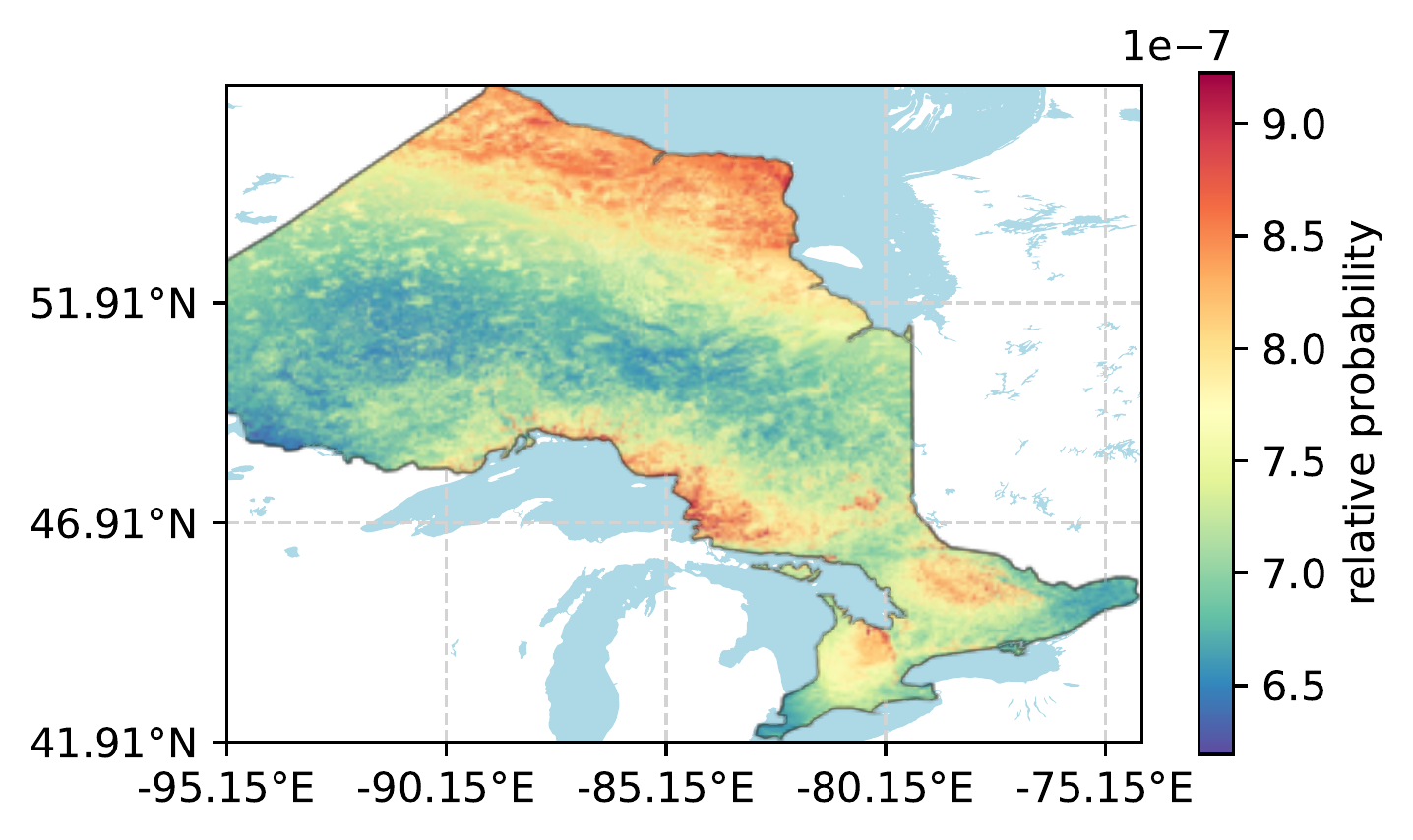}
    \end{tabular}
    \caption{Estimated mean and standard deviation of cumulative estimated probabilities across all species in the CAN region calculated over 10 random seeds under different hyper-parameter settings (batch size and weight decay).}
    \label{fig:allmaps_std}
\end{figure}

\section{Computational time}
\label{annex:computation_time}

{Table~\ref{tab:comp} summarizes the computational time of the different training procedures on the GeoPlant dataset (>4 thousand species, 5 million occurrences) on a laptop. The results indicate that the GPU implementation significantly accelerates the learning of DeepMaxent, whereas the CPU-based training required a considerably longer execution times. Additionally, even though Maxent was fitted with only 10,000 sites drawn among the millions of background sites, training DeepMaxent on a CPU (including all background sites) was twice quicker.}
\begin{table}[ht]
\centering
\begin{tabular}{l c}
\hline
\textbf{Method} & {Time} \\ \hline
Maxent (CPU) & $\sim$ 6.9 hours \\ 
DeepMaxent (CPU) & $\sim$ 3 hours \\
DeepMaxent (GPU) & $\sim$ 40 minutes \\
\hline
\end{tabular}
\caption{{Approximate training time for each model on the GeoPlant dataset. MaxEnt is trained as a single-species model, whereas DeepMaxent is trained jointly on all species for 100 epoch on CPU and GPU}}
\label{tab:comp}
\end{table}

\section{Dataset description}

\label{annex:dataset}

\subsection{NCEAS}

\begin{figure}[h!]
    \centering
    \includegraphics[width=1\linewidth]{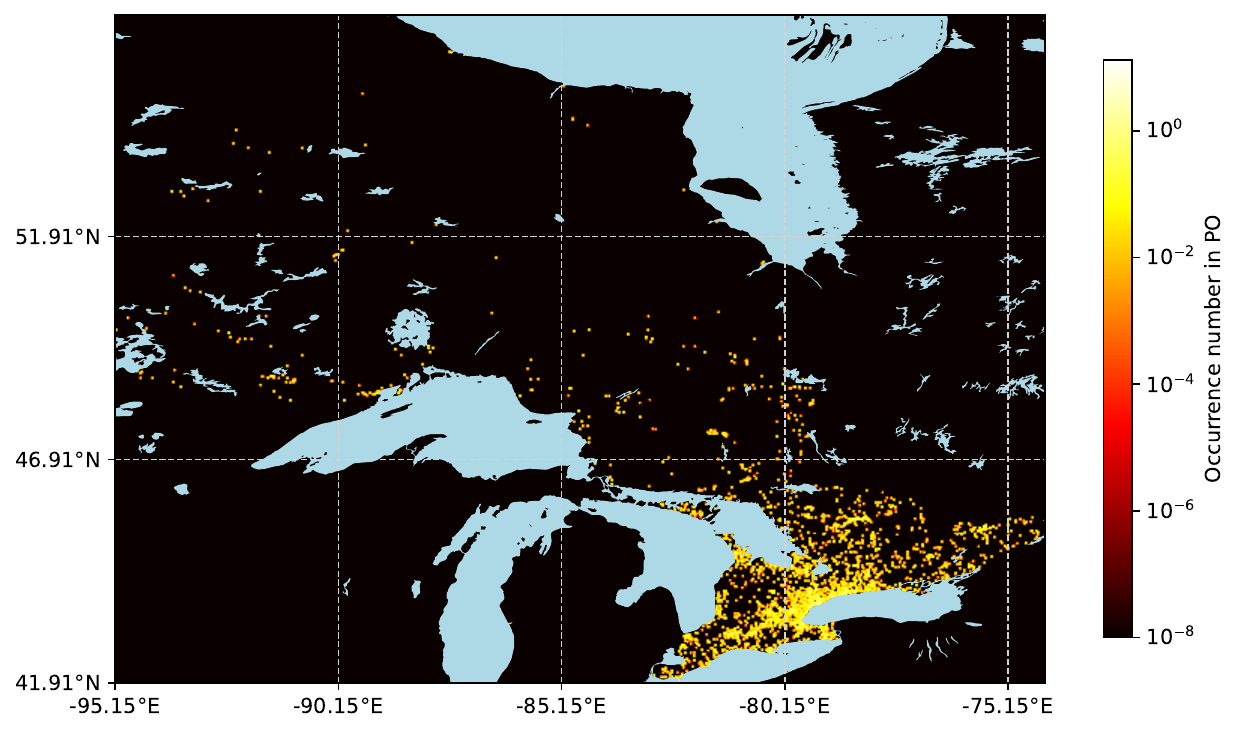}
    \caption{Location of presence-only (PO) occurrences for all 20 bird species in Canada}
    \label{fig:PO_occu_map}
\end{figure}

\begin{figure}[h!]
    \centering
    \includegraphics[width=0.65\linewidth]{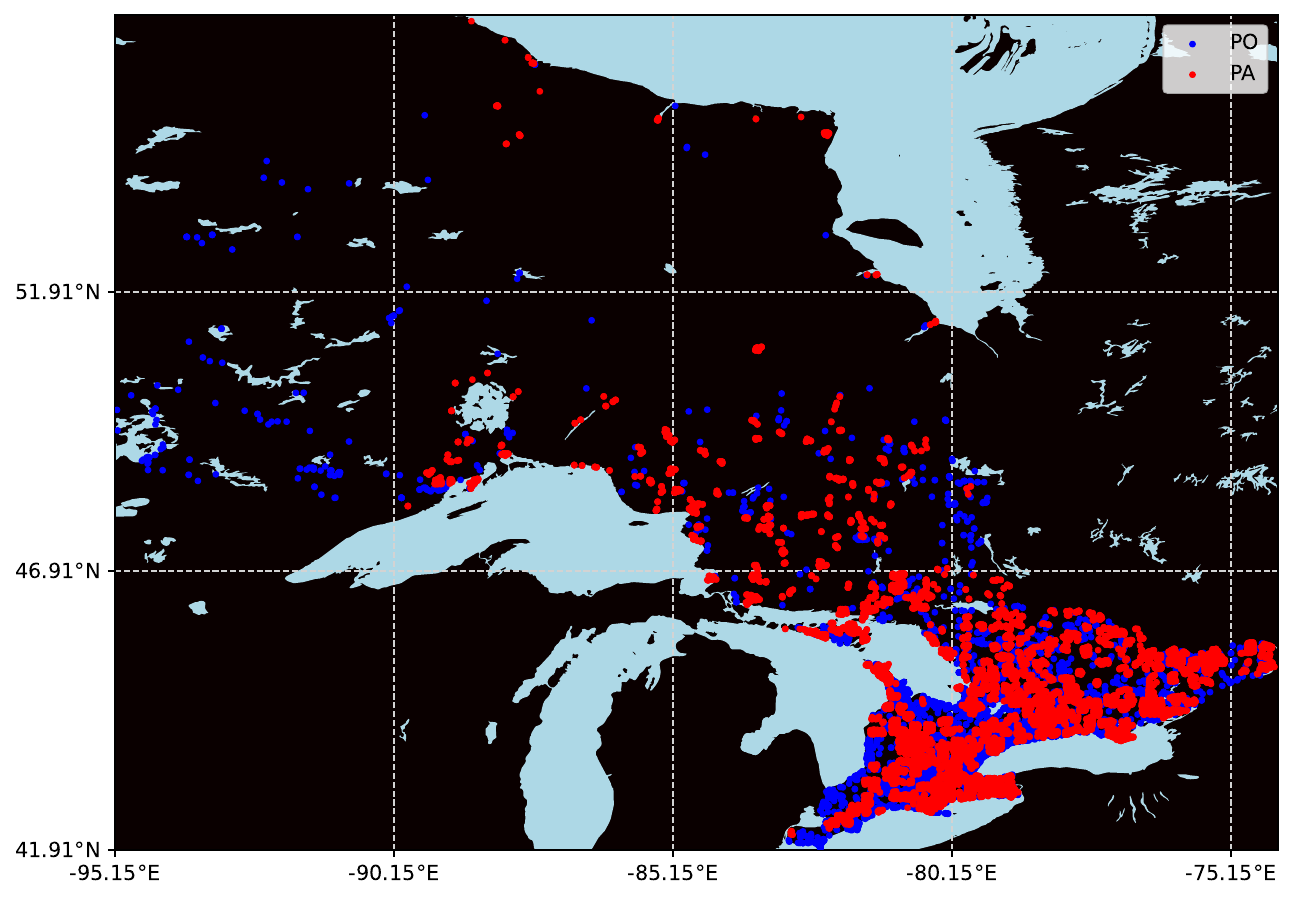}
    \caption{Occurrences for all 20 bird species in Canada, for PO and PA data)}
    \label{fig:PO_PA}
\end{figure}

\begin{figure}[h!]
    \centering
    \begin{subfigure}[b]{0.45\linewidth}
        \centering
        \includegraphics[width=0.60\linewidth]{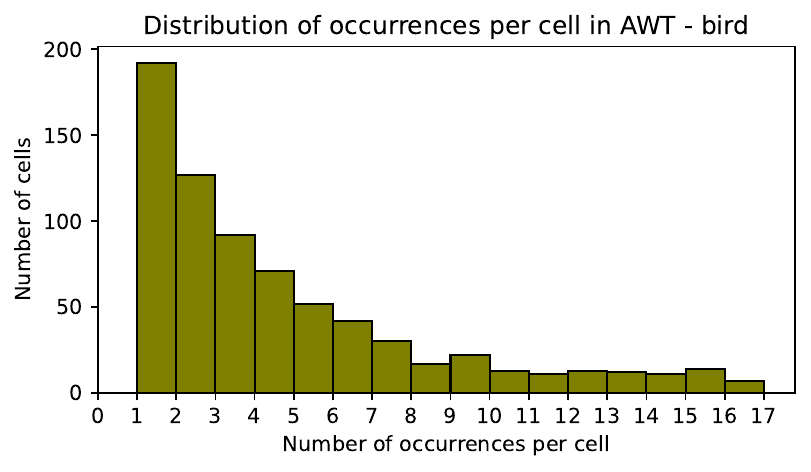}
        \caption{}
    \end{subfigure}
        \centering
    \begin{subfigure}[b]{0.45\linewidth}
        \centering
        \includegraphics[width=0.65\linewidth]{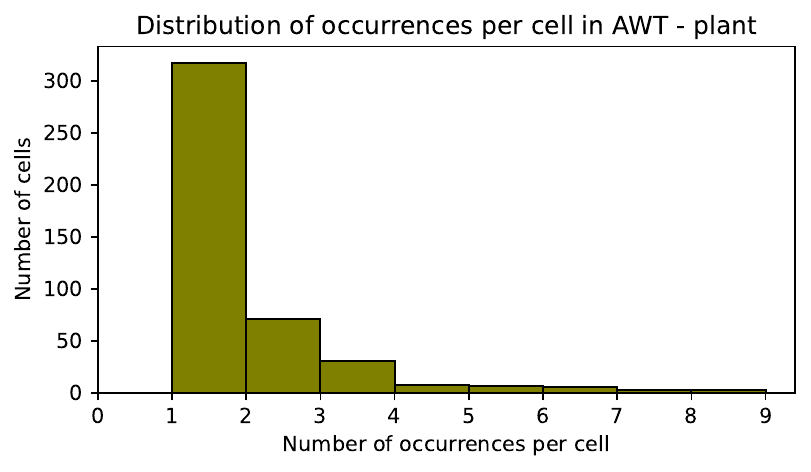}
        \caption{}
    \end{subfigure}
    \begin{subfigure}[b]{0.45\linewidth}
        \centering
        \includegraphics[width=0.60\linewidth]{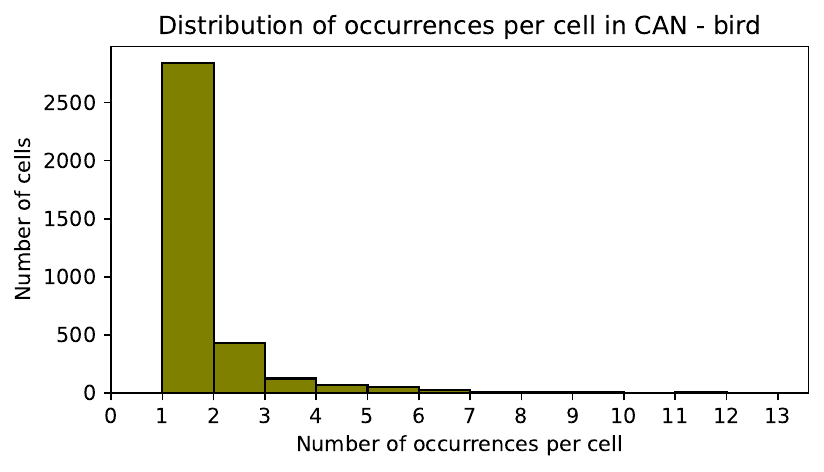}
        \caption{}
    \end{subfigure}
    \begin{subfigure}[b]{0.45\linewidth}
        \centering
        \includegraphics[width=0.60\linewidth]{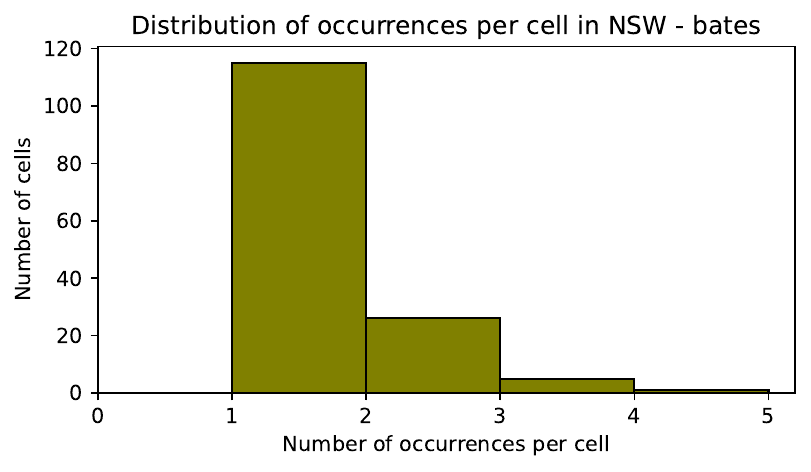}
        \caption{}
    \end{subfigure}
\begin{subfigure}[b]{0.45\linewidth}
        \centering
        \includegraphics[width=0.60\linewidth]{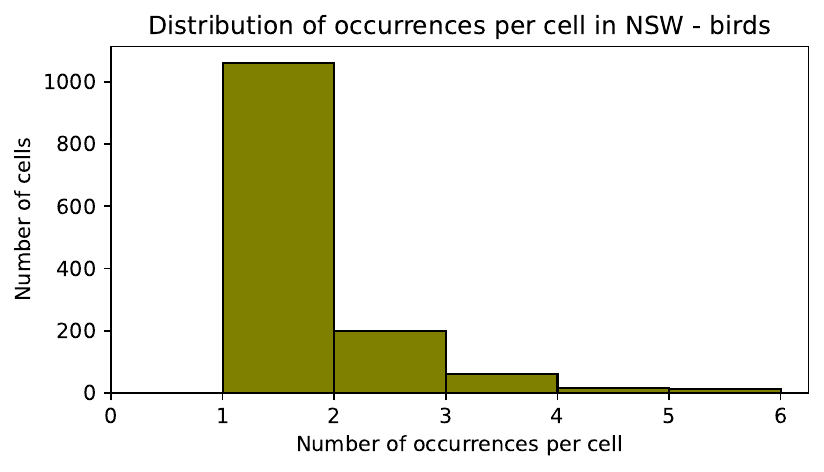}
        \caption{}
    \end{subfigure}
\begin{subfigure}[b]{0.45\linewidth}
        \centering
        \includegraphics[width=0.60\linewidth]{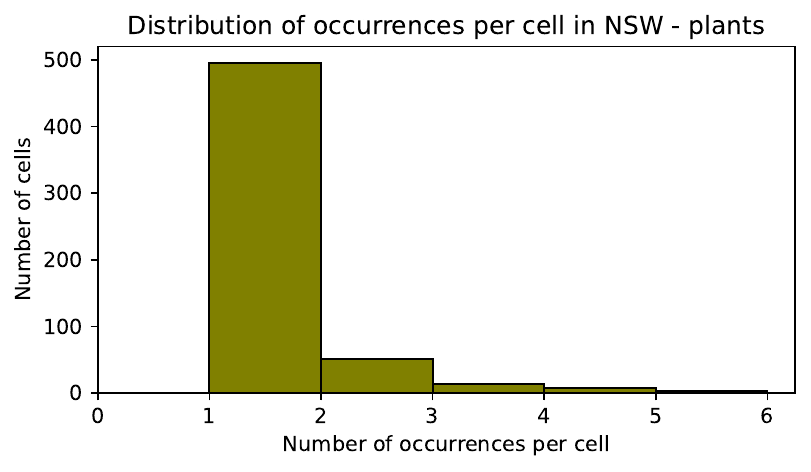}
        \caption{}
    \end{subfigure}
\begin{subfigure}[b]{0.45\linewidth}
        \centering
        \includegraphics[width=0.60\linewidth]{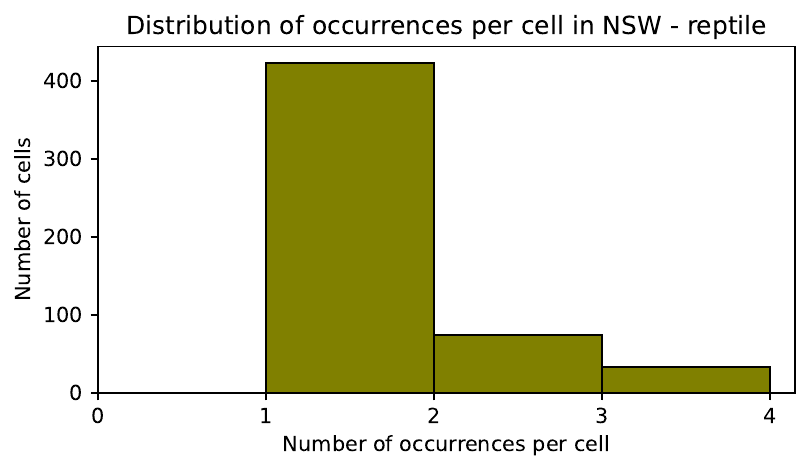}
        \caption{}
    \end{subfigure}
\begin{subfigure}[b]{0.45\linewidth}
        \centering
        \includegraphics[width=0.60\linewidth]{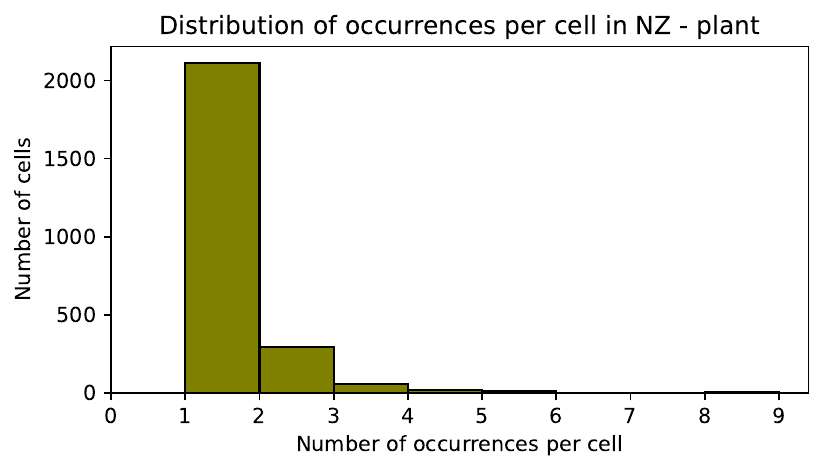}
        \caption{}
    \end{subfigure}
    \begin{subfigure}[b]{0.45\linewidth}
        \centering
        \includegraphics[width=0.60\linewidth]{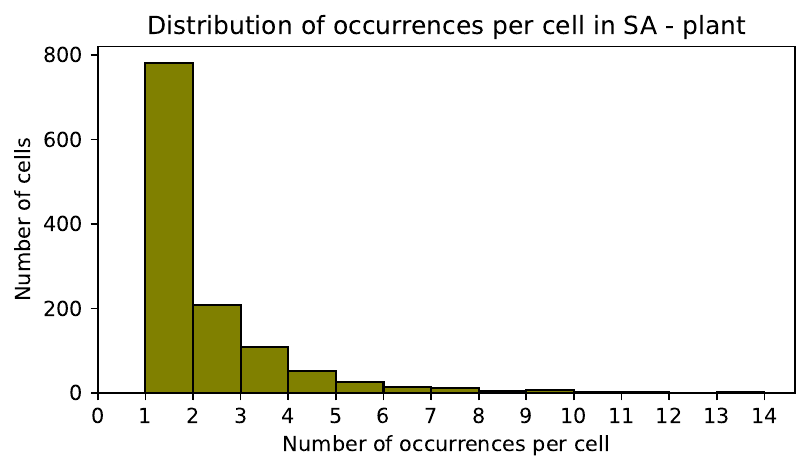}
        \caption{}
    \end{subfigure}
        \begin{subfigure}[b]{0.45\linewidth}
        \centering
        \includegraphics[width=0.60\linewidth]{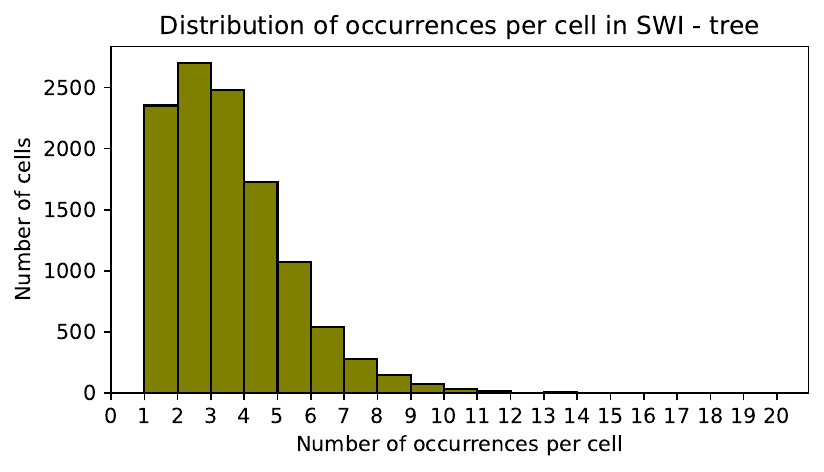}
        \caption{} 
    \end{subfigure}
    \caption{Distribution of occurrence numbers in cells that contain at least one occurrence for each region and biological group: (a) AWT bird, (b) AWT plant, (c) CAN bird, (d) NSW bates, (e) NSW bird, (f) NSW plant, (g) NSW reptile, (h) NZ plant, (i) SA plant and (j) SWI tree } 
\end{figure}

\begin{figure}[h!]
    \centering
    \begin{subfigure}[b]{0.45\linewidth}
        \centering
        \includegraphics[width=0.75\linewidth]{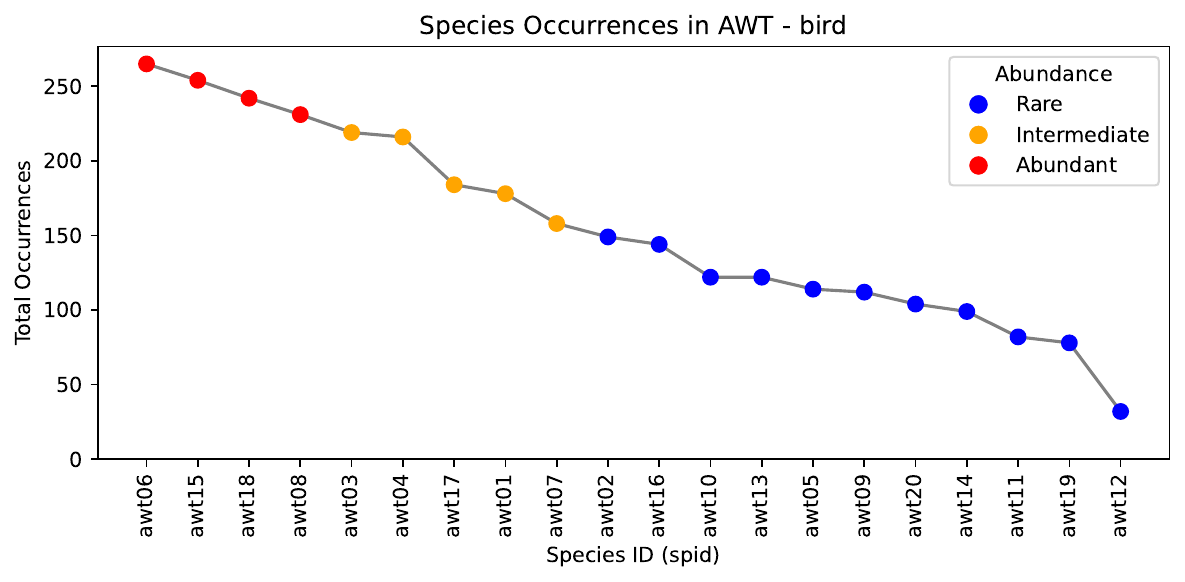}
        \caption{}
    \end{subfigure}
        \centering
    \begin{subfigure}[b]{0.45\linewidth}
        \centering
        \includegraphics[width=0.75\linewidth]{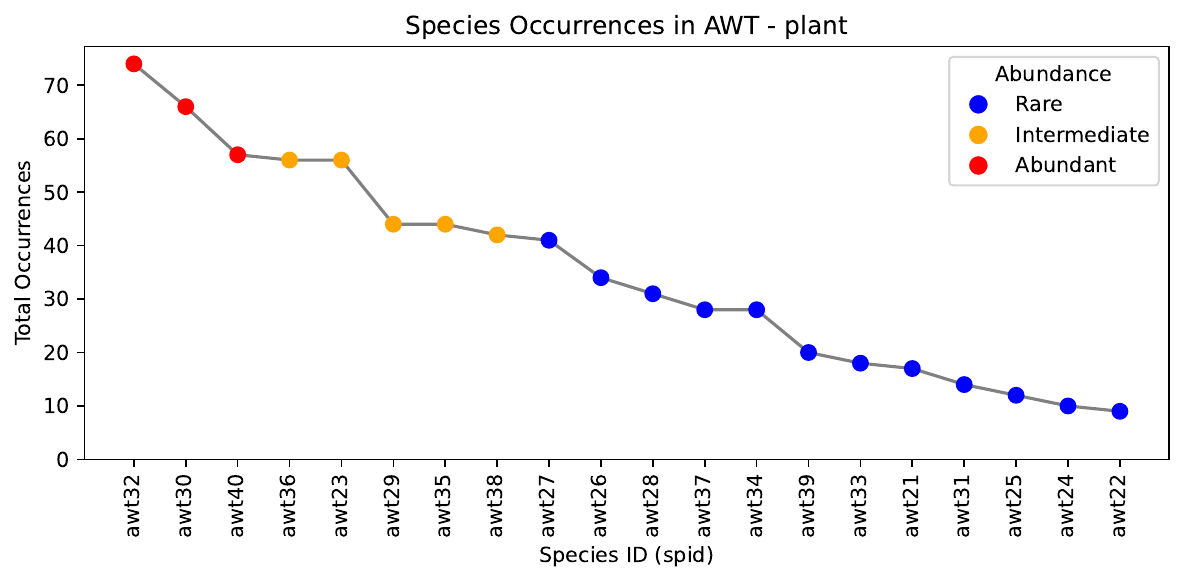}
        \caption{}
    \end{subfigure}
    \begin{subfigure}[b]{0.45\linewidth}
        \centering
        \includegraphics[width=0.75\linewidth]{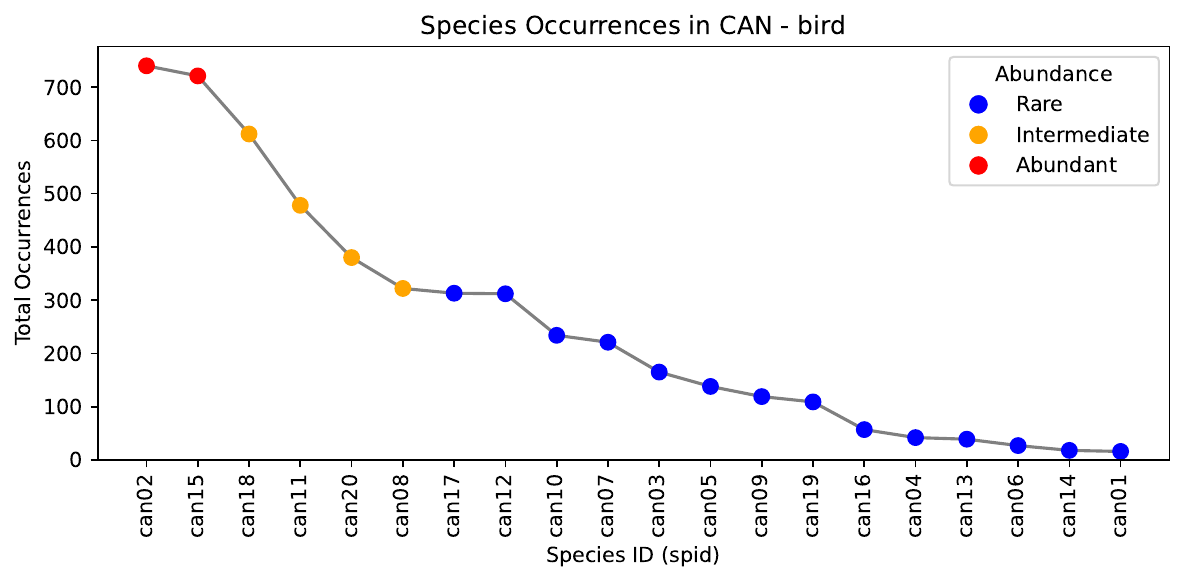}
        \caption{}
    \end{subfigure}
    \begin{subfigure}[b]{0.45\linewidth}
        \centering
        \includegraphics[width=0.75\linewidth]{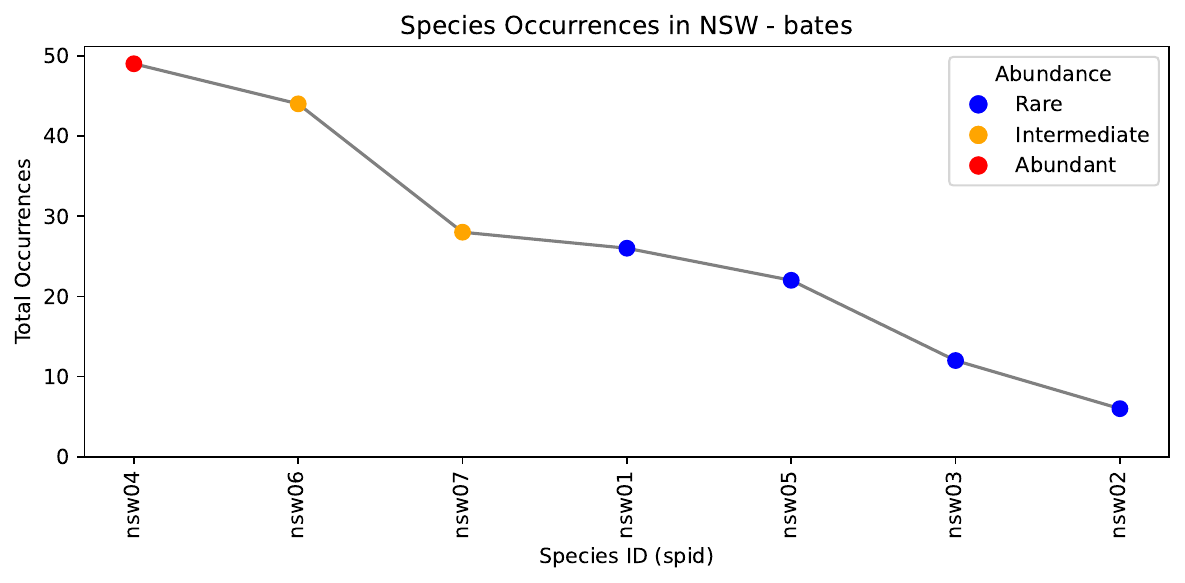}
        \caption{}
    \end{subfigure}
\begin{subfigure}[b]{0.45\linewidth}
        \centering
        \includegraphics[width=0.75\linewidth]{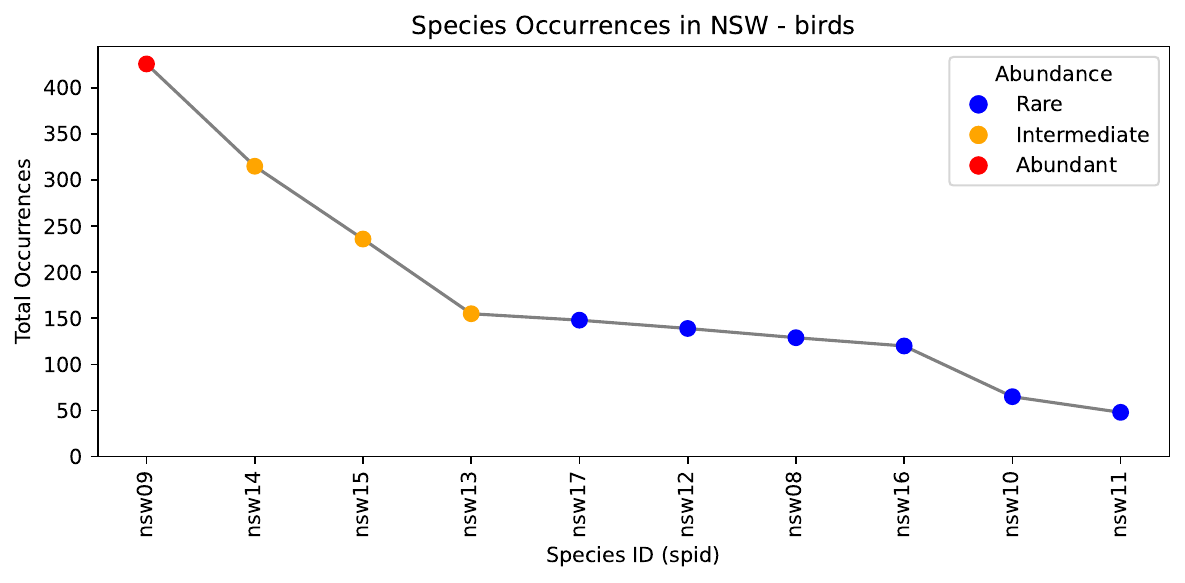}
        \caption{}
    \end{subfigure}
\begin{subfigure}[b]{0.45\linewidth}
        \centering
        \includegraphics[width=0.75\linewidth]{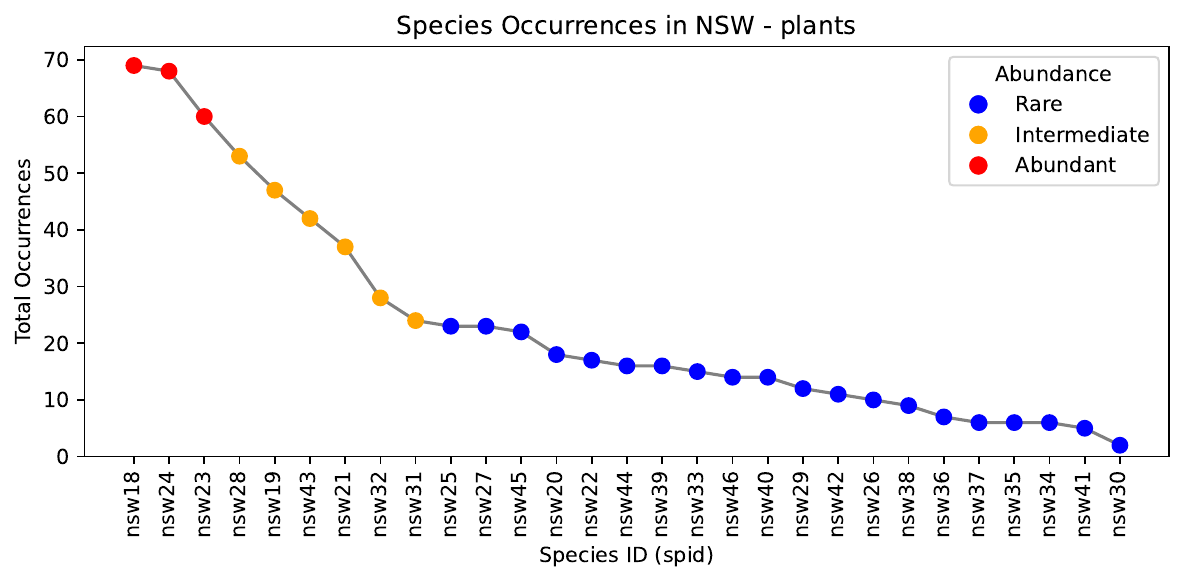}
        \caption{}
    \end{subfigure}
\begin{subfigure}[b]{0.45\linewidth}
        \centering
        \includegraphics[width=0.75\linewidth]{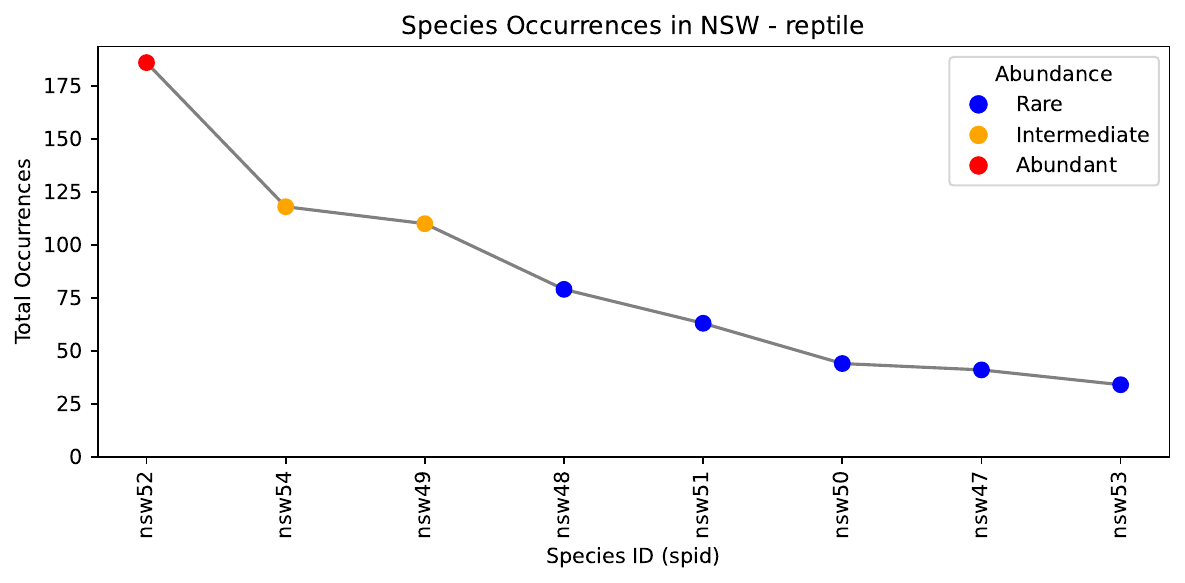}
        \caption{}
    \end{subfigure}
\begin{subfigure}[b]{0.45\linewidth}
        \centering
        \includegraphics[width=0.75\linewidth]{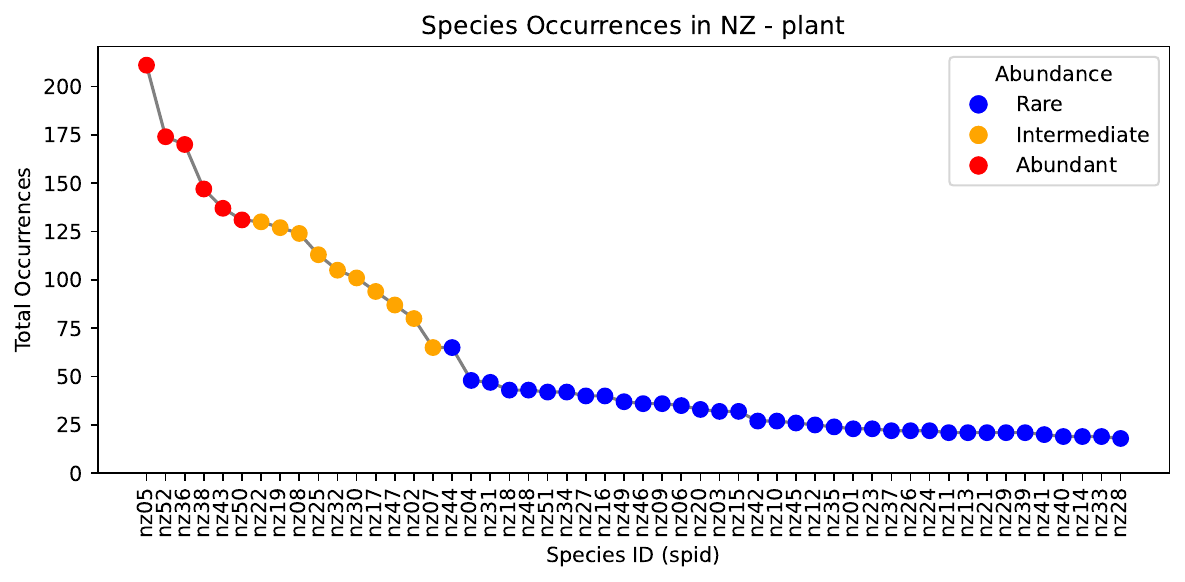}
        \caption{}
    \end{subfigure}
    \begin{subfigure}[b]{0.45\linewidth}
        \centering
        \includegraphics[width=0.75\linewidth]{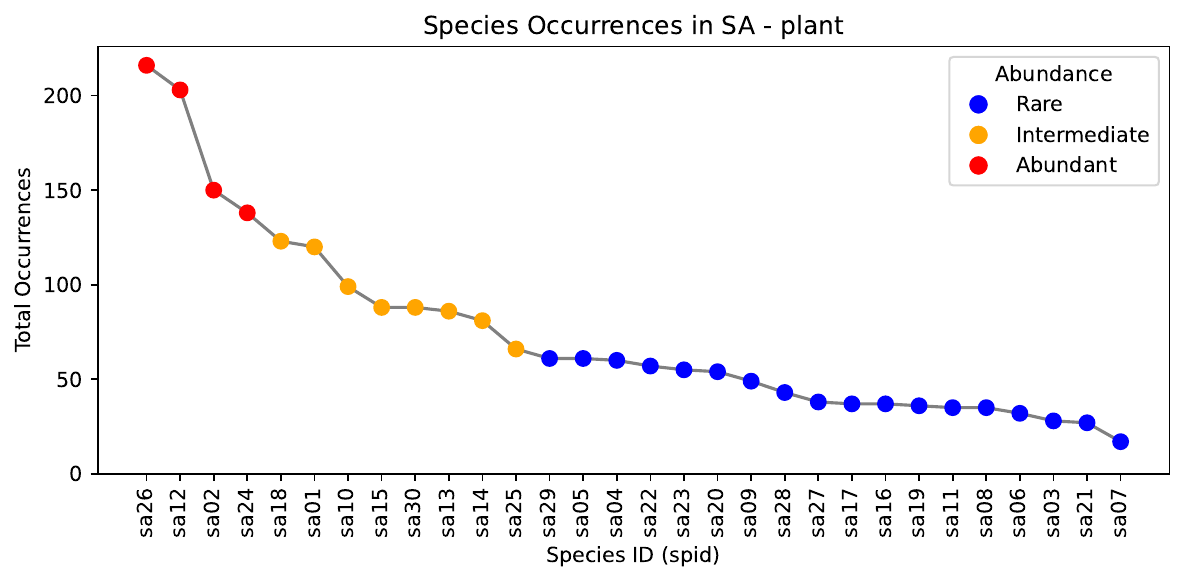}
        \caption{}
    \end{subfigure}
        \begin{subfigure}[b]{0.45\linewidth}
        \centering
        \includegraphics[width=0.75\linewidth]{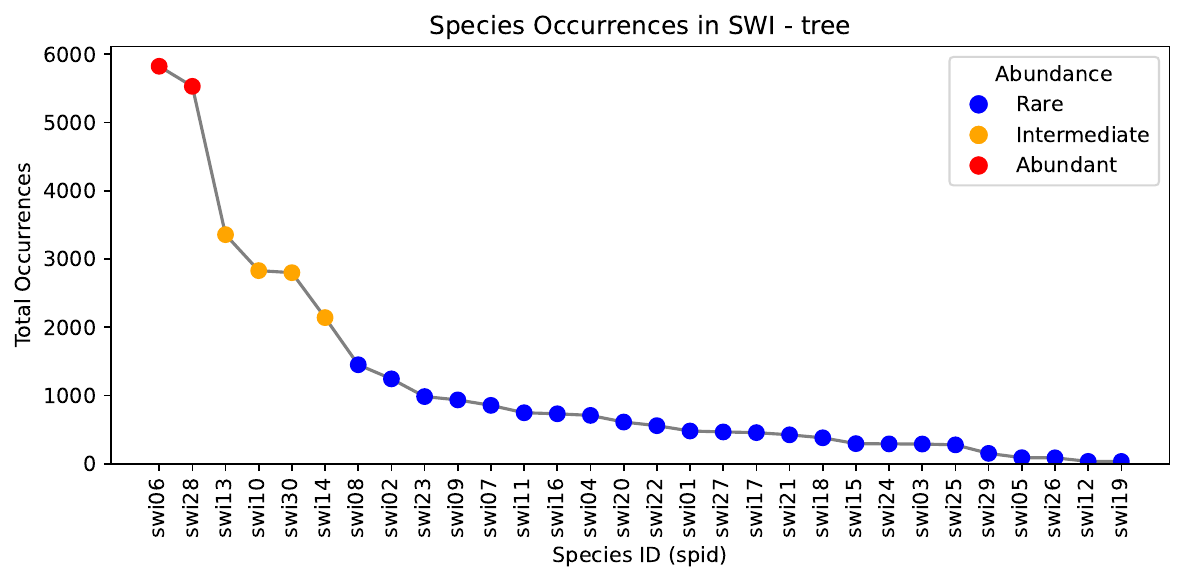}
\caption{}
    \end{subfigure}
            \caption{Species occurrence numbers in PO data for each region and biological group: (a) AWT bird, (b) AWT plant, (c) CAN bird, (d) NSW bates, (e) NSW bird, (f) NSW plant, (g) NSW reptile, (h) NZ plant, (i) SA plant and (j) SWI tree. It's displayed in descending order. Colours correspond to abundance classes: Common, Intermediate, and Rare} 
            \label{annex:distribution_biodivdata}
\end{figure}

\subsection{GeoPlant}

\begin{figure}[h!]
    \centering
    \includegraphics[width=0.65\linewidth]{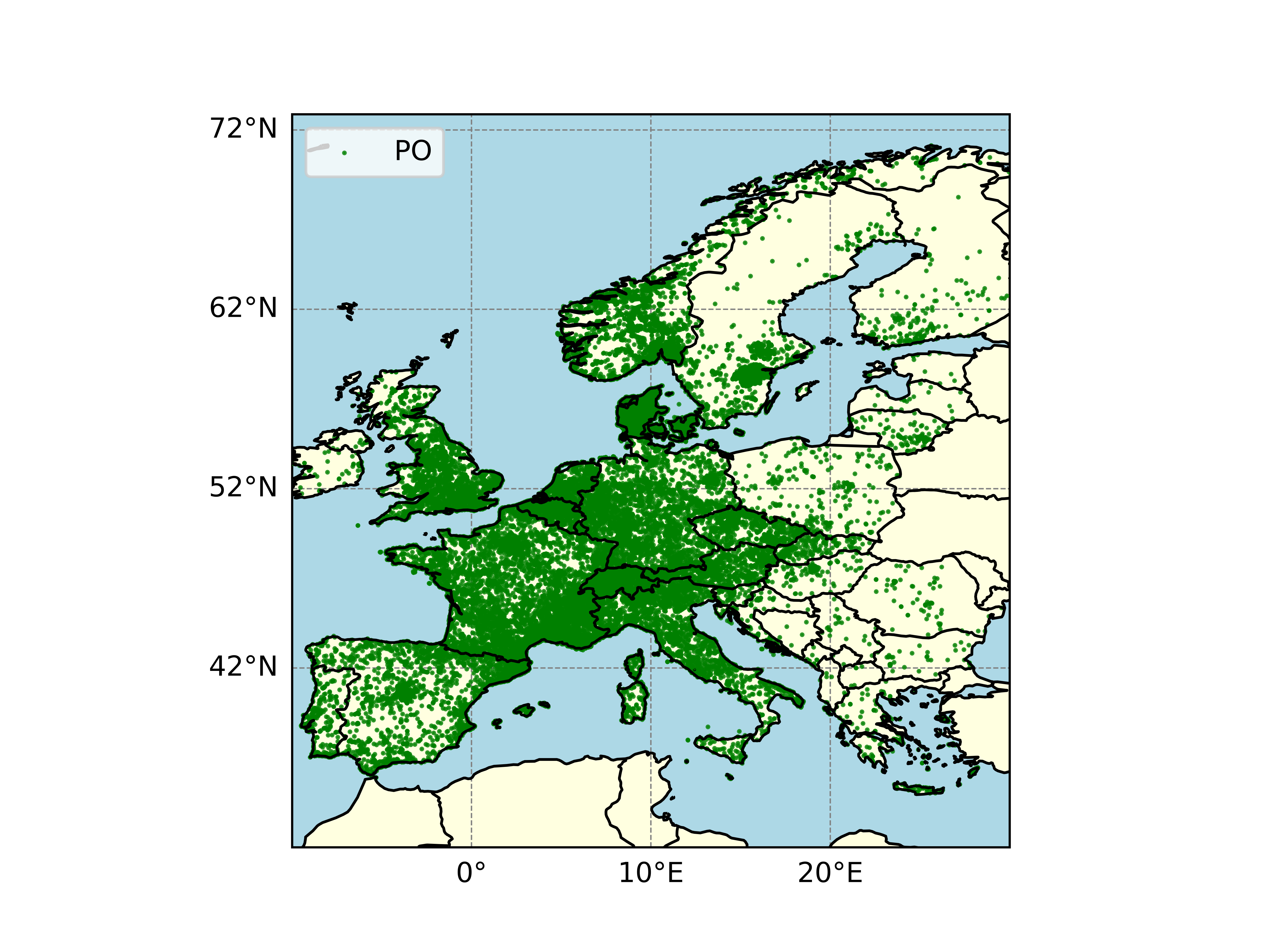}
    \caption{{A map representing 1\% of GeoPlant PO observations.}}
    \label{fig:geoplant_po}
\end{figure}

\begin{figure}[h!]
    \centering
    \includegraphics[width=0.65\linewidth]{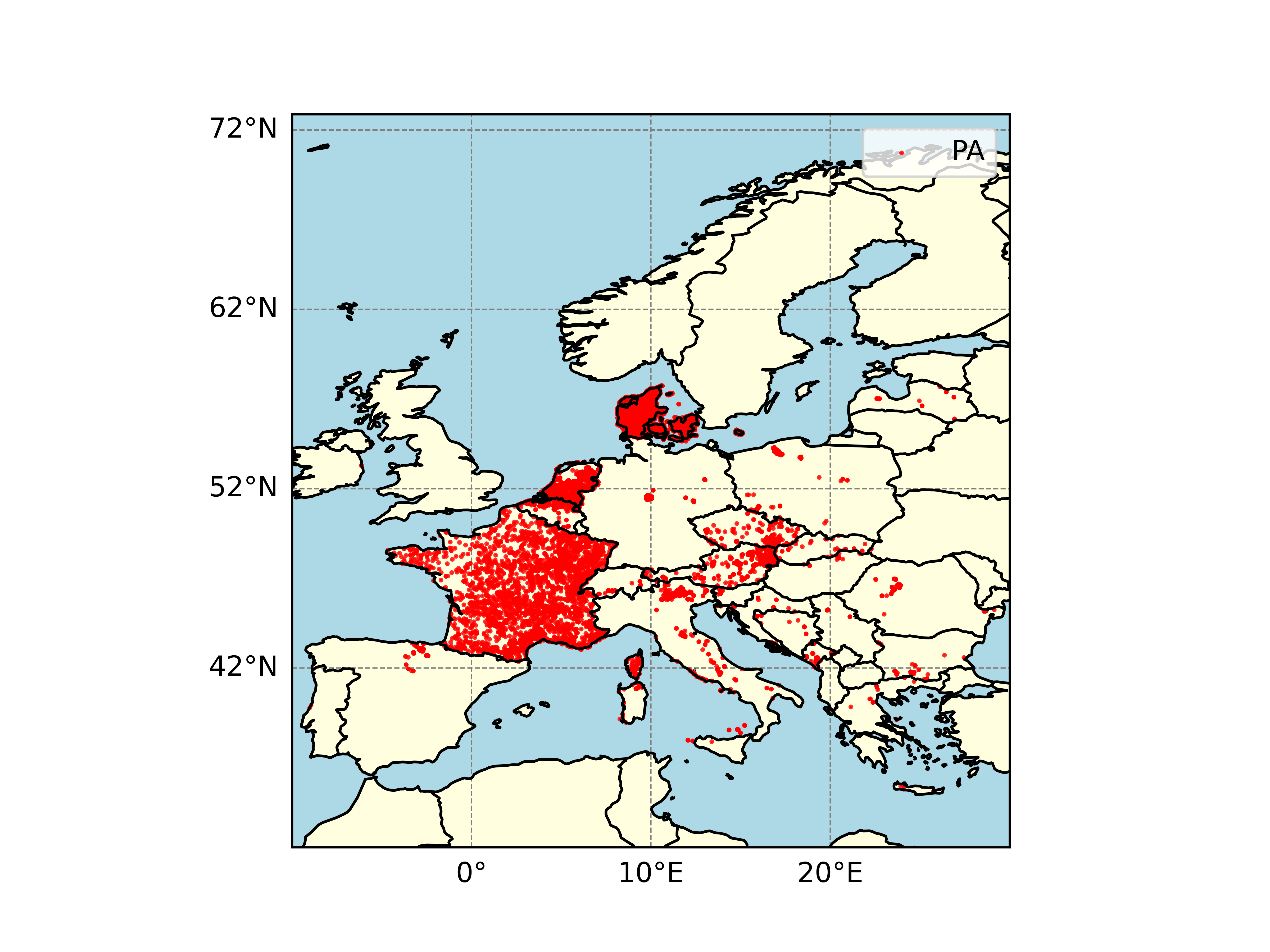}
    \caption{{A map representing 1\% of GeoPlant PA data.}}
    \label{fig:geoplant_pa}
\end{figure}

\begin{figure}[h!]
    \centering
    \includegraphics[width=0.75\linewidth]{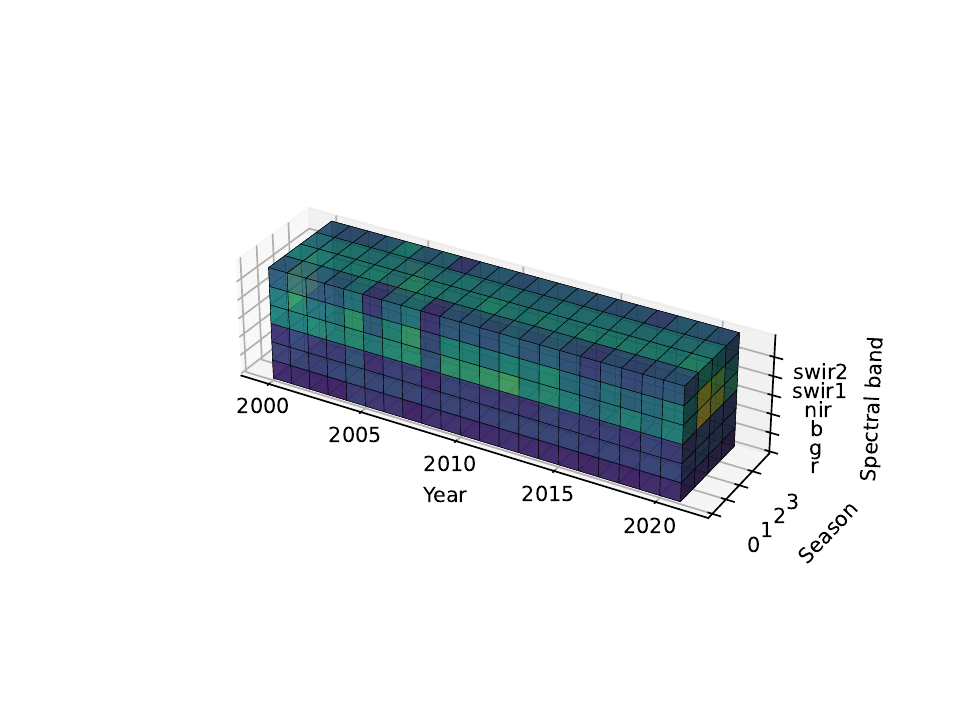}
    \caption{{A cube representation of Landsat time series for a single observation at a specific location (latitude, longitude). The cube has dimensions of 6 spectral bands × 4 seasons × 21 years, where each value represents the seasonal median of a given spectral band for a specific year}}
    \label{fig:landsatCUBE}
\end{figure}

\end{document}